%% file: main_nips.tex
\documentclass{article}

\usepackage[preprint]{neurips_2021}

\input{support_algo}

\input{support_english}

\input{support_figure}
\input{support_linking}

\input{support_math}

\input{support_mathenv}
\input{support_mathvar}

\input{support_refer}

\input{support_table}
\input{support_misc}

\newcommand{\manus}{paper} %

\newif\ifthesis
\newif\ifpaper
\papertrue %

\title{Examining average and discounted reward optimality criteria in reinforcement learning}

\author{%
  Vektor Dewanto, Marcus Gallagher \\
  School of Information Technology and Electrical Engineering \\
  University of Queensland, Australia \\
  \texttt{v.dewanto@uqconnect.edu.au, marcusg@uq.edu.au} \\
}

\begin{document}
\maketitle

\input{abstract}

\input{intro}
\input{prelim}

\input{disrew}

\input{gamma}

\input{avgrew}

\input{discuss}
\input{xprmt_setup.tex}

\bibliography{main}
\bibliographystyle{apalike}

\end{document}

%% file: support_algo.tex
\usepackage[linesnumbered,ruled,vlined]{algorithm2e}

\SetKwComment{Comment}{$\triangleright$\ }{}

\SetKwInput{KwInput}{Input}                %
\SetKwInput{KwOutput}{Output}              %

\SetKw{Andd}{and} %
\SetKw{Break}{break}
\SetKw{Continue}{continue}
\SetKw{Return}{return}

%% file: support_english.tex
\newcommand{\cf}{cf.~}

\newcommand{\eg}{e.g.~}

\newcommand{\first}{\textbf{Firstly},~}
    \newcommand{\second}{\textbf{Secondly},~}
    \newcommand{\third}{\textbf{Third},~}

\newcommand{\ie}{i.e.~}

%% file: support_figure.tex
\usepackage{tikz}
\usetikzlibrary{positioning, shapes.geometric, backgrounds}

\usetikzlibrary{arrows.meta}

\usepackage{subcaption}

\usepackage{grffile}

\usepackage{graphicx} %
\DeclareGraphicsExtensions{.pdf,.jpeg,.png,.jpg}

\usepackage{rotating}

\usepackage{wrapfig}

%% file: support_linking.tex
\usepackage[hidelinks, bookmarksnumbered, backref=page]{hyperref}
\renewcommand*{\backref}[1]{}
\renewcommand*{\backrefalt}[4]{
  \ifcase #1 (Broken backref)
  \or        (page~#2)
  \else      (pages~#2)
  \fi
}

\hypersetup{
  colorlinks   = true,
  urlcolor     = blue,
  linkcolor    = blue,
  citecolor   = blue
}

\usepackage{url}            %

%% file: support_math.tex
\usepackage{scalerel}
\usepackage{amsmath}
\usepackage{amsthm}
\usepackage{amssymb}
\usepackage{amsfonts}       %
\usepackage{mathtools}
\usepackage{cancel}
\usepackage{bbm} %

\usepackage{mathtools}
\DeclarePairedDelimiter\ceil{\lceil}{\rceil}

\newcommand{\abs}[1]{\left|#1\right|}
\newcommand{\clip}{\mathrm{clip}}

\newcommand{\E}[2]{\mathbb{E}_{#1} \left[ #2 \right]} %

\newcommand{\norm}[1]{\|#1\|} %

\newcommand{\prob}[1]{\mathrm{Pr}\{#1\}}

\newcommand{\real}[1]{\mathbb{R}^{#1}}

\newcommand{\setname}[1]{\mathcal{#1}}
\newcommand{\setsize}[1]{|\mathcal{#1}|}

\newcommand{\eqdef}{\coloneqq}
\newcommand{\eqdefr}{\eqqcolon}

\DeclareMathOperator*{\argmax}{\arg\!\max}

\newcommand{\vecb}[1]{\boldsymbol{#1}}
\newcommand{\mat}[1]{\boldsymbol{#1}}

\let\originalleft\left
\let\originalright\right
\renewcommand{\left}{\mathopen{}\mathclose\bgroup\originalleft}
\renewcommand{\right}{\aftergroup\egroup\originalright}

%% file: support_mathvar.tex
\newcommand{\szrat}{s_{\mathrm{zrat}}} %
\newcommand{\azrat}{a_{\mathrm{zrat}}}
\newcommand{\pzrat}{p_{\mathrm{zrat}}}
\newcommand{\rzrat}{r_{\mathrm{zrat}}}
\newcommand{\isdzrat}{\isd_{\mathrm{zrat}}}
\newcommand{\ssetzrat}{ \setname{S}_{\mathrm{zrat}}^+ }
\newcommand{\asetzrat}{ \setname{A}_{\mathrm{zrat}}^+ }

\newcommand{\sreset}{s_{\mathrm{rst}}}
\newcommand{\areset}{a_{\mathrm{rst}}}
\newcommand{\preset}{p_{\mathrm{rst}}}
\newcommand{\rreset}{r_{\mathrm{rst}}}
\newcommand{\isdreset}{\isd_{\mathrm{rst}}}
\newcommand{\ssetreset}{ \setname{S}_{\mathrm{rst}}^+ }
\newcommand{\asetreset}{ \setname{A}_{\mathrm{rst}}^+ }

\newcommand{\sref}{s_{\mathrm{ref}}}

\newcommand{\ppimat}{\mat{P}_{\!\!\pi}} %

\newcommand{\rpivec}{\vecb{r}_{\!\!\pi}}

\newcommand{\vgvec}{\vecb{v}_{\!g}}

\newcommand{\piset}[1]{\Pi_{\mathrm{#1}}}

\newcommand{\isd}{\mathring{p}} %

\newcommand{\tmax}{t_{\mathrm{max}}}
\newcommand{\tmaxrv}{T_{\mathrm{max}}} %
\newcommand{\tepmax}{\hat{t}_{\mathrm{max}}^{\mathrm{xep}}} %
\newcommand{\tepmaxi}{\hat{t}_{\mathrm{max}}^{\mathrm{xep}(i)}} %

\newcommand{\rmax}{r_{\mathrm{max}}}

\newcommand{\nvis}{n_{\mathrm{vis}}} %
\newcommand{\nvisrv}{N_{\mathrm{vis}}} %

\newcommand{\nxep}{n_{\mathrm{xep}}}
\newcommand{\nxeplast}{n_{\mathrm{xep}}^{\mathrm{last}}}

\newcommand{\nrecur}{n_{\mathrm{re}}}

\newcommand{\gammabw}{\gamma_{\mathrm{Bw}}} %
\newcommand{\pistarbw}{\pi^*_{\mathrm{Bw}}}
\newcommand{\pistarbwset}{\Pi^*_{\mathrm{Bw}}}

\newcommand{\psitotasy}{\bar{\psi}_{\mathrm{tot}}^{\mathrm{final}}}

\newcommand{\bigO}[1]{\mathcal{O}\left(#1\right)}

%% file: support_refer.tex
\newcommand{\appref}[1]{Appendix~\ref{#1}} %

\newcommand{\chref}[1]{Chapter~\ref{#1}}

\newcommand{\figref}[1]{Fig~\ref{#1}}

\newcommand{\figrefand}[2]{Figs~\ref{#1} and~\ref{#2}}
\newcommand{\figrefc}[3]{Figs~\ref{#1}, \ref{#2}, and \ref{#3}}

\newcommand{\fnoteref}[1]{Footnote~\ref{#1}}

\newcommand{\secref}[1]{Sec~\ref{#1}}

\newcommand{\secrefc}[3]{Secs~\ref{#1}, \ref{#2} and \ref{#3}}
\newcommand{\secrefand}[2]{Secs~\ref{#1} and \ref{#2}}

\newcommand{\alg}[1]{Algo~#1}

\newcommand{\app}[1]{Appendix~#1}

\newcommand{\ch}[1]{Ch~#1}

\newcommand{\cor}[1]{Corollary~#1} %

\newcommand{\deff}[1]{Def~#1}

\newcommand{\equ}[1]{Eqn~#1}

\newcommand{\fig}[1]{Fig~#1}
\newcommand{\fnote}[1]{Footnote~#1}

\newcommand{\lmm}[1]{Lemma~#1}

\newcommand{\page}[1]{p#1}

\newcommand{\prop}[1]{Prop~#1}

\newcommand{\tbl}[1]{Table~#1}
\newcommand{\secc}[1]{Sec~#1}

\newcommand{\thm}[1]{Thm~#1}

%% file: support_table.tex
\usepackage{booktabs}       %
\usepackage{multirow}
\usepackage{longtable}

\usepackage{colortbl}

%% file: support_misc.tex
\usepackage{enumitem}
\setlist{nosep}

\usepackage{xcolor}
\colorlet{darkgreen}{green!50!black}

\definecolor{stateblue}{cmyk}{0.96,0,0,0}
\definecolor{lightgray}{gray}{0.85}
\definecolor{halfgreen}{rgb}{0,0.5,0}

\usepackage{pdflscape}
\usepackage{afterpage}

\usepackage{nicefrac}       %
\usepackage{microtype}      %
\usepackage[yyyymmdd,hhmmss]{datetime}
\usepackage{lipsum}

\usepackage[utf8]{inputenc} %
\usepackage[T1]{fontenc}    %

\usepackage{import}

\usepackage{epigraph}

\usepackage{acronym}

\usepackage{multicol}

%% file: abstract.tex
\begin{abstract}
In reinforcement learning (RL), the goal is to obtain an optimal policy,
for which the optimality criterion is fundamentally important.
Two major optimality criteria are average and discounted rewards.
While the latter is more popular, it is problematic to apply in environments
without an inherent notion of discounting.
This motivates us to revisit
a)~the progression of optimality criteria in dynamic programming,
b)~justification for and complication of an artificial discount factor, and
c)~benefits of directly maximizing the average reward criterion, which is discounting-free.
Our contributions include a thorough examination of the relationship between
average and discounted rewards, as well as a discussion of their pros and cons in~RL.
We emphasize that average-reward RL methods possess the ingredient and mechanism
for applying a family of discounting-free optimality criteria
\citep{veinott_1969_sen} to~RL.
\end{abstract}

%% file: intro.tex
\section{Introduction}

Reinforcement learning (RL) is concerned with sequential decision making,
where a decision maker has to choose an action based on its current state.
Determining the best actions amounts to finding an \emph{optimal} mapping from
every state to a probability distribution over actions available at that state.
Thus, one fundamental component of any RL method is the optimality criterion,
by which we define what we mean by such an optimal mapping.

The most popular optimality criterion in RL is the discounted reward
\citep{duan_2016_benchrl, machado_2018_arcade, henderson_2018_rlmatter}.
On the other hand, there is growing interest in the average reward optimality,
\ifpaper
as surveyed by \citet{mahadevan_1996_avgrew, dewanto_2020_avgrew}.
\fi
In this \manus, we discuss both criteria in order to obtain
a comprehensive understanding of their properties, relationships, and differences.
This is important because the choice of optimality criteria affects
almost every aspect of RL methods, including the policy evaluation function,
the policy gradient formulation, and the resulting optimal policy,
where the term policy refers to the above-mentioned mapping.
Thus, the choice of optimality criteria eventually impacts the performance of an RL system
(and the choices made within, \eg approximation techniques and hyperparameters).

This \manus~presents a thorough examination of the connection between
average and discounted rewards, as well as a discussion of their pros and cons in RL.
Our examination here is devised through broader lens of refined optimality criteria
(which generalize average and discounted rewards), inspired by
the seminal work of \citet{mahadevan_1996_sensitivedo}.
It is also broader in the sense of algorithmic styles: value- and policy-iteration,
as well as of tabular and function approximation settings in RL.
In this \manus, we also attempt to compile and unify various justifications for
deviating from the common practice of maximizing the discounted reward criterion in RL.

There are a number of papers that specifically compare and contrast
average to discounted rewards in RL.
For example, \citet{mahadevan_1994_cmpqr} empirically investigated average versus
discounted reward Q-learning (in the context of value-iteration RL).
\citet{tsitsiklis_2002_avgdisctd} theoretically compared
average versus discounted reward temporal-difference (TD) methods for
policy evaluation (in the context of policy-iteration RL).
We provide updates and extensions to those existing comparison works
in order to obtain a comprehensive view on discounting and discounting-free RL.

Since RL is \emph{approximate} dynamic programming (DP), we begin with
reviewing optimality criteria in DP,
model classification (which plays an important role in average-reward optimality),
and the interpretation of discounting in \secref{sec:prelim_avgdirew}.
We then provide justifications for discounting for
environments without any inherent notion of discounting (\secref{sec:disrew}).
This is followed by the difficulties that arise from introducing
an artificial discount factor (\secref{sec:gamma}).
We discuss the benefits from maximizing the average-reward criterion in \secref{sec:avgrew},
as well as our finding and viewpoint in \secref{sec:discuss_avgdisrew}.

%% file: prelim.tex
\section{Preliminaries} \label{sec:prelim_avgdirew}

Sequential decision making is often formulated as a Markov decision process (MDP)
with a state set~$\setname{S}$, an action set~$\setname{A}$, a reward set~$\setname{R}$,
and a decision-epoch set~$\setname{T}$.
Here, all but~$\setname{T}$ are finite sets,
yielding an infinite-horizon finite MDP.
At the current decision-epoch $t \in \setname{T}$,
a decision maker (henceforth, an agent) is in a current state $s_t \in \setname{S}$,
and chooses to then execute a current action $a_t \in \setname{A}$.
Consequently at the next decision-epoch $t+1$, it arrives
in the next state $s_{t+1} \in \setname{S}$ and
earns an (immediate) scalar reward $r_{t+1} \in \real{}$.
The decision-epochs occur every single time unit (\ie timestep)
from $t=0$ till the maximum timestep (denoted as $\tmax$);
hence, we have a discrete-time MDP.

For $t= 0, 1, \ldots, \tmax = \infty$, the agent experiences
a sequence (trajectory) of states~$s_t$, actions~$a_t$ and rewards~$r_{t+1}$.
That is, $s_0, a_0, r_1, s_1, a_1, r_2, \ldots, s_{\tmax}$.
The initial state, the next state, and the next reward are
governed by the environment dynamics that is fully specified by
three time-homogenous (time-invariant) entities:
i)~the initial state distribution $\isd$, from which $S_{t=0} \sim \isd$,
ii)~the one-(time)step state transition distribution $p(\cdot| s_t, a_t)$,
from which $S_{t+1} | S_t = s_t, A_t = a_t \sim p(\cdot| s_t, a_t)$, and
iii)~the reward function $r(s_t, a_t)
= \E{p(\cdot| s_t, a_t)}{\sum_{r \in \setname{R}} \prob{r | s_t, a_t, S_{t+1}} \cdot r}
\eqdefr r_{t+1}$,
where $S_t$ and $A_t$ denote the state and action random variables, respectively.
The existence of a probability space that holds this infinite sequence of random variables
($S_0, A_0, S_1, A_1, \ldots$) can be shown using the Ionescu-Tulcea theorem
\citep[\page{513}]{lattimore_2020_bandit}.
We assume that the rewards are bounded,
\ie  $| r(s, a) |\le \rmax < \infty, \forall (s, a) \in \setname{S} \times \setname{A}$,
and the MDP is unichain and aperiodic (see \secref{sec:mdpclf} for MDP classification).

The solution to a sequential decision making problem is an \emph{optimal} mapping
from every state to a probability distribution over the (available) action set
$\setname{A}$ in that state.\footnote{
    In general, different states may have different action sets.
}
It is optimal with respect to some optimality criterion,
as discussed later in \secref{sec:optcrit}.
Any of such a mapping (regardless whether it is optimal) is called a \emph{policy}
and generally depends on timestep~$t$.
It is denoted as $\pi_t: \setname{S} \mapsto [0, 1]^{\setsize{A}}$, or alternatively
$\pi_t: \setname{S} \times \setname{A} \mapsto [0, 1]$,
where $\pi_t(a_t|s_t) \eqdef \prob{A_t = a_t|S_t = s_t}$ indicating
the probability of selecting action $a_t \in \setname{A}$ given
the state $s_t \in \setname{S}$ at timestep~$t \in \setname{T}$.
Thus, each action is sampled from a conditional action distribution,
\ie $A_t | S_t = s_t \sim \pi_t(\cdot|s_t)$.

The most specific solution space is
the stationary and deterministic policy set $\piset{SD}$
whose policies $\pi \in \piset{SD}$ are \emph{stationary} (time-invariant),
\ie $\pi \eqdef \pi_0 = \pi_1 = \ldots = \pi_{\tmax - 1}$, as well as
\emph{deterministic}, \ie $\pi(\cdot|s_t)$ has a single action support
(hence the mapping is reduced to $\pi: \setname{S} \mapsto \setname{A}$).
In this work, we consider a more general policy set,
that is the stationary policy set $\piset{S}$.
It includes the stationary randomized (stochastic) set~$\piset{SR}$ and
its degenerate counterpart: the stationary deterministic set~$\piset{SD}$.

\input{prelim_optcrit}
\input{prelim_mdpclf}

\input{prelim_origamma}

%% file: prelim_optcrit.tex
\subsection{Optimality criteria} \label{sec:optcrit}

In a basic notion of optimality, a policy with the largest \emph{value} is optimal.
That is,
\begin{equation}
v_x(\pi_x^*) \ge v_x(\pi),\quad \forall \pi \in \piset{S}.
\label{equ:opt_vf}
\end{equation}
Here, the function $v_x$ measures the value (utility) of a policy~$\pi$ based on
the \emph{infinite} reward sequence that is earned by an agent following~$\pi$.
The subscript $x$ indicates the specific type of value functions, which
induces a specific $x$-optimality criterion,
hence the $x$-optimal policy, denoted as $\pi_x^*$.

One intuitive\footnote{
    This intuition seems to lead to the so-called reward hypothesis
    \citep[\page{53}]{sutton_2018_irl}.
}
value function is the \emph{expected} total reward.
That is,
\begin{equation}
v_{\mathrm{tot}}(\pi, s) \eqdef
    \lim_{\tmax \to \infty}
    \E{A_t \sim \pi(\cdot|s_t), S_{t+1} \sim p(\cdot|s_t, a_t)}{
        \sum_{t = 0}^{\tmax - 1} r(S_t, A_t) \Big| S_0 = s, \pi},
\quad \forall \pi \in \piset{S}, \forall s \in \setname{S}.
\label{equ:v_total}
\end{equation}
However, $v_{\mathrm{tot}}$ may be infinite
(unbounded, divergent, non-summable).\footnote{
    If a limit is equal to infinity, then we assert that such a limit does not exist.
    No comparison can be made between finite and infinite policy values,
    as well as between infinite policy values.
}
\citet{howard_1960_dpmp} therefore, examined the \emph{expected} average reward
(also referred to as the \emph{gain}) defined as
\begin{align}
v_g(\pi, s)
& \eqdef \lim_{\tmax \to \infty} \frac{1}{\tmax}
    \E{A_t \sim \pi(\cdot|s_t), S_{t+1} \sim p(\cdot|s_t, a_t)}
        {\sum_{t = 0}^{\tmax - 1} r(S_t, A_t) \Big| S_0 = s, \pi},
\label{equ:v_gain}
\end{align}
which is finite for all $\pi \in \piset{S}$ and all $s \in \setname{S}$.
For more details, including interpretation about the gain,
we refer the reader to
\ifpaper
\citep{dewanto_2020_avgrew}.
\fi

Alternatively, \citet[\secc{4}]{blackwell_1962_ddp} attempted tackling
the infiniteness of \eqref{equ:v_total} through
the \emph{expected} total discounted reward\footnote{
    There are other discounting schemes, %
    \eg $1/(t^2 + t)$,
    $1/(1 + \kappa t)$ for $\kappa > 0$ (hyperbolic), and
    $t^{-\kappa}$ for $\kappa > 1$
    \citep{hutter_2006_gdisc, lattimore_2014_disc}, but they do not exhibit
    advantageous properties as the geometric discounting \eqref{equ:v_disc},
    such as its decomposition via \eqref{equ:laurent_expansion}.
    Hence, those other schemes are not considered here.
},
which was studied before by \citet{bellman_1957_dp}. %
That is,
\begin{equation}
v_\gamma(\pi, s) \eqdef
    \lim_{\tmax \to \infty}
    \E{A_t \sim \pi(\cdot|s_t), S_{t+1} \sim p(\cdot|s_t, a_t)}
        {\sum_{t = 0}^{\tmax - 1} \gamma^t r(S_t, A_t) \Big| S_0 = s, \pi},
\quad \forall \pi \in \piset{S}, \forall s \in \setname{S},
\label{equ:v_disc}
\end{equation}
with a discount factor $\gamma \in [0, 1)$.
In particular, according to what is later known as
the truncated Laurent series expansion \eqref{equ:laurent_expansion_truncated},
Blackwell suggested finding policies that are $\gamma$-discounted optimal
for all discount factors $\gamma$ \emph{sufficiently} close to~1.
He also established their existence in finite MDPs.
Subsequently, \citet{smallwood_1966_reg} identified that the discount factor interval
can be divided into a \emph{finite} number of intervals\footnote{
    See also \citet{hordijk_1985_disc} and \citet[\thm{2.16}]{feinberg_2002_hmdp}.
},
\ie $[0 = \gamma_{m}, \gamma_{m - 1}), [\gamma_{m - 1}, \gamma_{m - 2}), \ldots,
[\gamma_0, \gamma_{-1} = 1)$,
in such a way that there exist policies $\pi^*_{\gamma_i}$ for $0 \le i \le m$,
that are $\gamma$-discounted optimal for all $\gamma \in [\gamma_i, \gamma_{i-1})$.
This leads to the concept of Blackwell optimality.
A policy $\pistarbw$ is Blackwell optimal if there exists
a critical\footnote{
    It is critical in that it specifies the sufficiency of being close to~1
    for attaining Blackwell optimal policies.
} discount factor $\gammabw \in [0, 1)$ such that
\begin{equation}
v_\gamma(\pistarbw, s) \ge v_\gamma(\pi, s), \quad
\text{for}\ \gammabw \le \gamma < 1,\
\text{and for all $\pi \in \piset{S}$ and all $s \in \setname{S}$}.
\label{equ:bw_optim}
\end{equation}
Note that whenever the policy value function $v$ depends not only on policies $\pi$
but also on states $s$ from which the value is measured,
the basic optimality \eqref{equ:opt_vf} requires that the optimal policy has
a value greater than or equal to the other policies' values in all state $s \in \setname{S}$.

In order to obtain Blackwell optimal policies, \citet[\equ{27}]{veinott_1969_sen}
introduced a family of new optimality criteria.
That is, a policy $\pi_n^*$ is $n$-discount optimal for $n = -1, 0, \ldots$, if
\begin{equation*}
\lim_{\gamma \to 1} \frac{1}{(1 - \gamma)^n}
    \Big( v_\gamma(\pi_n^*, s) - v_\gamma(\pi, s) \Big) \ge 0,
\qquad \forall \pi \in \piset{S}, \forall s \in \setname{S}.
\end{equation*}
He showed that $(n=\setsize{S})$-discount optimality\footnote{
    This is eventually refined to $(n=\setsize{S} - \nrecur(\pistarbw))$-discount optimality,
    where $\nrecur(\pistarbw)$ is the number of recurrent classes under
    the Blackwell optimal policy \citep[\thm{8.3}, \secc{8.1.4}]{feinberg_2002_hmdp}.
} is equivalent to Blackwell optimality because the selectivity increases with~$n$
(hence, $n_i$-discount optimality implies $n_j$-discount optimalities for all $i > j$).
Then, he developed a policy-iteration algorithm (for finding $n$-discount optimal policies)
that utilizes the full Laurent series expansion of~$\vecb{v}_\gamma(\pi)$
for any $\pi \in \piset{S}$, where the policy value vector
$\vecb{v}_x(\pi) \in \real{\setsize{S}}$ is obtained by stacking
the state-wise values $v_x(\pi, s)$ altogether for all $s \in \setname{S}$.
That is,
\begin{align}
\vecb{v}_\gamma(\pi)
& = \frac{1}{\gamma} \Big(
    \frac{\gamma}{1 - \gamma} \vecb{v}_{-1}(\pi) + \vecb{v}_0(\pi)
    + \sum_{n=1}^{\infty} \Big( \frac{1 - \gamma}{\gamma} \Big)^n \vecb{v}_n(\pi)
    \Big)
    \hspace{20mm} \text{(Full expansion)}
    \label{equ:laurent_expansion} \\
& = \frac{1}{1 - \gamma} \vecb{v}_{-1}(\pi) + \vecb{v}_0(\pi)
    + \Bigg\{ \frac{1 - \gamma}{\gamma} \vecb{v}_0(\pi)
        + \frac{1}{\gamma} \sum_{n=1}^{\infty}
            \Big( \frac{1 - \gamma}{\gamma} \Big)^n \vecb{v}_n(\pi) \Bigg\}
    \tag{Using the additive identity: $\vecb{v}_0 - \vecb{v}_0 = \vecb{0}$} \\
& = \frac{1}{1 - \gamma} \vecb{v}_{-1}(\pi) + \vecb{v}_0(\pi)
    + \bigg\{ \vecb{e}(\pi, \gamma) \bigg\},
    \hspace{31mm} \text{(Truncated expansion)}
    \label{equ:laurent_expansion_truncated}
\end{align}
where $\vecb{v}_n \in \real{\setsize{S}}$ for $n=-1, 0, \ldots$ denotes
the expansion coefficients.
The truncated form first appeared (before the full one) in
\citet[\thm{4a}]{blackwell_1962_ddp}, where
$\vecb{v}_{-1}$ is equivalent to the gain from all states $\vgvec \in \real{\setsize{S}}$,
$\vecb{v}_0$ is equivalent to the so-called bias, and
$\vecb{e}(\pi, \gamma)$ converges to~$\vecb{0}$ as $\gamma$ approaches~1,
that is $\lim_{\gamma \to 1} \bigO{1 - \gamma} = \vecb{0}$.

%% file: prelim_mdpclf.tex
\subsection{MDP classification} \label{sec:mdpclf}

A stationary policy~$\pi$ of a finite MDP induces
a stationary finite Markov chain (MC)
with a $\setsize{S}$-by-$\setsize{S}$ one-step transition stochastic matrix
$\ppimat$, whose $[s, s']$-th entry indicates
\begin{equation}
p_\pi(s'|s) \eqdef \prob{S_{t+1} = s' | S_{t} = s; \pi}
= \sum_{a \in \setname{A}} p(s'|s, a) \pi(a|s),
\quad \forall s, s' \in \setname{S}.
\label{equ:p_pi}
\end{equation}
Therefore, an MDP can be classified on the basis of the set of MCs induced
by its stationary policies.
To this end, we need to review the classification of states, then of MCs below;
mainly following \citet[\app{A}]{puterman_1994_mdp} and
\citet[\ch{6.2}]{douc_2018_markov}.

\paragraph{\textbf{State classification:}}
Let $\ppimat^t$ denote the $t$-th power of $\ppimat$ of a Markov chain.
This $\ppimat^t$ is another $\setsize{S}$-by-$\setsize{S}$ stochastic matrix,
whose $[s, s']$-entry indicates
$p_\pi^t(s'|s) \eqdef \prob{S_{t} = s' | S_{0} = s; \pi}$,
\ie the probability of being in state $s' \in \setname{S}$ in $t$ timesteps
starting from a state $s \in \setname{S}$.

A state $s' \in \setname{S}$ is \emph{accessible} from state $s \in \setname{S}$,
denoted as $s \to s'$, if $p_\pi^t(s'|s) > 0$ for some $t \ge 0$.
Furthermore, $s$ and $s'$ \emph{communicate} if $s \to s'$ and $s' \to s$.

A subset $\tilde{\setname{S}}$ of $\setname{S}$ is called \emph{a closed set}
if no state outside $\tilde{\setname{S}}$ is accessible from any state
in $\tilde{\setname{S}}$.
Furthermore, a subset $\tilde{\setname{S}}$ of~$\setname{S}$ is called
\emph{a recurrent class} (or a recurrent chain)
if all states in $\tilde{\setname{S}}$ communicate
and $\tilde{\setname{S}}$ is closed.
In a finite MC, there exists at least one recurrent class.

Let $\nvis^{s,\pi}$ denote the number of visits to a state $s$ under a policy $\pi$.
That is, $\nvis^{s,\pi} \eqdef {\sum_{t=0}^\infty \mathbb{I}[S_t = s | \pi]}$,
where $\mathbb{I}$ is an identity (indicator) operator.
Then, the \emph{expected} number of visits to a state $s \in \setname{S}$ starting
from the state $s$ itself under a policy $\pi$ is given by
$\E{}{\nvisrv^{s,\pi}} = \sum_{t=0}^\infty p_\pi^t(s|s)$.

A state $s \in \setname{S}$ is \emph{recurrent} if and only if
$\E{}{\nvisrv^{s,\pi}} = \infty$.
This means that a recurrent state will be re-visited infinitely often.
Equivalently, starting from a recurrent state $s \in \setname{S}$,
the probability of returning to $s$ itself for the first time in finite time is~1.

A state $s \in \setname{S}$ is \emph{transient} if and only if
$\E{}{\nvisrv^{s,\pi}} < \infty$.
This means that a transient state will never be re-visited again
after some point in time.
Equivalently, starting from a transient state $s \in \setname{S}$, the probability of
returning to $s$ itself for the first time in finite time is less than 1.
A transient state is a non-recurrent state
(hence, every state is either recurrent or transient).

The period of a state~$s \in \setname{S}$ is defined as the greatest common divisors
$t_{\mathrm{gcd}}$ of all~$t \ge 1$ for which $p_\pi^t(S_t = s | S_0=s) > 0$.
Whenever the period $t_{\mathrm{gcd}} > 1$, we clasify $s$ as \emph{periodic}.
Otherwise, $s$ is \emph{aperiodic} (not periodic).
Intuitively, if returning to a state~$s$ occurs at irregular times,
then $s$ is aperiodic.
Note that periodicity is a class property, implying that
all states in a recurrent class have the same period.

An \emph{ergodic class} (or an ergodic chain) is a class that is
both recurrent and aperiodic.
Its recurrent and aperiodic states are referred to as \emph{ergodic states}.
Note however, that the term ``ergodic'' is also used in the ergodic theory (mathematics),
in which its definition does not involve the notion of aperiodicity.

\paragraph{\textbf{Markov chain (MC) classification:}}
An MC is \emph{irreducible} if the set of all its states forms
a single recurrent class.
An irreducible MC is aperiodic if all its states are aperiodic.
Otherwise, it is periodic.
An MC is \emph{reducible} if it has both recurrent states and transient states.
The state set of a reducible MC can be partitioned into
one or more disjoint recurrent classes, plus a set of transient states.

An MC is \emph{unichain} if it consists of a single (uni-) recurrent class (chain),
plus a (possibly empty) set of transient states.
Otherwise, it is \emph{multichain}.

\paragraph{\textbf{MDP classification based on the pattern of MCs
induced by all stationary policies:}}
An MDP is unichain if the induced MC corresponding to every stationary policy
is unichain.
Otherwise, it is multichain (\ie at least one induced MC is multichain,
containing two or more recurrent classes).
This is a restrictive definition of a multichain MDP.
In the literature, its loose definition is also used, where
it simply means a \emph{general} MDP.

A \emph{recurrent MDP} is a special case of unichain MDPs,
whenever the MC corresponding to every stationary policy is irreducible.
In other words, a recurrent MDP is a unichain MDP whose all induced MCs have
an empty transient state set.

\paragraph{\textbf{MDP classification based on the pattern of state accessibility
under some stationary policy:}}
An MDP is \emph{communicating} (also termed \emph{strongly connected}) if,
for every pair of states $s$ and $s'$ in the state set $\setname{S}$,
there exists a stationary policy (which may depend on $s$ and $s'$)
under which $s'$ is accessible from $s$.
A recurrent MDP is always communicating,
but a communicating MDP may not be recurrent, it may be unichain or multichain.

An MDP is \emph{weakly communicating} (also termed \emph{simply connected}) if
there exists a closed state set $\tilde{\setname{S}} \subseteq \setname{S}$,
for which there exists a stationary policy under which
$s$ is accessible from $s'$ for every pair of states $s, s' \in \tilde{\setname{S}}$,
plus a (possibly empty) set of transient states under every stationary policy.
This weakly communicating classification is more general than the communicating
in that any communicating MDP is weakly communicating without
any state that is transient under every stationary policy.
Unichain MDPs are always weakly communicating.

There exists an MDP that is \emph{not weakly communicating}.
Such a non-weakly-communicating MDP is always multichain,
but a multichain MDP may also be weakly communicating (or communicating).

%% file: prelim_origamma.tex
\subsection{Discounting in environments with an inherent notion of discounting}
\label{sec:origamma}

A environment with an inherent notion of discounting has
a discount factor~$\gamma$ that encodes one of the following entities.
Thus, $\gamma$ is part of the environment specification (definition, description).

\textbf{Firstly}, \emph{the time value of rewards, \ie the value of a unit reward
$t$ timesteps in the future is $\gamma^t$:}
This is related to psychological concepts.
For example, some people prefer rewards now rather than latter \citep{mischel_1972_grat},
hence they assign greater values to early rewards through a small $\gamma$ (being shortsighted).
It is also natural to believe that there is more certainty about near- than far-future,
because immediate rewards are (exponentially more) likely due to recent actions.
The time preference is also well-motivated in economics \citep{samuelson_1937_util}.
This includes $\gamma$ for taking account of the decreasing value of money
(because of inflation), as well as
the interpretation of $(1 - \gamma)/\gamma$ as a positive interest rate.
Moreover, commercial activities have failure (abandonment) risk due to
changing government regulation and consumer preferences over time.

\textbf{Secondly}, \emph{the uncertainty about random termination independent of the agent's actions:}
Such termination comes from external control beyond the agent, \eg
someone shutting down a robot, engine failure (due to weather/natural disaster),
or death of any living organisms.

In particular, whenever the random termination time $\tmaxrv$ follows
a geometric distribution ${Geo(p= 1 - \gamma)}$, we have the following identity
between the total and discounted rewards, %
\begin{equation}
v_{\tmaxrv}(\pi, s)
\eqdef \E{S_t, A_t}{
    \E{\tmaxrv}{\sum_{t = 0}^{\tmaxrv - 1} r(S_t, A_t) \Big| S_0 = s, \pi}}
= \underbrace{v_\gamma(\pi, s)}_\text{See \eqref{equ:v_disc}},
\quad \forall \pi \in \piset{S}, \forall s \in \setname{S},
\label{equ:rndtermination}
\end{equation}
where the discount factor $\gamma$ plays the role of
the geometric distribution parameter \citep[\prop{5.3.1}]{puterman_1994_mdp}.
This discounting implies that at every timestep (for any state-action pair),
the agent has a probability of $(1 - \gamma)$ for entering
the 0-reward absorbing terminal state, see
\ifpaper
\figref{gridnav2_szrat}.
\fi
Note that because $\gamma$ is invariant to states and actions (as well as time),
this way of capturing the randomness of~$\tmaxrv$ may be inaccurate in cases
where termination depends on states, actions, or both.

%% file: disrew.tex
\section{Discounting without an inherent notion of discounting}
\label{sec:disrew}

From now on, we focus on environments without \emph{inherent} notion of discounting,
where $\gamma$ is not part of the environment specification
(\cf \secref{sec:origamma}).
We emphasize the qualification ``inherent'' since any MDP can always be thought of
having some notion of discounting from the Blackwell optimality point of view
\eqref{equ:bw_optim}.
This is because a Blackwell optimal policy is guaranteed to exist in finite MDPs
\citep[\thm{10.1.4}]{puterman_1994_mdp};
implying the existence of its discount factor $\gammabw \in (0, 1]$ and
of a (potentially very long) finite-horizon MDP model that gives exactly
the same Blackwell-optimal policy as its infinite-horizon counterpart
\citep[\ch{1.3}]{chang_2013_mdp}.

When there is no inherent notion of discounting, the discount factor $\gamma$
is imposed for bounded sums (\secref{sec:optcrit}) and
becomes part of the solution method (algorithm).
This is what we refer to as \emph{artificial} discounting, which induces
\emph{artificial} interpretation as for instance, those described
in \secref{sec:origamma}.
The $\gammabw$ (mentioned in the previous paragraph) is
one of such artificial discount factors.
In addition to bounding the sum, we observe other justifications that
have been made for introducing an artificial~$\gamma$.
They are explained in the following \secrefc{sec:avgrewapprox}{sec:disrew_bwoptim}{sec:discrew_mathplus}.

\input{disrew_approxgain.tex}
\input{disrew_bwoptim.tex}
\input{disrew_mathplus.tex}

%% file: disrew_approxgain.tex
\subsection{Approximation to the average reward (the gain) as $\gamma$ approaches 1}
\label{sec:avgrewapprox}

For recurrent MDPs, the gain optimality is the most selective\footnote{
    For an intuitive explanation about the selectivity of optimality criteria,
    refer to \figref{fig:unichain_zrat_n1}.
}
because there are no transient states \citep[\secc{3.1}]{feinberg_2002_hmdp}.
This implies that a gain optimal policy is also Blackwell optimal in recurrent MDPs,
for which one should target the gain optimality criterion.
Nonetheless, the following relationships exist between the average reward $v_g$
and discounted reward~$v_\gamma$ value functions of a policy $\pi \in \piset{S}$.

\textbf{Firstly}, for every state $s_0 \in \setname{S}$,
\begin{align}
v_g(\pi, s_0)
& = \lim_{\gamma \to 1}\ (1 - \gamma)\ v_\gamma(\pi, s_0)
    \hspace{30mm} \text{\citep[\cor{8.2.5}]{puterman_1994_mdp}}
    \label{equ:gain2disrew} \\
& = (1 - \gamma) \sum_{s \in \setname{S}} p_\pi^\star(s|s_0)\ v_\gamma(\pi, s),
    \quad \forall \gamma \in [0, 1),
    \hspace{6mm} \text{\citep[\secc{5.3}]{singh_1994_rlpomdp}}
    \label{equ:gain2disrew_2}
\end{align}
where $p_\pi^\star(s|s_0)$ denotes the stationary probability of a state $s$, \ie
the long-run (steady-state) probability of being in state $s$ when the MC begins in $s_0$.
Here, \eqref{equ:gain2disrew} is obtained by multiplying
the left and right hand sides of \eqref{equ:laurent_expansion_truncated}
by $(1 - \gamma$), then taking the limit of both sides as $\gamma \to 1$.
The derivation of \eqref{equ:gain2disrew_2} begins with taking
the expectation of $v_\gamma(\pi, S)$ with respect to $p_\pi^\star$,
then utilizes the discounted-reward Bellman (expectation) equation,
and the stationarity of $p_\pi^\star$.
That is,
\begin{align*}
\sum_{s \in \setname{S}} p_\pi^\star(s|s_0) \bigg\{ v_\gamma(\pi, s) \bigg\}
& = \sum_{s \in \setname{S}} p_\pi^\star(s|s_0) \bigg\{ \underbrace{
    r_\pi(s) + \gamma \sum_{s' \in \setname{S}} p_\pi(s'|s) v_\gamma(\pi, s')
    }_\text{$v_\gamma(\pi, s)$ via the Bellman equation}
    \bigg\} \\
& = \underbrace{
        \sum_{s \in \setname{S}} p_\pi^\star(s|s_0) r_\pi(s)
        }_\text{Gain $v_g(\pi, s_0)$ in \eqref{equ:v_gain}}
    +\ \gamma \sum_{s' \in \setname{S}} \underbrace{
        \sum_{s \in \setname{S}} p_\pi^\star(s|s_0) p_\pi(s'|s)
        }_\text{$p_\pi^\star(s'|s_0)$ due to stationarity}
        v_\gamma(\pi, s'),
\end{align*}
which can be re-arranged to obtain \eqref{equ:gain2disrew_2}.
Here, $r_\pi(s) \eqdef \sum_{a \in \setname{A}} \pi(a|s) r(s, a)$ whereas
$p_\pi(s'|s)$ is defined in \eqref{equ:p_pi}.
It is interesting that any discount factor $\gamma \in [0, 1)$ maintains
the equality in \eqref{equ:gain2disrew_2}, which was also proved by
\citet[\page{254}]{sutton_2018_irl}.

\textbf{The second relationship} pertains to the gradient of the gain when
a parameterized policy $\pi(\vecb{\theta})$ is used.
By notationally suppressing the policy parameterization $\vecb{\theta}$ and
the dependency on $s_0$, as well as using $\nabla \eqdef \partial/\partial \vecb{\theta}$,
this relation can be expressed as
\begin{align}
\nabla v_g(\pi)
& = \lim_{\gamma \to 1} \Big\{
    \sum_{s \in \setname{S}} \sum_{s' \in \setname{S}}
    p_\pi^\star(s) \underbrace{
        \sum_{a \in \setname{A}} p(s'|s, a) \pi(a|s) \nabla \log \pi(a|s)
        }_{\nabla p_\pi(s'|s)}
        v_\gamma(\pi, s')
    \Big\}
    \label{equ:gain2disrew_grad_2} \\
& = \underbrace{
    \sum_{s \in \setname{S}} \sum_{a \in \setname{A}}
    p_\pi^\star(s) \pi(a|s) \Big[ q_\gamma(\pi, s, a) \nabla \log \pi(a|s) \Big]
}_\text{involving the \emph{discounted} state-action value $q_\gamma$}
    + \underbrace{
        (1 - \gamma) \sum_{s \in \setname{S}} p_\pi^\star(s)
        \Big[ v_\gamma(\pi, s) \nabla \log p_\pi^\star(s) \Big]
        }_\text{involving the \emph{discounted} state value $v_\gamma$},
    \label{equ:gain2disrew_grad}
\end{align}
for all $\gamma \in [0, 1)$.
Notice that the right hand sides (RHS's) of \eqref{equ:gain2disrew_grad_2} and \eqref{equ:gain2disrew_grad}
involve the discounted-reward value functions, \ie the state value function
$v_\gamma(\pi, s), \forall s \in \setname{S}$ in \eqref{equ:v_disc}
and the corresponding (state-)action value function
$q_\gamma(\pi, s, a), \forall (s, a) \in \setname{S} \times \setname{A}$.
They are related via $v_\gamma(\pi, s) = \E{A \sim \pi}{q_\gamma(\pi, s, A)}$.
The identity \eqref{equ:gain2disrew_grad_2} was shown by \citet[\thm{2}]{baxter_2001_ihpge},
whereas \eqref{equ:gain2disrew_grad} was derived from \eqref{equ:gain2disrew_2}
by \citet[\app{A}]{morimura_2010_sdpg}.

Thus for attaining average-reward optimality, one can maximize $v_\gamma$
but merely as an approximation to $v_g$ because the equality
in \eqref{equ:gain2disrew} is only in the limit,
\ie setting $\gamma$ exactly to~1 is prohibited by definition \eqref{equ:v_disc}.
The similiar limiting behaviour applies to \eqref{equ:gain2disrew_grad_2}, where
$v_\gamma$ is used to approximately compute the gain gradient $\nabla v_g$,
yielding approximately gain-optimal policies.
In particular, \citet[\lmm{3}]{jin_2021_amdp} proved that
an $\epsilon$-gain-optimal policy is equivalent to an
$\frac{\epsilon}{3(1 - \gamma)}$-discounted-optimal policy with a certain $\gamma$ value
that depends on $\epsilon$ and a parameter describing some property of the target MDP.
Here, a policy $\pi$ is said to be $\varepsilon$-$x$-optimal for any positive~$\varepsilon$
and a criterion~$x$ if its values $v_x(\pi, s) \ge v_x(\pi_x^*, s) - \varepsilon$,
for all $s \in \setname{S}$.

The aforementioned approximation to $v_g$ by $v_\gamma$ with $\gamma \to 1$
seems to justify the use of stationary state distribution $p_\pi^\star$
(instead of the (improper) discounted state distribution $p_\pi^\gamma$ in
\eqref{equ:pstar_pgamma} below) to weight the errors in recurrent states in
approximate discounted policy evaluation, see \eg \citet[\equ{4}]{dann_2014_petd}.
In fact, the identity in \eqref{equ:gain2disrew} implies the following connection,
\begin{align}
& \underbrace {\ppimat^\star \rpivec}_{\vgvec(\pi)}
    = \lim_{\gamma \to 1} (1 - \gamma)
        \underbrace{\ppimat^\gamma \rpivec}_{\vecb{v}_\gamma(\pi)},
    \quad \text{such that}\
    \ppimat^\star = \lim_{\gamma \to 1} (1 - \gamma) \ppimat^\gamma,\
    \text{hence}\
    p_\pi^\star = \lim_{\gamma \to 1}
                \underbrace{(1 - \gamma) p_\pi^\gamma}_\text{a proper distribution},
    \label{equ:pstar_pgamma} \\
& \text{where}\quad
\ppimat^\star
    = \lim_{\tmax \to \infty} \frac{1}{\tmax} \sum_{t = 0}^{\tmax - 1} \ppimat^t,
\quad \text{and} \quad
\ppimat^\gamma
    = \lim_{\tmax \to \infty} \sum_{t = 0}^{\tmax - 1} (\gamma \ppimat)^t.
    \label{equ:pstar_pgamma_def}
\end{align}
Here, the $s_0$-th row of $\ppimat^\star$ contains
the probability values of the stationary state distribution $p_\pi^\star(\cdot|s_0)$.
On the other hand, the $s_0$-th row of $\ppimat^\gamma$ contains
the values of the improper discounted state distribution~$p_\pi^\gamma(\cdot|s_0)$,
which is improper because
$\sum_{s' \in \setname{S}} p_\pi^\gamma(s'|s_0) = 1/ (1 -\gamma) \ne 1$.
Importantly, this connection suggests that $p_\pi^\star$ is suitable as
weights in the discounted value approximator's error function only when $\gamma \to 1$.
Otherwise, $ p_\pi^\gamma$ may be more suitable.

It is also an approximation in \eqref{equ:gain2disrew_2} whenever $v_\gamma$ is
weighted by some initial state distribution~$\isd$ (such as in \eqref{equ:dispg_obj})
or \emph{transient} state-distributions $p_\pi^t$,
which generally differs from~$p_\pi^\star$.
In \eqref{equ:gain2disrew_grad}, the second RHS term is typically ignored
since calculating $\nabla \log p_\pi^\star(s)$ is difficult in RL settings.\footnote{
    The difficulty of computing the gradient of state distributions,
    such as  $\nabla \log p_\pi^\star(s)$, in RL motivates
    the development of the policy gradient theorem \citep[\ch{13.2}]{sutton_2018_irl}.
}
Consequently, $\nabla v_g(\pi)$ is approximated
(closely whenever $\gamma$ is close to 1) by the first RHS term
of \eqref{equ:gain2disrew_grad}, then by sampling the state $S \sim p_\pi^\star$
(after the agent interacts long ``enough'' with its environment).

Moreover, approximately maximizing the average reward via discounting is favourable
because discounting formulation has several mathematical virtues,
as described in \secref{sec:discrew_mathplus}.

%% file: disrew_bwoptim.tex
\subsection{A technique for attaining the most selective optimality
    with $\gamma \in [\gammabw, 1)$}
\label{sec:disrew_bwoptim}

As discussed in the previous \secref{sec:avgrewapprox},
$\gamma$-discounted optimality approximates the gain optimality as $\gamma \to 1$.
This is desirable since the gain optimality is the most selective in recurrent MDPs.
In unichain MDPs however, the gain optimality is generally underselective since
the gain ignores the rewards earned in transient states
(for an illustrative example, see
\ifpaper
\figref{fig:unichain_zrat_n1}).
\fi
Consequently, multiple gain-optimal policies prescribe different action selections
(earning different rewards) in transient states.
The underselectiveness of gain optimality (equivalent to $(n=-1)$-discount optimality)
can be refined up to the most selective optimality by increasing the value of $n$
from $-1$ to~$0$ (or higher if needed up to $n = (\setsize{S} -2)$ for unichain MDPs)
in the family of $n$-discount optimality (\secref{sec:optcrit}).

Interestingly, such a remedy towards the most selective criterion
can also be achieved by specifying a discount factor $\gamma$ that
lies in the Blackwell's interval, \ie $\gamma \in [\gammabw, 1)$,
for the $\gamma$-discounted optimality~\eqref{equ:bw_optim}.
This is because the resulting $\pi_{\gamma \in [\gammabw, 1)}^*$, which is also
called a Blackwell optimal policy, is also optimal for all $n= -1, 0, \ldots$
in $n$-discount optimality \citep[\thm{10.1.5}]{puterman_1994_mdp}.
Moreover, Blackwell optimality is always the most selective regardless of
the MDP classification, inheriting the classification invariance property of
$\gamma$-discounted optimality.
Thus, artificial discounting can be interpreted as a technique to attain
the most selective criterion (\ie the Blackwell optimality)
whenever $\gamma \in [\gammabw, 1)$ not only in recurrent but also unichain MDPs,
as well as the most general multichain MDPs.

\input{fig/zrat_n1.tex}

Targetting the Blackwell optimality (instead of gain optimality) is imperative,
especially for episodic environments\footnote{\label{fnote:episodic_env}
    Episodic environments are those with at least one terminal state.
    Once the agent enters the terminal state, the agent-environment interaction terminates.
}
that are commonly modelled as infinite-horizon MDPs
(so that the stationary policy set~$\piset{S}$ is
a sufficient space to look at for optimal policies).\footnote{
    In finite-horizon MDPs, the optimal policy is \emph{generally} non-stationary,
    where it depends on the number of remaining timesteps (in addition to the state).
    Intuitively, if we (approximately) had only 1~week to live, would we act
    the same way as if we (approximately) had 10~years to live?
    In basketball, a long shot attempt (from beyond half court) is optimal
    only at the final seconds of the game.
}
Such modelling is carried out by augmenting the state set with
a 0-reward absorbing terminal state (denoted by~$\szrat$),
as shown in \figref{gridnav2_szrat}.
This yields a unichain MDP with a non-empty set of transient states.
For such $\szrat$-models, the gain is trivially $0$ for all stationary policies
so that gain optimality is underselective.
The $(n=0)$-discount optimality improves the selectivity.
It may be the most selective (hence, it is equivalent to Blackwell optimality)
in some cases.
Otherwise, it is underselective as well, hence
some higher $n$-discount optimality criterion should be used,
\eg $n=1$ for an MDP shown in \figref{fig:unichain_zrat_n1}.
It is also worth noting that in $\szrat$-models,
($n=0$)-discount optimality is equivalent to the total reward optimality
whose $v_{\mathrm{tot}}$ \eqref{equ:v_total} is finite
\citep[\prop{10.4.2}]{puterman_1994_mdp}.
The total reward optimality therefore may also be underselective in some cases.

Towards obtaining Blackwell optimal policies in unichain MDPs, the relationship
between maximizing $\gamma$-discounted and $n$-discount criteria can be summarized
as follows (see also \figref{fig:gamma_n_venn}).
\begin{align}
& \underbrace{
    \argmax_{\pi \in \piset{S}} v_{\gamma \in [\gammabw, 1)}(\pi, s)
}_\text{Blackwell-optimal policies} \notag \\
& = \begin{cases}
\underset{\pi \in \piset{S}}{\argmax}\
    \Big[ \underbrace{
        v_{-1}(\pi, s)
            = \underset{\gamma \to 1}{\lim}\ (1 - \gamma) v_\gamma(\pi, s)
    }_\text{Based on \eqref{equ:gain2disrew}}  \Big]
    \quad \text{if MDPs are recurrent (see \secref{sec:avgrewapprox})},\\
\underset{\pi \in \piset{n=-1}^*}{\argmax}\
    \Big[\underbrace{
        v_0(\pi, s) =
            \underset{\gamma \to 1}{\lim}\ \frac{(1 - \gamma) v_\gamma(\pi, s)\
            -\ v_{-1}(\pi, s)}{1 - \gamma}
    }_\text{Based on \eqref{equ:laurent_expansion_truncated}}
    \Big]\ \text{\parbox[t]{33mm}{if MDPs are unichain and $(n=0)$-discount
        is the most selective,}} \\
\underset{\pi \in \piset{n -1}^*}{\argmax}\
    \big[ v_{n}(\pi, s) \big]\ \text{for $n = 1, 2, \ldots, (\setsize{S} - 3)$}
    \qquad \text{\parbox[t]{43mm}{if MDPs are unichain and $n$-discount
        is the most selective,}} \\
\underset{\pi \in \piset{n = (\setsize{S} - 3)}^*}{\argmax}\
    \big[ v_{n = (\setsize{S} - 2)}(\pi, s) \big]
    \quad \text{if MDPs are unichain},
\end{cases}
\label{equ:bwoptim_gamma_ndiscount}
\end{align}
for all $s \in \setname{S}$, where
$\piset{n}^*$ denotes the $n$-discount optimal policy set for $n=-1, 0, \ldots$.

From the second case in the RHS of \eqref{equ:bwoptim_gamma_ndiscount},
we know that $v_0$ can be computed by taking the limit of
a function involving $v_\gamma$ and $v_{-1} (= v_g)$ as $\gamma$ approaches~1.
This means that for unichain MDPs, the Blackwell optimal policies may be obtained
by setting $\gamma$ very close to~1,
similar to that for recurrent MDPs (the first case) but not the same since
the function of which the limit is taken differs.

In practice, the limits in the first and second cases in \eqref{equ:bwoptim_gamma_ndiscount}
are computed approximately using a discount factor $\gamma$ close to unity,
which is likely in the Blackwell's interval, \ie ${\gammabw \le (\gamma \approx 1) < 1}$.
Paradoxically however, this does not necessarily attain the Blackwell optimality
because the finite-precision computation involving $\gamma \approx 1$ yields
quite accurate estimation to the limit values:
maximizing the first and second cases in the RHS of \eqref{equ:bwoptim_gamma_ndiscount}
with a high $\gamma \approx 1$ attains (approximately) gain ($n=-1$) and bias ($n=0$) optimality
respectively, which may be underselective in unichain MDPs.
Consequently in practice, the most selective Blackwell optimality can always be
achieved using $\gamma$ that is at least as high as $\gammabw$ but not too close to 1.
That is, $\gammabw \le \gamma \le \bar{\gamma} < 1$ for
some practical upper bound $\bar{\gamma}$, which is
computation (hence, implementation) dependent.

%% file: fig/zrat_n1.tex
\begin{figure}[t]
\centering

\resizebox{0.75\textwidth}{!}{
\begin{tikzpicture}[
>={Latex[length=5mm]}, %
node distance = 5cm and 5cm, on grid,
-{Latex[length=5mm]}, %
semithick, %
strans/.style={circle, fill=white,
draw=black, text=black, minimum width = 1.5cm},
srecur/.style={circle, fill=white,
draw=black, solid, text=black, minimum width = 1.5cm},
]
\node[strans](A)[] {\Large $s^0$};
\node[srecur](B)[right=of A] {\Large $s^1$};
\node[srecur](C)[right=of B] {\Large $s^2$};

\path (A) edge [bend left, blue] node[above] {$reward = 2$} (C);
\path (A) edge [bend right, red] node[below] {$reward = 1$} (B);
\path (B) edge [bend right, halfgreen] node[below] {$reward = 1$} (C);
\path (C) edge [out=-30, in=30, loop, black] node[right] {$reward= 0$} (C);

\end{tikzpicture}
} %

\caption{A symbolic diagram of a unichain MDP with
$\setname{S} = \{s^0, s^1, s^2 \}$ and deterministic transitions.
There are two stationary policies that differ in their blue or red action selection
in the left-most state $s^0$, hence we call them: red and blue policies.
Under both policies, the right-most state $s^2$ is
a 0-reward (absorbing) recurrent state $\szrat$,
hence this unichain MDP is a $\szrat$-model.
Both policies are gain-optimal and $(n=0)$-discount optimal,
but only the blue policy is $(n=1)$-discount optimal.
Therefore, $(n=1 = \setsize{S}-2)$-discount optimality is the most selective,
which is equivalent to Blackwell optimality.
In particular for this simple environment, the Blackwell discount factor is
trivially $\gammabw=0$.
Note that the total reward and the ($n=0$)-discount values of both policies are equal,
\ie $v_{\mathrm{tot}}(\mathrm{red}, s^0) = v_{\mathrm{tot}}(\mathrm{blue}, s^0)
= v_{n=0}(\mathrm{red}, s^0) = v_{n=0}(\mathrm{blue}, s^0) = 2$
such that the total reward and ($n=0$)-discount criteria are underselective.
This example MDP is adopted from \citet[\fig{10.1.1}]{puterman_1994_mdp} and
\citet[\fig{1}]{mahadevan_1996_sensitivedo}.
}
\label{fig:unichain_zrat_n1}
\end{figure}
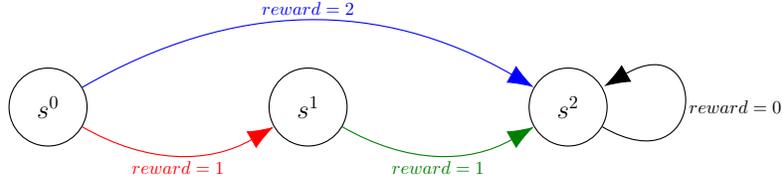

%% file: disrew_mathplus.tex
\subsection{Contraction, variance reduction, and independence from MDP classification}
\label{sec:discrew_mathplus}

Discounted reward optimality is easier to deal with than
its average reward counterpart.
This can be attributed to three main factors as follows.

\first the discounted-reward theory holds regardless of the classification of
the induced MCs (\secref{sec:mdpclf}), whereas that of the average reward
involves such classification.
Because in RL settings, the transition probability $p(s'|s, a)$ is unknown,
average reward algorithms require estimation or assumption about
the chain classification, specifically whether unichain or multichain.
Nevertheless, note that such (assumed) classification is needed in order to apply
a specific (simpler) class of average-reward algorithms:
leveraging the fact that a unichain MDP has a single scalar gain
(associated with its single chain) that is constant across all states,
whereas a multichain MDP generally has different gain values
associated with its multiple chains.

\second the discounted-reward Bellman optimality operator $\mathbb{B}_\gamma^*$
is contractive, where the discount factor $\gamma$ serves as the contraction modulus
(hence, $\mathbb{B}_\gamma^*$ is a $\gamma$-contraction).
That is,
\begin{equation*} %
\norm{\mathbb{B}_\gamma^*[\vecb{v}] - \mathbb{B}_\gamma^*[\vecb{v}']}_\infty
\le \gamma \norm{\vecb{v} - \vecb{v}'}_\infty,
\quad \text{for any vectors $\vecb{v},\vecb{v}' \in \real{\setsize{S}}$},
\end{equation*}
where in state-wise form, $\mathbb{B}_\gamma^*[\vecb{v}](s) \eqdef
\max_{a \in \setname{A}} \{ r(s, a) + \gamma
\sum_{s' \in \setname{S}} p(s'|s, a) v(s') \}, \forall s \in \setname{S}$,
and $\norm{\vecb{v}}_\infty \eqdef \max_{s \in \setname{S}} |v(s)|$
denotes the maximum norm.
This means that $\mathbb{B}_\gamma^*$ makes $\vecb{v}$ and $\vecb{v}'$
closer by at least $\gamma$ such that the sequence of iterates
$\vecb{v}^{k+1} \gets \mathbb{B}_\gamma^*[\vecb{v}^k]$ converges
to the unique fixed point of $\mathbb{B}_\gamma^*$ as $k \to \infty$
from any initial $\vecb{v}^{k=0} \in \real{\setsize{S}}$.
This is based on the Banach's fixed-point theorem \citet[\thm{A.9}]{szepesvari_2010_arl}.
In particular as $k \to \infty$, we obtain
$\mathbb{B}_\gamma^*[\vecb{v}^k] = \vecb{v}^k = \vecb{v}_\gamma^*$,
where $\vecb{v}_\gamma^*$ denotes the optimal discounted value,
\ie $\vecb{v}_\gamma(\pi_\gamma^*)$.
Thus, $\lim_{k \to \infty} \norm{\vecb{v}^k - \vecb{v}_\gamma^*}_\infty = \vecb{0}$.
In the absense of $\gamma$, the contraction no longer holds.
This is the case for the average-reward Bellman optimality operator
$\mathbb{B}_g^*[\vecb{v}](s) \eqdef
\max_{a \in \setname{A}} \{ r(s, a) +
\sum_{s' \in \setname{S}} p(s'|s, a) v(s') \}, \forall s \in \setname{S}$.
As a result, the basic value iteration based on $\mathbb{B}_g^*$ is not guaranteed
to converge \citep[\secc{2.4.3}]{mahadevan_1996_avgrew}.

The aforementioned contraction property also applies to
the discounted-reward Bellman \emph{expectation} operator, \ie
$\mathbb{B}_\gamma^\pi[\vecb{v}](s) \eqdef
\E{A \sim \pi}{ r(s, A) + \gamma
\sum_{s' \in \setname{S}} p(s'|s, A) v(s')}, \forall s \in \setname{S}$.
As a consequence of Banach’s fixed-point theorem, we have
${\lim_{k \to \infty} \norm{\vecb{v}^k - \vecb{v}_\gamma^\pi}_\infty} = \vecb{0}$,
where $\vecb{v}^{k+1} \gets \mathbb{B}_\gamma^\pi[\vecb{v}^k]$ is iteratively applied
on any initial $\vecb{v}^{k=0} \in \real{\setsize{S}}$ for $k=0, 1, \ldots$,
whereas $\vecb{v}_\gamma^\pi \eqdef \vecb{v}_\gamma(\pi)$ is
the discounted policy value of a policy $\pi$.
In other words, $\vecb{v}_\gamma^\pi$ is the unique fixed point of
$\mathbb{B}_\gamma^\pi$ such that
$\vecb{v}_\gamma^\pi = \mathbb{B}_\gamma^\pi[\vecb{v}_\gamma^\pi]$,
which is known as the discounted-reward Bellman evaluation equation ($\gamma$-BEE).
This $\gamma$-BEE is further utilized to derive a TD-based parametric value approximator
whose convergence depends on the fact that $\gamma \in [0, 1)$.
That is, the approximator's error minimizer formula involves
the inverse $(\mat{I} - \gamma \ppimat)^{-1}$, which exists \citep[p206]{sutton_2018_irl}.
This is in contrast to the matrix $(\mat{I} - \ppimat)$ whose
inverse does not exist \citep[\page{596}]{puterman_1994_mdp}.

In addition, the contractive nature induced by $\gamma$ is also utilized for
computing state similarity metrics \citep{castro_2020_sim, ferns_2006_met}.
There, $\gamma$ guarantees the convergence of the metric operator to its fixed point
(when such an operator is iteratively applied).
Note that the notion of state similarity plays an important role in for example,
state aggregation and state representation learning.

\third discounting can be used to reduce the variance of, for example
policy gradient estimates, at the cost of bias-errors
\citetext{\citealp[\thm{3}]{baxter_2001_ihpge}; \citealp[\thm{1}]{kakade_2001_avgrew}}.
In particular, the variance (\emph{the bias-error}) increases (\emph{decreases})
as a function of $1/(1 - \gamma) = \sum_{t=0}^\infty \gamma^t$.
This is because the effective number of timesteps (horizon) can be controlled
by~$\gamma$.
This is also related to the fact that the infinite-horizon discounted reward
$v_\gamma$ \eqref{equ:v_disc} can be $\epsilon$-approximated by
a finite horizon $\tau$ proportional to $\log_\gamma (1 - \gamma)$,
as noted by \citet[\fnote{1}]{tang_2010_islr}.
That is,
\begin{equation*}
\epsilon
\eqdef \E{}{\sum_{t = 0}^\infty \gamma^t r(S_t, A_t)}
    - \E{}{\sum_{t=0}^{\tau - 1} \gamma^t r(S_t, A_t)}
= \E{}{\sum_{t = \tau}^\infty \gamma^t r(S_t, A_t)}
\le \sum_{t = \tau}^\infty \gamma^t \rmax
    = \frac{\gamma^{\tau} \rmax}{1 - \gamma}.
\end{equation*}
This is then re-arranged to obtain $\gamma^{\tau} \ge ((1 - \gamma) \epsilon) / \rmax$,
whose both sides are taken to the logarithm with base $\gamma$ to yield
\begin{equation*}
\tau \ge \log_\gamma \frac{(1 - \gamma) \epsilon}{\rmax},
\quad \text{hence, the smallest of such a finite horizon is}\
\tau = \Big\lceil \log_\gamma \frac{(1 - \gamma) \epsilon}{\rmax} \Big\rceil,
\end{equation*}
where $\ceil*{x}$ indicates the smallest integer greater than or equal to $x \in \real{}$.

%% file: gamma.tex
\section{Artificial discount factors are sensitive and troublesome}
\label{sec:gamma}

The artificial discount factor $\gamma$ (which is part of the solution method)
is said to be \emph{sensitive} because the performance of RL methods often depends
largely on~$\gamma$.
\figref{gn25_vargamma} illustrates this phenomenon using
$Q_\gamma$-learning with various $\gamma$ values.
As can be seen, higher $\gamma$ leads to slower convergence,
whereas lower $\gamma$ leads to suboptimal policies
(with respect to the most selective criterion, which in this case,
is the gain optimality since the MDP is recurrent).
This trade-off is elaborated more in \secrefand{sec:gamma_hi}{sec:gamma_lo}.
The sensitivity to $\gamma$ has also been observed in the error
(and even whether convergence or divergence) of
approximate policy evaluation methods with function approximators
\citetext{\citealp[\fig{1}]{scherrer_2010_poleval}; \citealp[Example~11.1]{sutton_2018_irl}}.

Additionally, \figref{fig:optim_landscape} shows the effect of
various $\gamma$ values on the optimization landscapes on which
policy gradient methods look for the maximizer of the discounted value $v_\gamma$.
It is evident that different $\gamma$ values induce different
maximizing policy parameters.
From the 2D visualization in \figref{fig:optim_landscape}, we can observe that
the landscape of $v_\gamma$ begins to look like that of the gain $v_g$ when
$\gamma$ is around $\gammabw$, then becomes more and more look like it as
$\gamma$ approaches 1.
When $\gamma$ is far less than $\gammabw$, the maximizer of $v_\gamma$
does not coincide with that of $v_g$.
In such cases, some local maximizer of $v_\gamma$ may be desirable
(instead of the global one) because of its proximity to the maximizer of $v_g$,
which represents the optimal point of the most selective criterion
for recurrent MDPs examined in \figref{fig:optim_landscape}.

The artificial $\gamma$ is \emph{troublesome} because its critical value,
\ie $\gammabw$, is difficult to determine, even in DP where
the transition and reward functions are known \citep{hordijk_1985_disc}.
This is exacerbated by the fact that $\gammabw$ is specific to
each environment instance (even from the same environment family,
as shown in \figref{gammabw_envspecific}).
Nevertheless, knowing this critical value $\gammabw$ is always desirable.
For example, despite the gain optimality can be attained by having $\gamma$ very close to~1,
setting $\gamma \gets \gammabw$ (or some value around it)
leads to not only convergence to the optimal gain (or close to it)
but also faster convergence, as demonstrated by $Q_\gamma$-learning (\figref{gn25_vargamma}).
We can also observe visually in \figref{fig:optim_landscape} that
the discounted $v_{\gamma \approx \gammabw}$-landscape already resembles
the gain $v_g$-landscape.
Thus, for obtaining the gain-optimal policy in recurrent MDPs, $\gamma$ does not
need to be too close to 1 (as long as it is larger than or equal to $\gammabw$);
see also \eqref{equ:bwoptim_gamma_ndiscount}.

Apart from that, $\gamma$ is troublesome because some derivation involving it
demands extra care, \eg for handling the improper discounted state distribution
$p_\pi^\gamma$ \eqref{equ:pstar_pgamma} in discounted-reward policy gradient algorithms
\citep{nota_2020_pgg, thomas_2014_biasnac}.

\input{gamma_xprmt.tex}
\input{gamma_hi.tex}

\input{gamma_lo.tex}
\input{gamma_landscape.tex}

%% file: gamma_xprmt.tex
\begin{figure*}
\centering

\begin{subfigure}{0.45\textwidth}
\includegraphics[width=\textwidth]{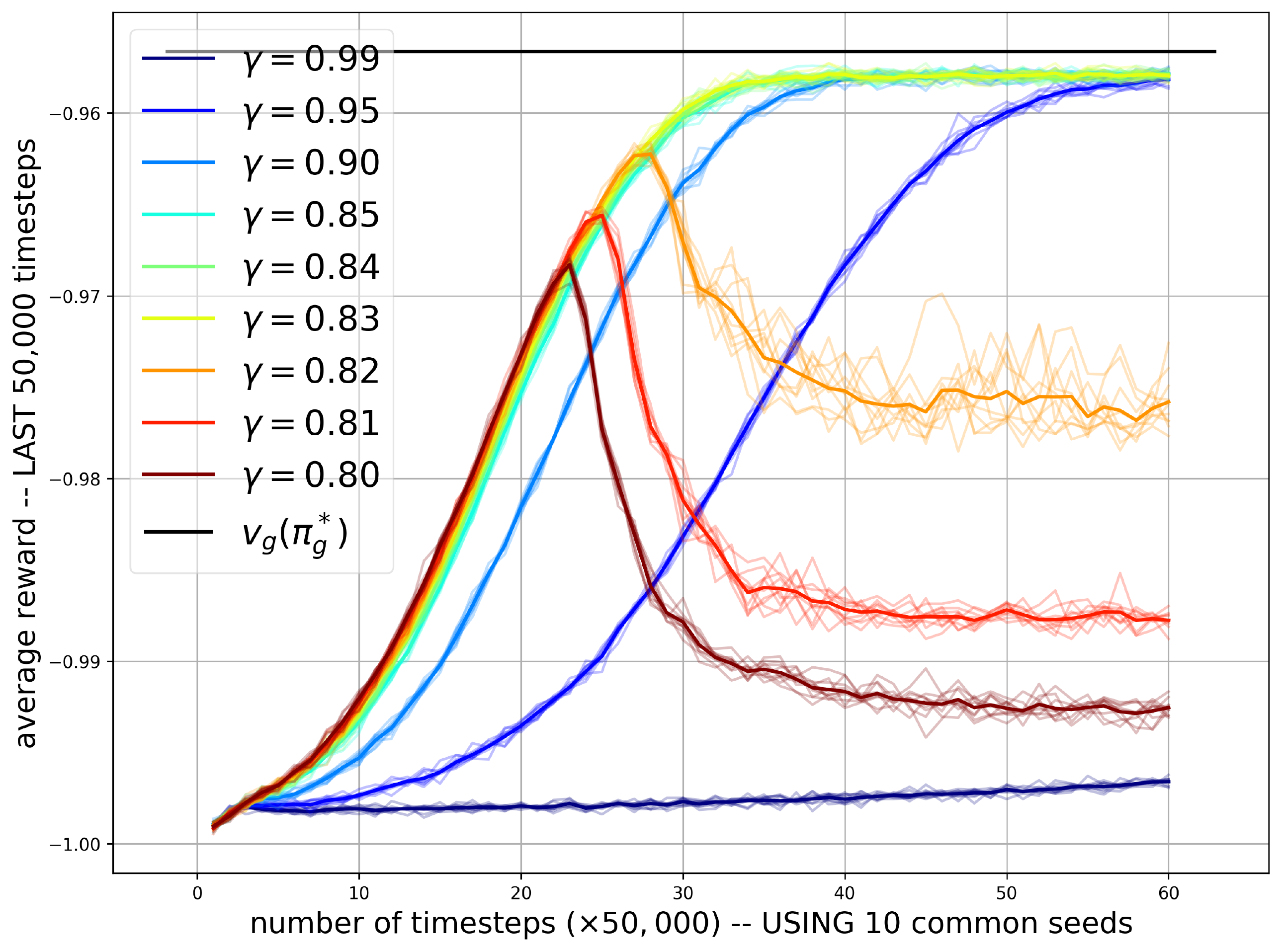}
\subcaption{Learning curves of $Q_\gamma$-learning with varying $\gamma$ values
on GridNav-25, which is episodic but modelled using $\sreset$ (\secref{sec:ep_repeat}).
Empirically, the critical discount factor is $\gammabw \approx 0.83$ (yellowish).
}
\label{gn25_vargamma}
\end{subfigure}
 \hspace{0.025\textwidth}
\begin{subfigure}{0.46\textwidth}
\includegraphics[width=\textwidth]{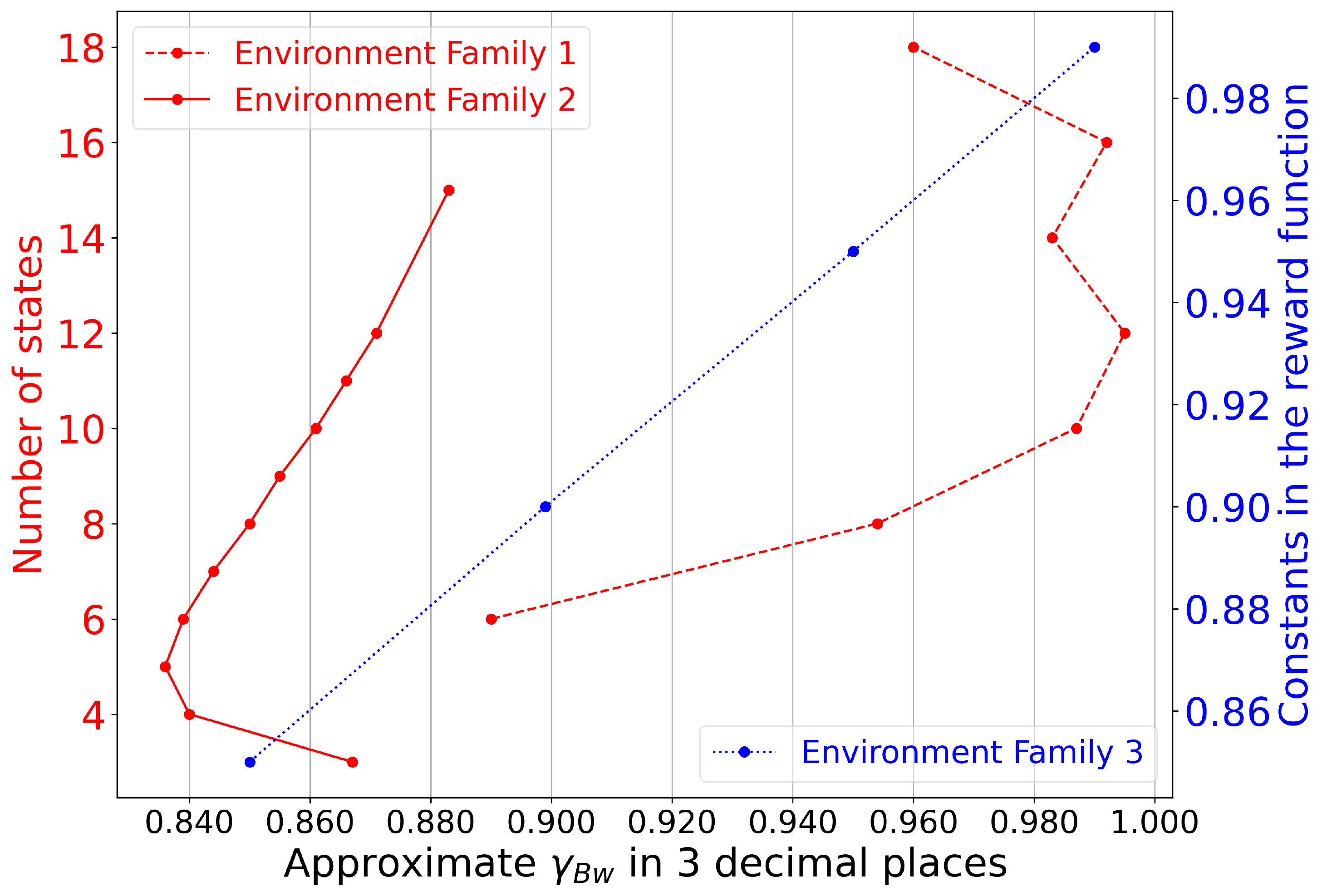}
\subcaption{The critical $\gammabw$ as a (non-trivial) function
{\color{red} of number of states}, and
{\color{blue} of some constant in the reward function}
on three environment families.}
\label{gammabw_envspecific}
\end{subfigure}

\caption{Empirical results illustrating the sensitivity and troublesomeness of
artificial discount factors $\gamma$.
In the left sub-figure~(a), the gain optimality is the most selective
(since the MDP is recurrent), where the optimal gain $v_g(\pi_g^*)$ is indicated by
the black solid horizontal line on top of the plot.
For experimental setup about the environments and learning methods,
see \secref{sec:xprmt_setup_avgdisrew}.
}
\label{fig:gamma_xprmt}
\end{figure*}

%% file: gamma_hi.tex
\subsection{Higher discount factors lead to slower convergence}
\label{sec:gamma_hi}

According to \eqref{equ:gain2disrew}, increasing the discount factor $\gamma$
closer to~1 makes the scaled discounted reward $(1 - \gamma) v_\gamma$ approximate
the average reward $v_g$ more closely.
This means that a discounted-reward method with such a setting obtains
more accurate estimates of gain-optimal policies.
However, it suffers from a lower rate of convergence (to the approximate gain optimality),
as well as from some numerical issue (since for example, it involves
the term $1/(1 - \gamma)$ that explodes as $\gamma \to 1$).
This becomes unavoidable whenever $\gammabw$ is indeed very close to unity because
the most selective Blackwell optimality (equivalent to gain optimality in recurrent MDPs)
requires that $\gamma \ge \gammabw$.

The slow convergence can be explained by examining the effect of
the effective horizon induced by~$\gamma$.
That is, as $\gamma$ approaches~1, the reward information is propagated to more states
\citep[\fig{1}]{beleznay_1999_vfe}.
From discounted policy gradient methods, we also know that an i.i.d state sample
from the discounted state distribution $p_\pi^\gamma$ in \eqref{equ:pstar_pgamma}
is the last state of a trajectory whose length is drawn from a geometric distribution
$Geo(p=1 - \gamma)$, see \citet[\alg{3}]{kumar_2020_zodpg}.
Evidently, the closer $\gamma$ to 1, the longer the required trajectory.
Also recall that such a geometric distribution has a mean of $1 / (1 - \gamma)$
and a variance of $\gamma / (1 - \gamma)^2$, which blow up as $\gamma \to 1$.

There are numerous works that prove and demonstrate slow convergence
due to higher $\gamma$.
From them, we understand that the error (hence, iteration/sample complexity)
essentially grows as a function of $1/ (1 - \gamma)$.
Those works include
\citetext{\citealp[\fig{3}]{thrun_1993_fnrl};
\citealp[\thm{1}]{melo_2007_qlearn};
\citealp[\thm{4.4}]{yang_2019_dqn}} for Q-learning with function approximators,
\citep[\equ{9.14, 12.8}]{sutton_2018_irl} for TD learning, and
\citep[\tbl{1, 2}]{agarwal_2019_polgrad} for policy gradient methods.
We note that for a specific environment type and with some additional hyperparameter,
\cite{devraj_2020_qlearn} proposed a variant of Q-learning whose sample complexity
is independent of $\gamma$.

%% file: gamma_lo.tex
\subsection{Lower discount factors likely lead to suboptimal policies}
\label{sec:gamma_lo}

Setting $\gamma$ further from~1 such that $\gamma < \gammabw$
yields $\gamma$-discounted optimal policies that are suboptimal
with respect to the most selective criterion (see \figref{gn25_vargamma}).
From the gain optimality standpoint, lower $\gamma$ makes $(1 - \gamma) v_\gamma$
deviate from $v_g$ in the order of $\bigO{1 - \gamma}$ as shown by \citep{baxter_2001_ihpge}.
More generally based on \eqref{equ:bwoptim_gamma_ndiscount},
$\gamma < \gammabw$ induces an optimal policy $\pi_{\gamma < \gammabw}^*$
that is not Blackwell optimal
(hence, $\pi_{\gamma < \gammabw}^*$ is also not gain optimal in recurrent MDPs).
This begs the question: is it ethical to run a suboptimal policy
(due to misspecifying the optimality criterion) in perpetuity?

For a parameterized policy in recurrent MDPs, $\gamma < \gammabw$ induces
a discounted $v_\gamma$-landscape that is different form the gain $v_g$-landscape.
\figref{fig:optim_landscape} shows such a discrepancy, which becomes
more significant as $\gamma$ is set further below 1.
Therefore, the maximizing parameters do not coincide, \ie
$\argmax_{\vecb{\theta}} v_{\gamma < \gammabw}(\pi(\vecb{\theta}))
\ne \argmax_{\vecb{\theta}} v_g(\pi(\vecb{\theta})) \eqdefr \vecb{\theta}_{\!g}^*$,
where $\vecb{\theta} \in \Theta$ denotes the policy parameter in some parameter set.
Interestingly, $v_{\gamma < \gammabw}(\pi(\vecb{\theta}_{\!g}^*))$ is a local maximum
in $v_\gamma$-landscape so that the $(\gamma < \gammabw)$-discounted-reward optimization
is ill-posed in that the (global) maximum is not what we desire in terms of obtaining
an optimal policy with respect to the most selective criterion
(that is, the gain optimal policy $\pi(\vecb{\theta}_{\!g}^*)$ in recurrent MDPs).

\citet[\thm{2}]{petrik_2008_ldisc} established the following error bound due to
misspecifying a discount factor $\gamma < \gammabw$.
That is,
\begin{equation*}
\norm{\vecb{v}_{\gammabw}^* - \vecb{v}_\gamma^*}_\infty
\le \frac{(\gammabw - \gamma)\ \rmax}{(1 - \gamma) (1 - \gammabw)},
\quad \text{for $\vecb{v}_{\gammabw}^*, \vecb{v}_\gamma^* \in \real{\setsize{S}}$,
and $\vecb{v}_\gamma^* \eqdef \vecb{v}_\gamma(\pi_\gamma^*)$
for any $\gamma \in [0, 1)$}.
\end{equation*}
Subsequently, \citet{jiang_2016_shp} refined the above error bound by
taking into account the transition and reward functions.
They also highlighted that such an error as a function of $\gamma$ is not
monotonically decreasing (with increasing $\gamma$).
This is consistent with what \citet{smallwood_1966_reg} observed, \ie
multiple disconnected $\gamma$-intervals that share
the same $\gamma$-discounted optimal policy.
We note that specifically for sparse-reward environments
(where non-zero rewards are not received in every timestep),
a lower discount factor $\gamma < \gammabw$ is likely to improve the performance
of RL algorithms \citep[\thm{10}]{petrik_2008_ldisc}.
They argue that lowering $\gamma$ decreases the value approximation error
$\norm{\vecb{v}_{\gamma}^* - \hat{\vecb{v}}_{\gamma}^*}_\infty$ more significantly
than it increases the $\gamma$-misspecification error
$\norm{\vecb{v}_{\gammabw}^* - \vecb{v}_{\gamma}^*}_\infty$.
Here, $\hat{\vecb{v}}_{\gamma}^*$ denotes an approximation to $\vecb{v}_{\gamma}^*$.

%% file: gamma_landscape.tex
\begin{landscape}
\begin{figure}
\centering

\begin{subfigure}{0.210\textwidth}
\includegraphics[width=\textwidth]{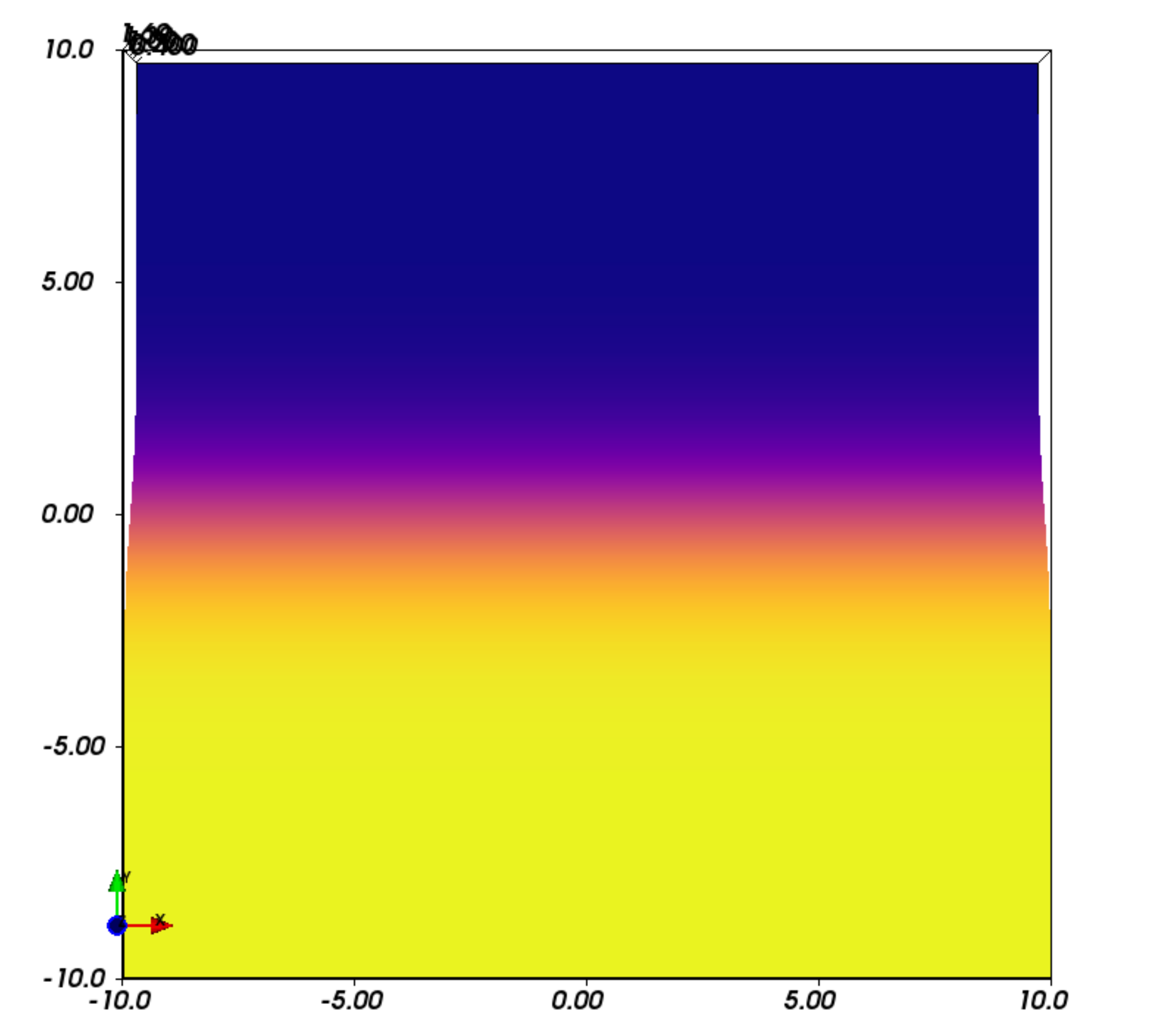}
\subcaption*{$\gamma=0.00$}
\end{subfigure}
\begin{subfigure}{0.210\textwidth}
\includegraphics[width=\textwidth]{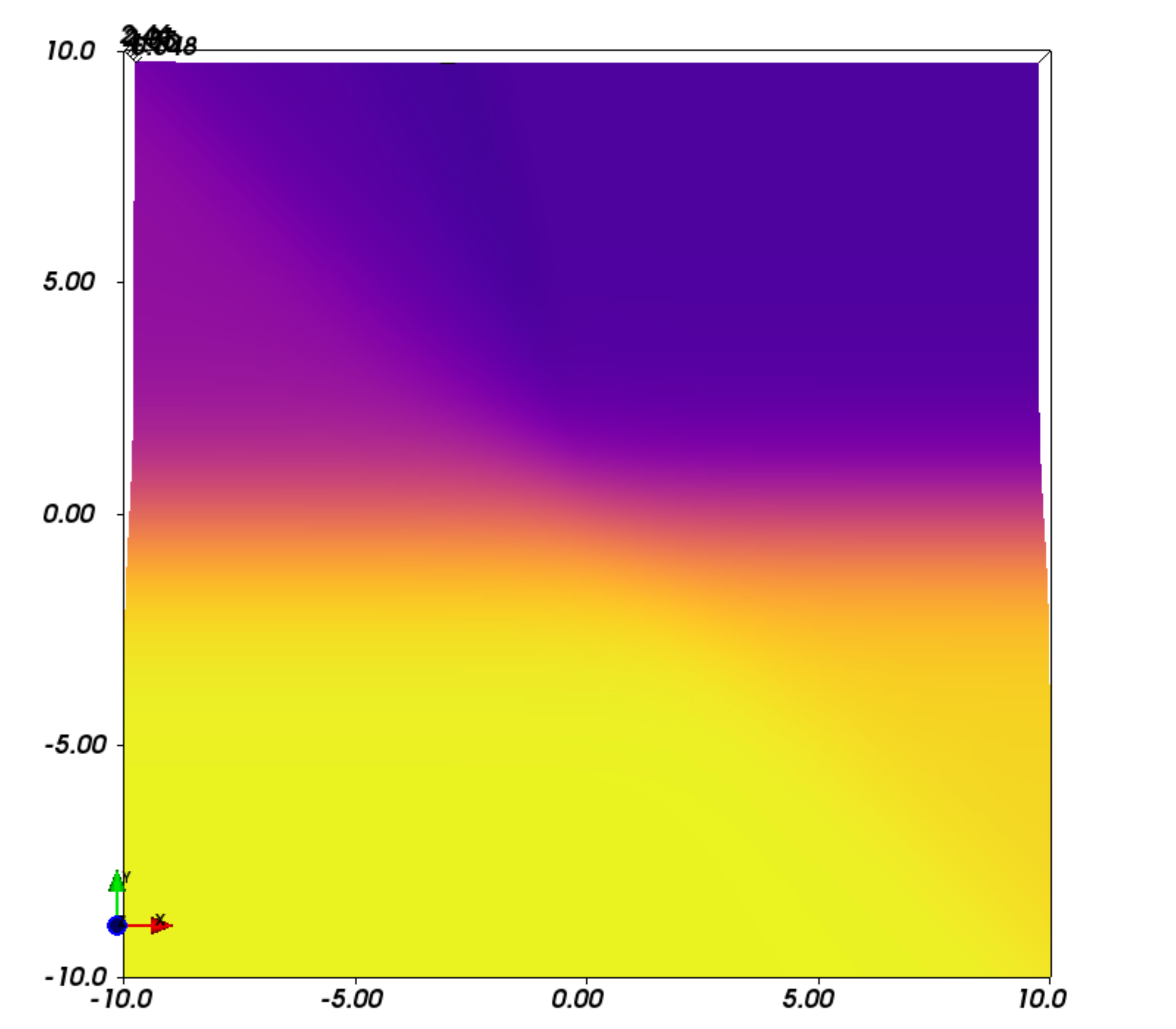}
\subcaption*{$\gamma=0.35$}
\end{subfigure}
\begin{subfigure}{0.210\textwidth}
\includegraphics[width=\textwidth]{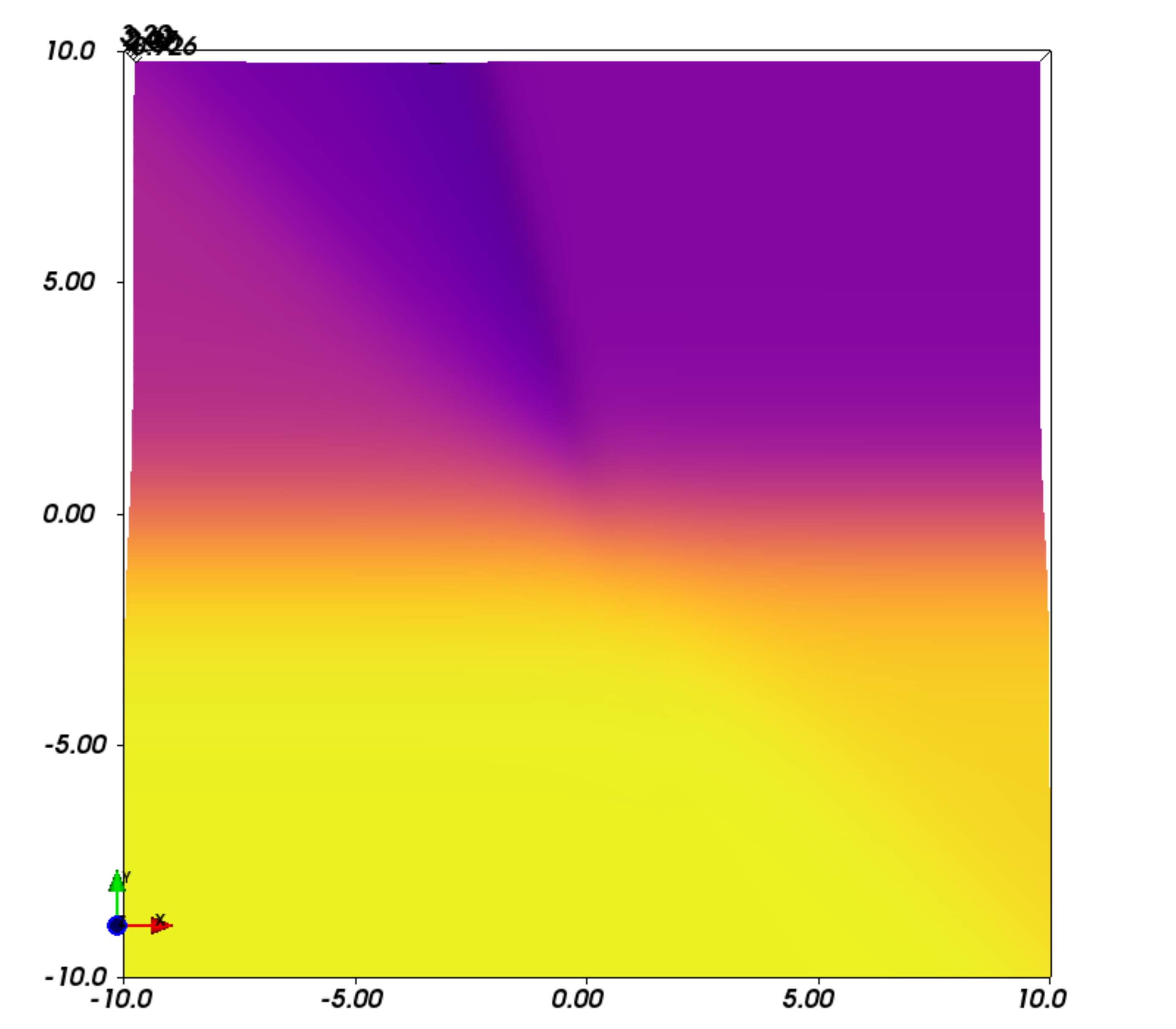}
\subcaption*{$\gamma=0.50$}
\end{subfigure}
\begin{subfigure}{0.210\textwidth}
\includegraphics[width=\textwidth]{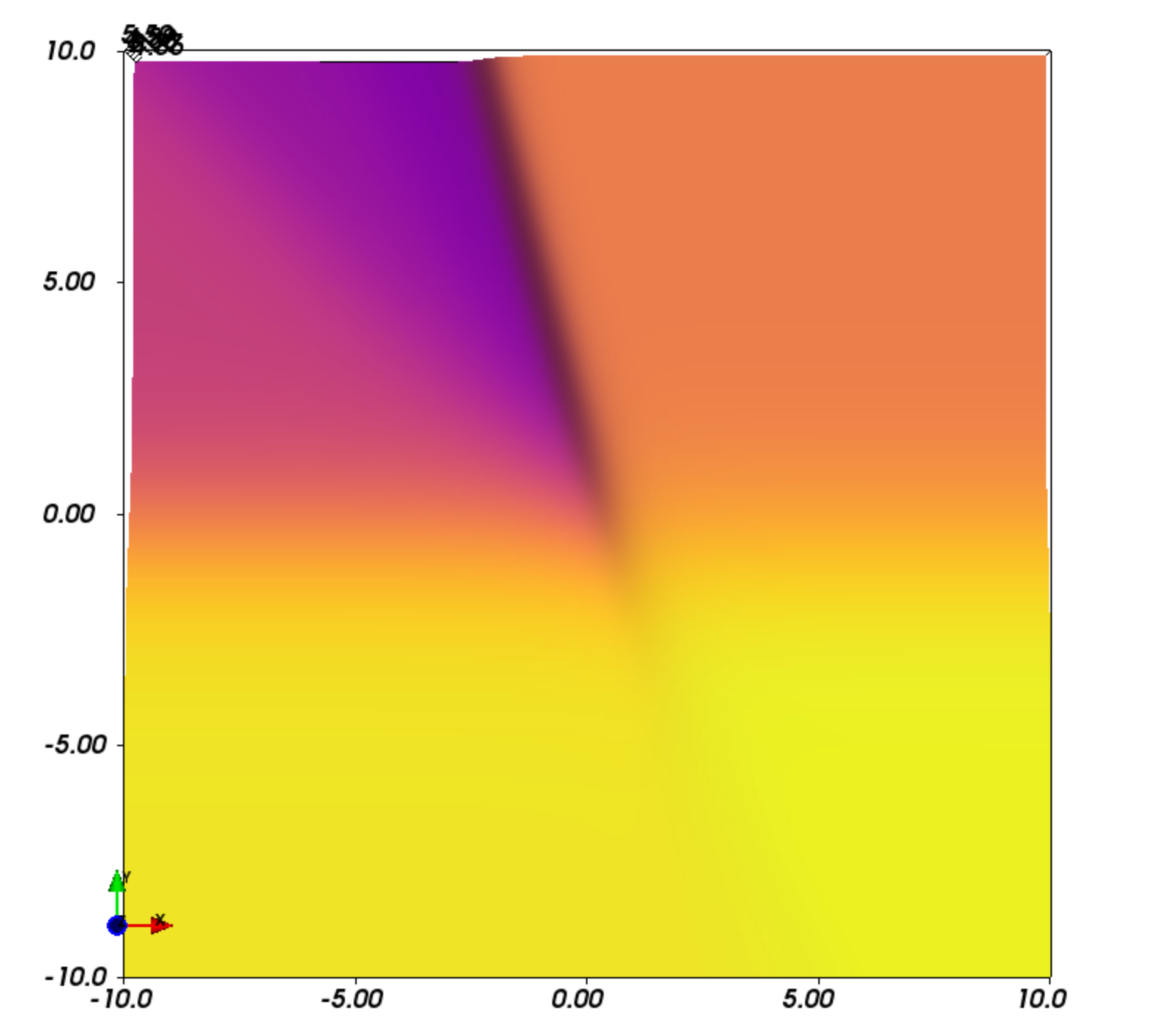}
\subcaption*{$\gamma=0.70$}
\end{subfigure}
\begin{subfigure}{0.210\textwidth}
\includegraphics[width=\textwidth]{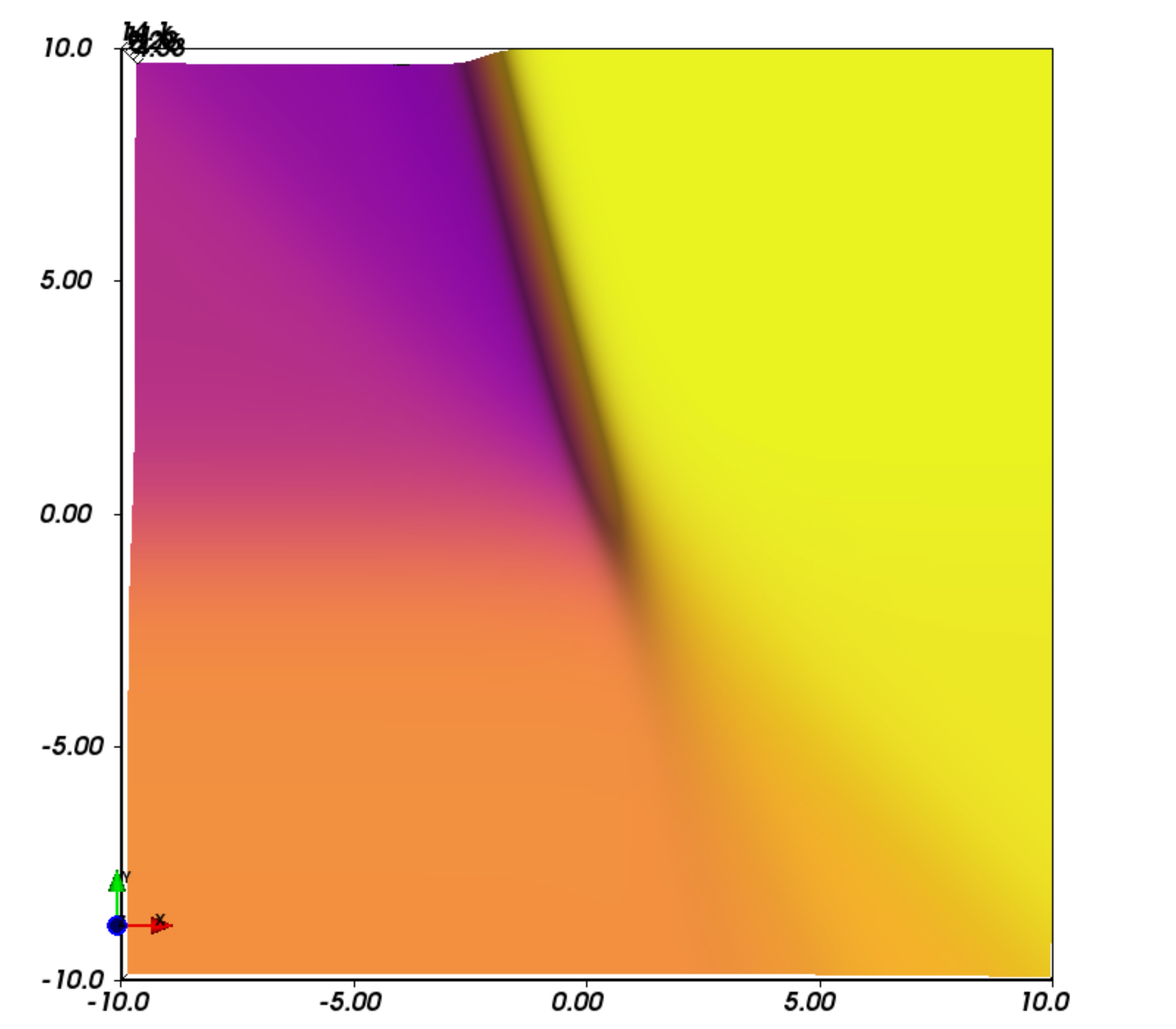}
\subcaption*{{\color{red}$\gamma=0.85 \approx \gammabw$}}
\end{subfigure}
\begin{subfigure}{0.210\textwidth}
\includegraphics[width=\textwidth]{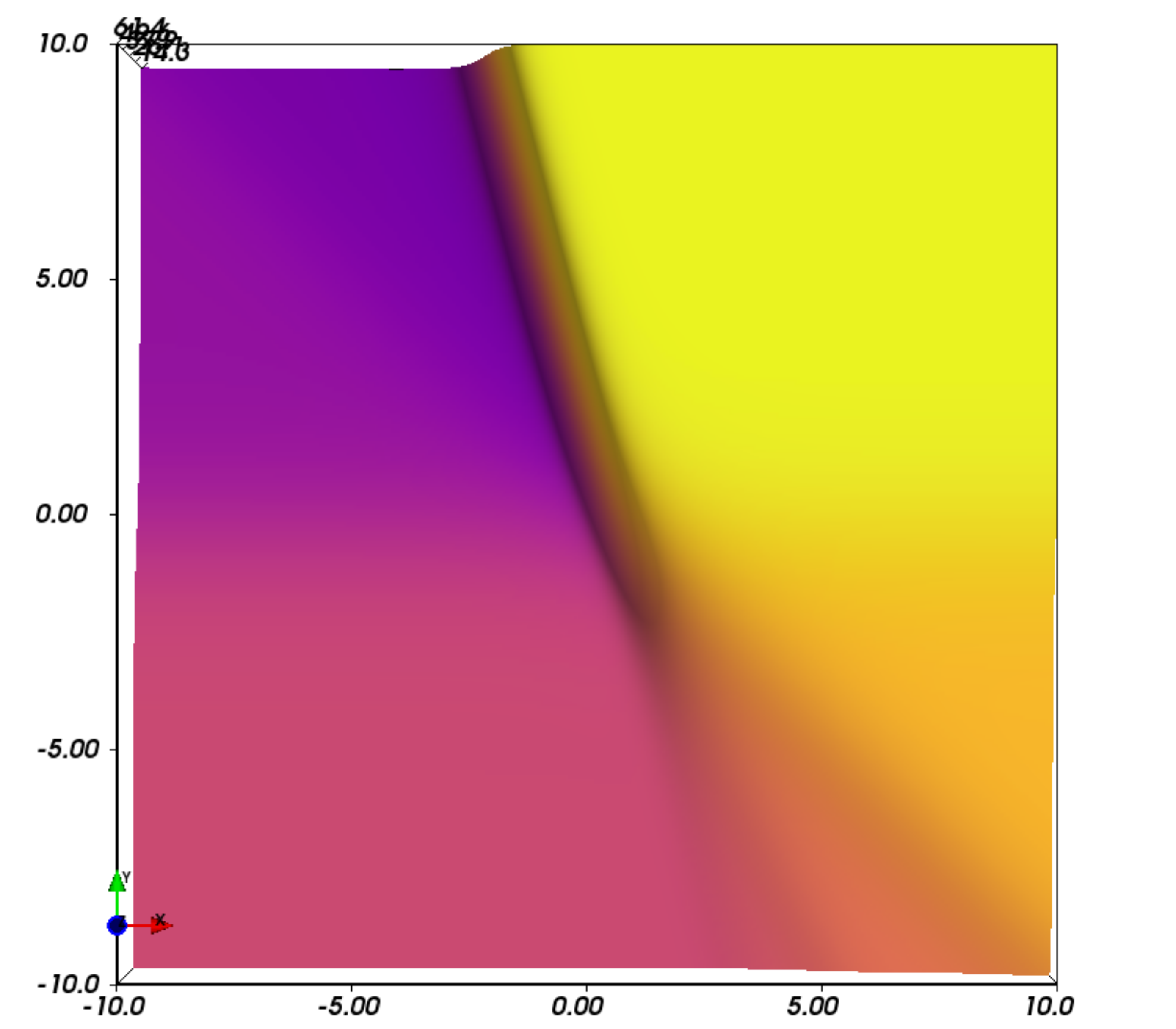}
\subcaption*{$\gamma=0.95$}
\end{subfigure}
\begin{subfigure}{0.210\textwidth}
\includegraphics[width=\textwidth]{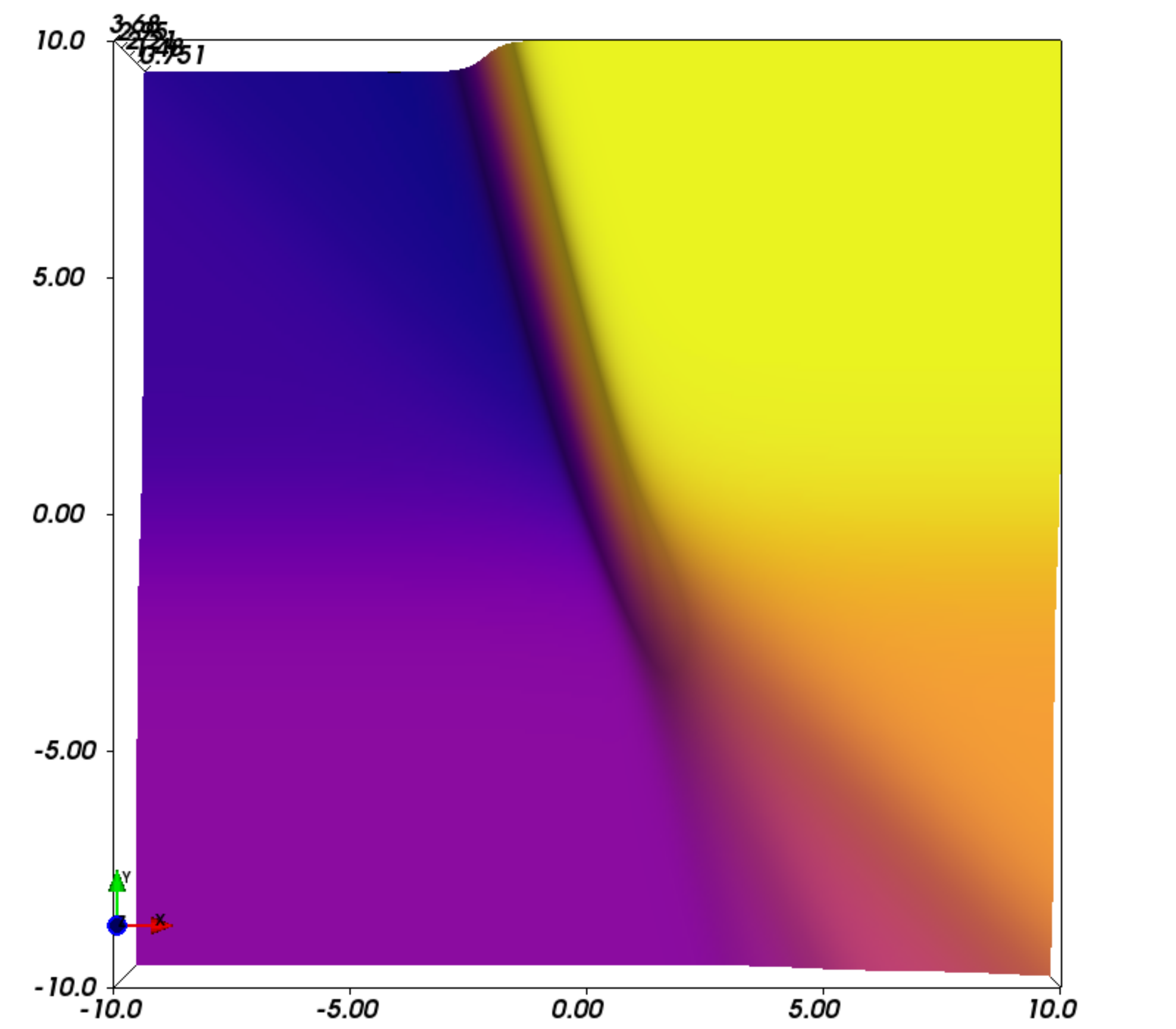}
\subcaption*{Gain}
\end{subfigure}

\begin{subfigure}{0.210\textwidth}
\includegraphics[width=\textwidth]{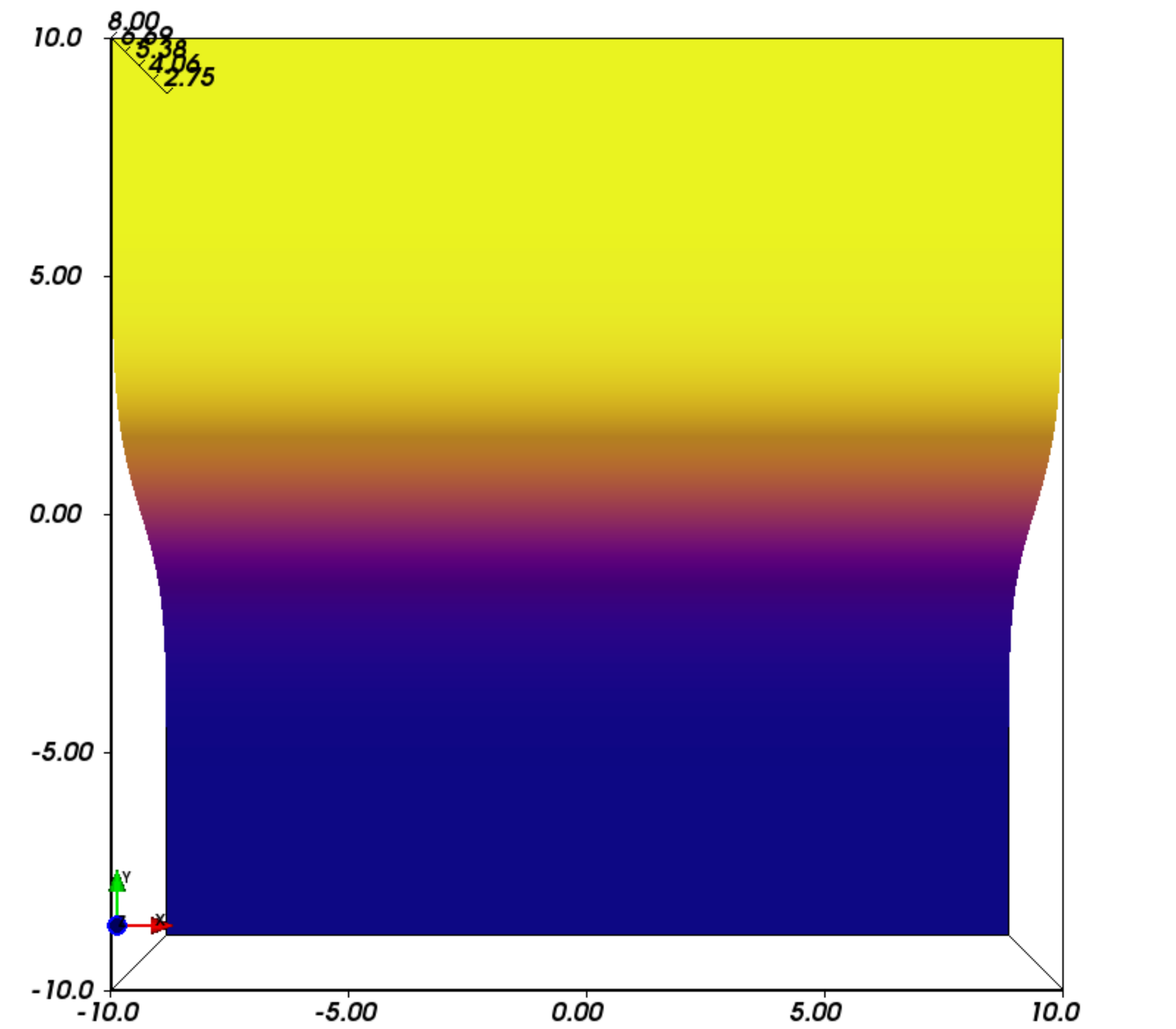}
\subcaption*{$\gamma=0.00$}
\end{subfigure}
\begin{subfigure}{0.210\textwidth}
\includegraphics[width=\textwidth]{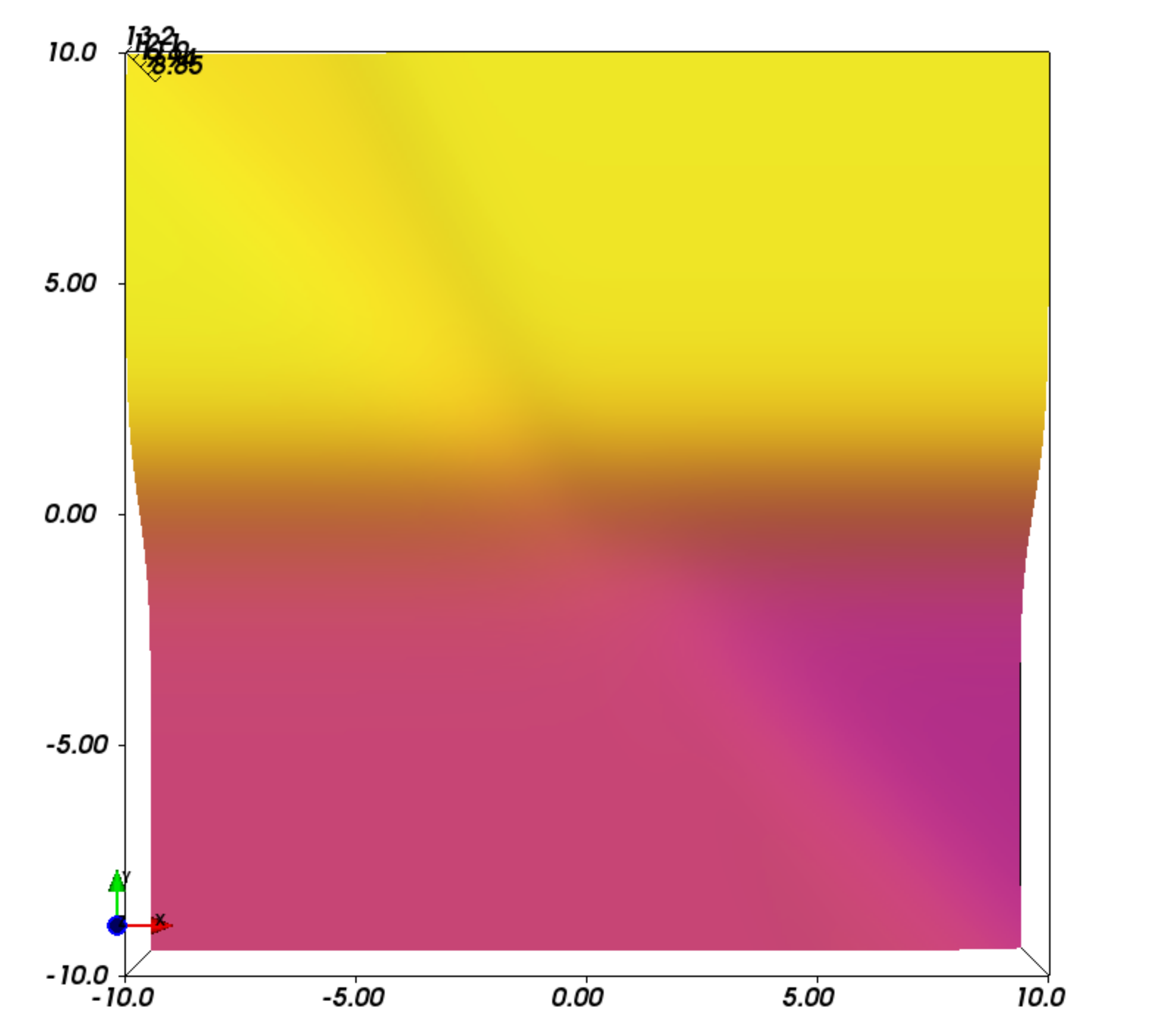}
\subcaption*{$\gamma=0.35$}
\end{subfigure}
\begin{subfigure}{0.210\textwidth}
\includegraphics[width=\textwidth]{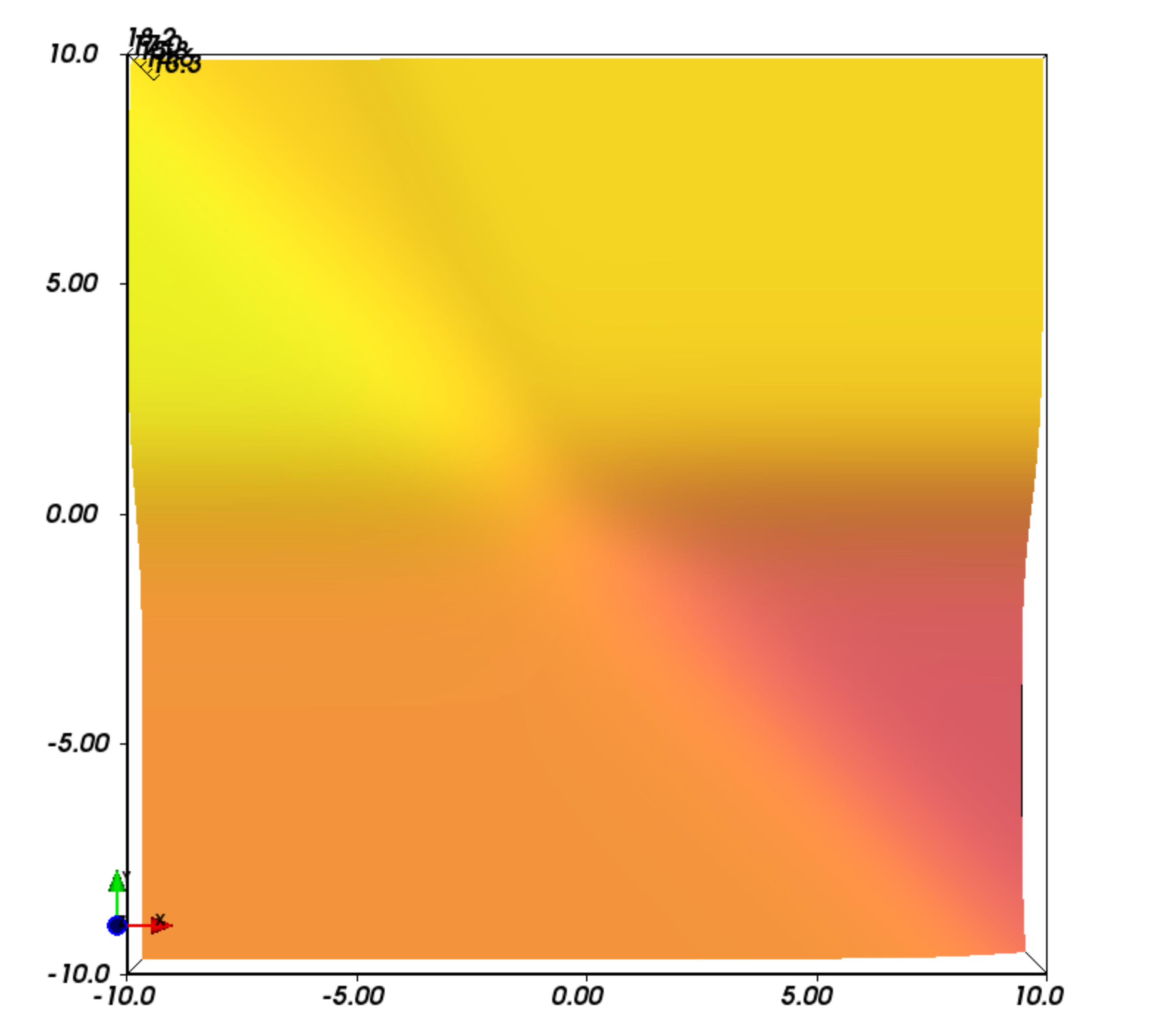}
\subcaption*{$\gamma=0.50$}
\end{subfigure}
\begin{subfigure}{0.210\textwidth}
\includegraphics[width=\textwidth]{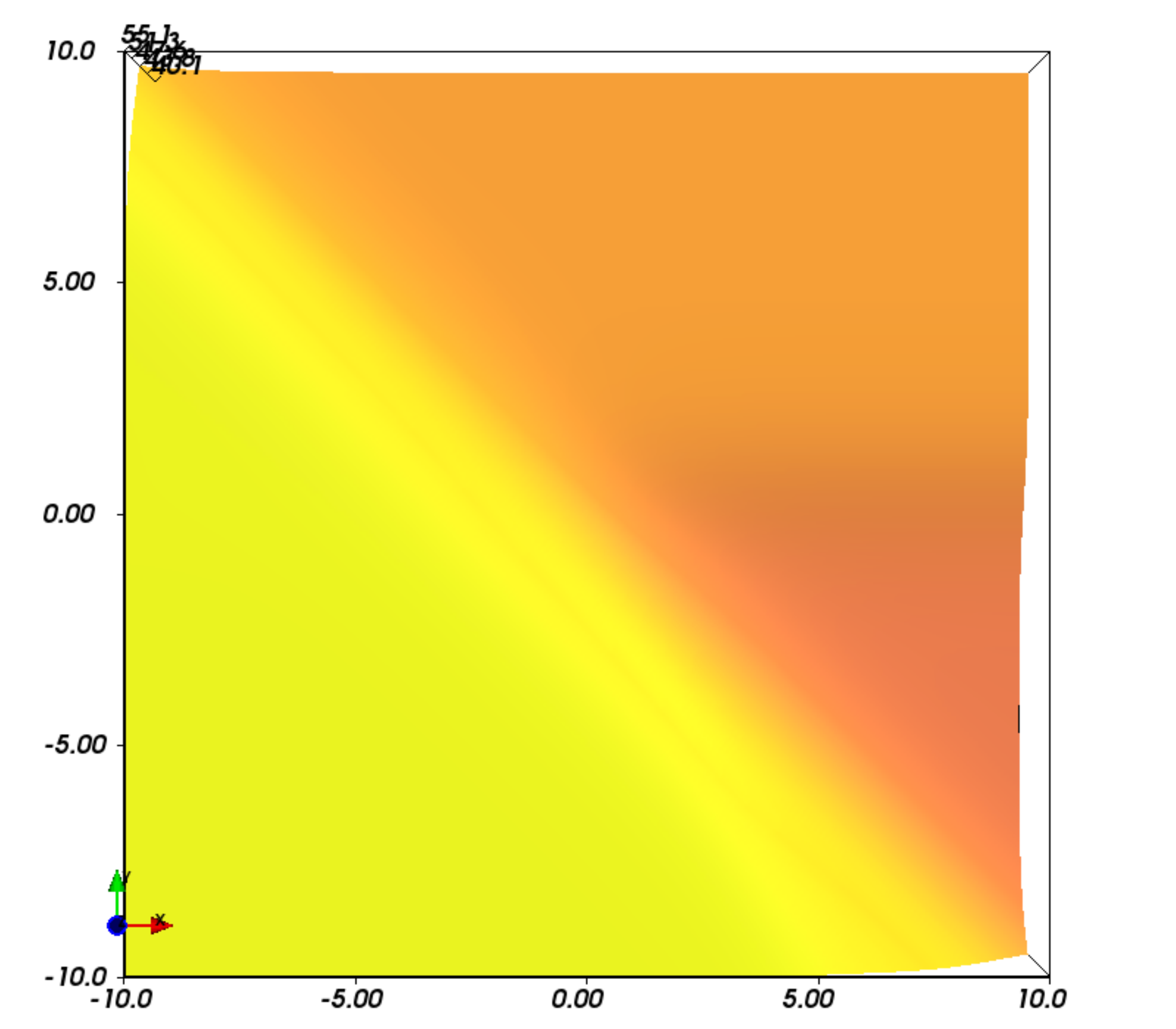}
\subcaption*{{\color{red}$\gamma=0.80 \approx \gammabw$}}
\end{subfigure}
\begin{subfigure}{0.210\textwidth}
\includegraphics[width=\textwidth]{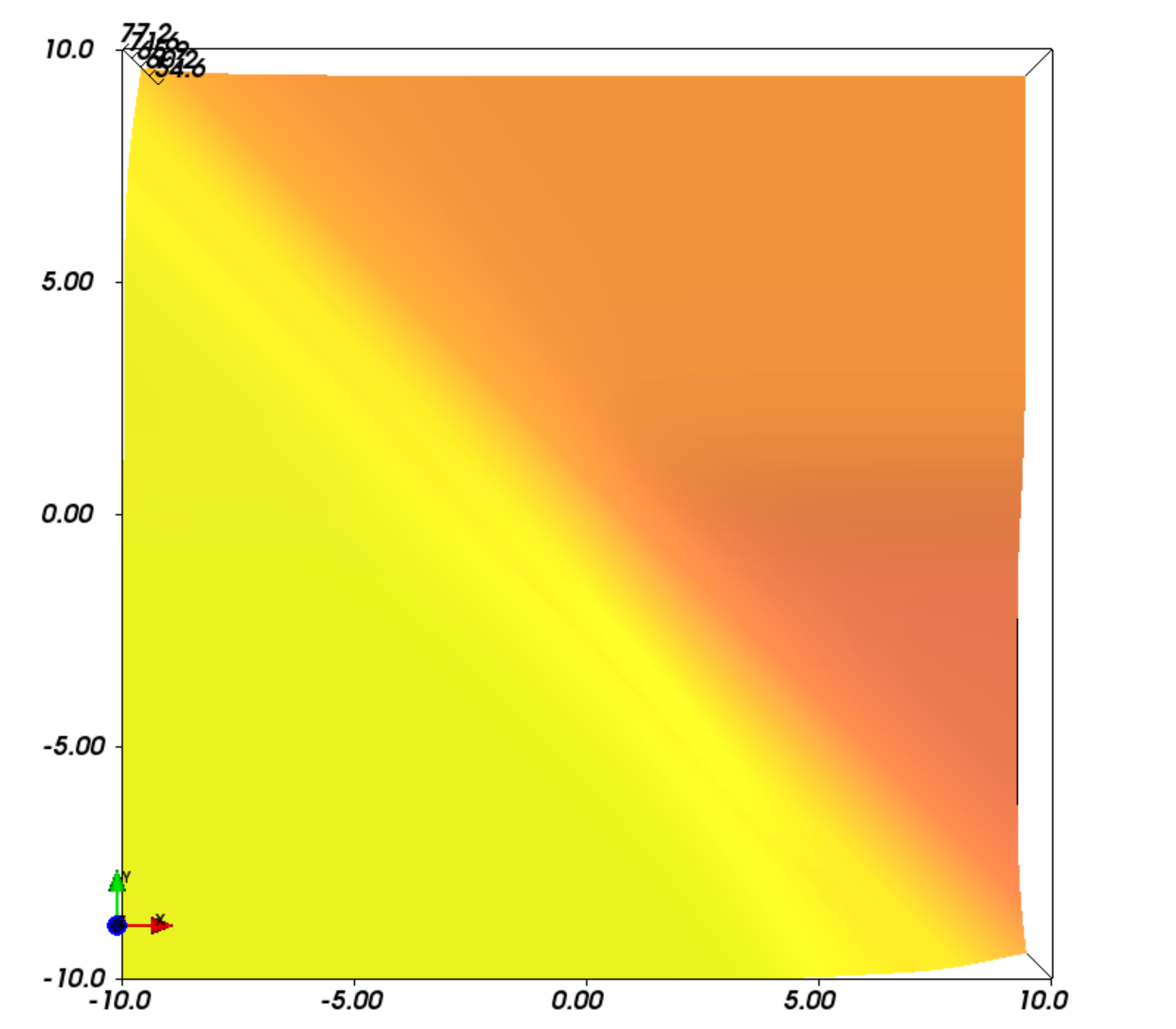}
\subcaption*{$\gamma=0.85$}
\end{subfigure}
\begin{subfigure}{0.210\textwidth}
\includegraphics[width=\textwidth]{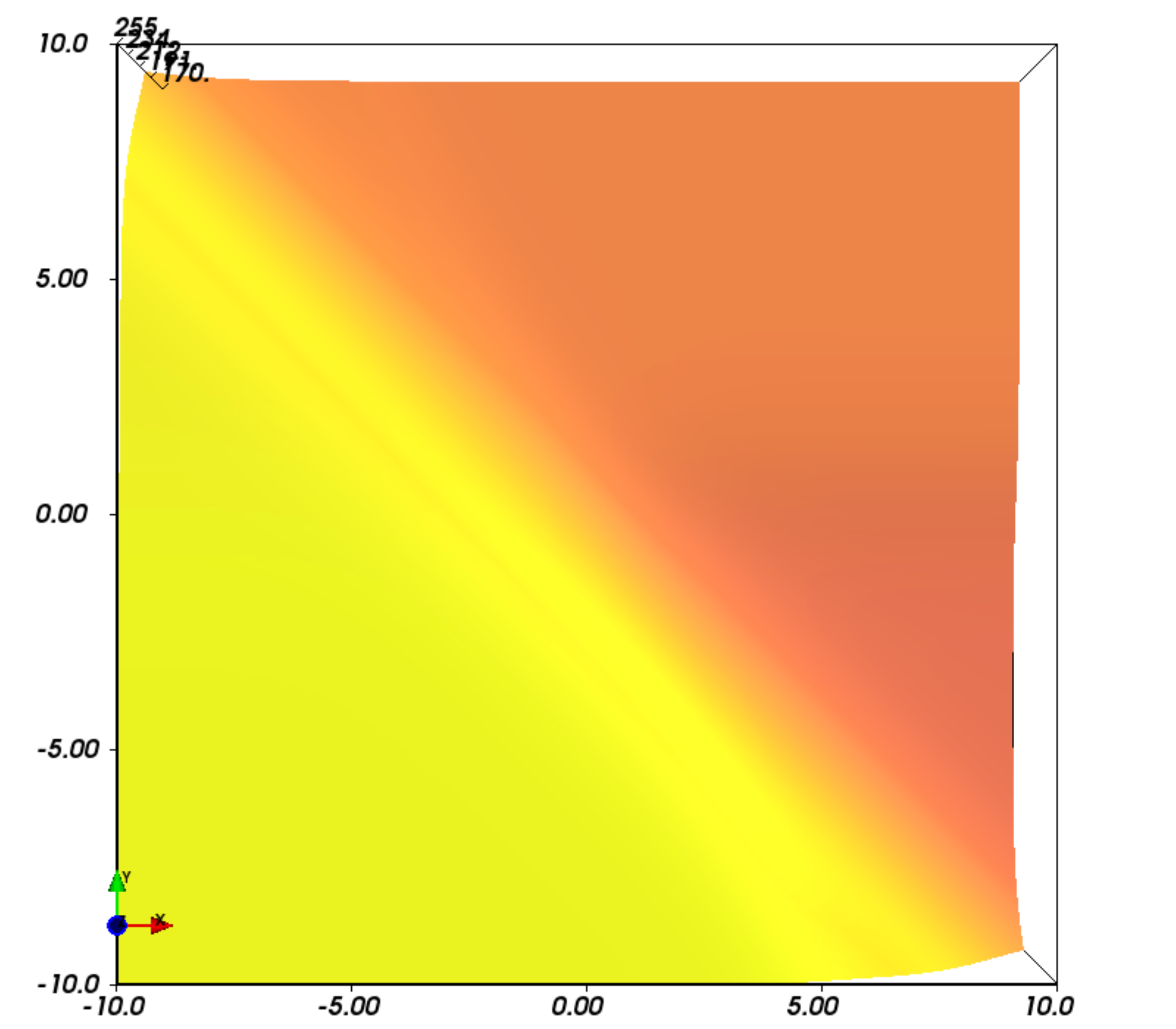}
\subcaption*{$\gamma=0.95$}
\end{subfigure}
\begin{subfigure}{0.210\textwidth}
\includegraphics[width=\textwidth]{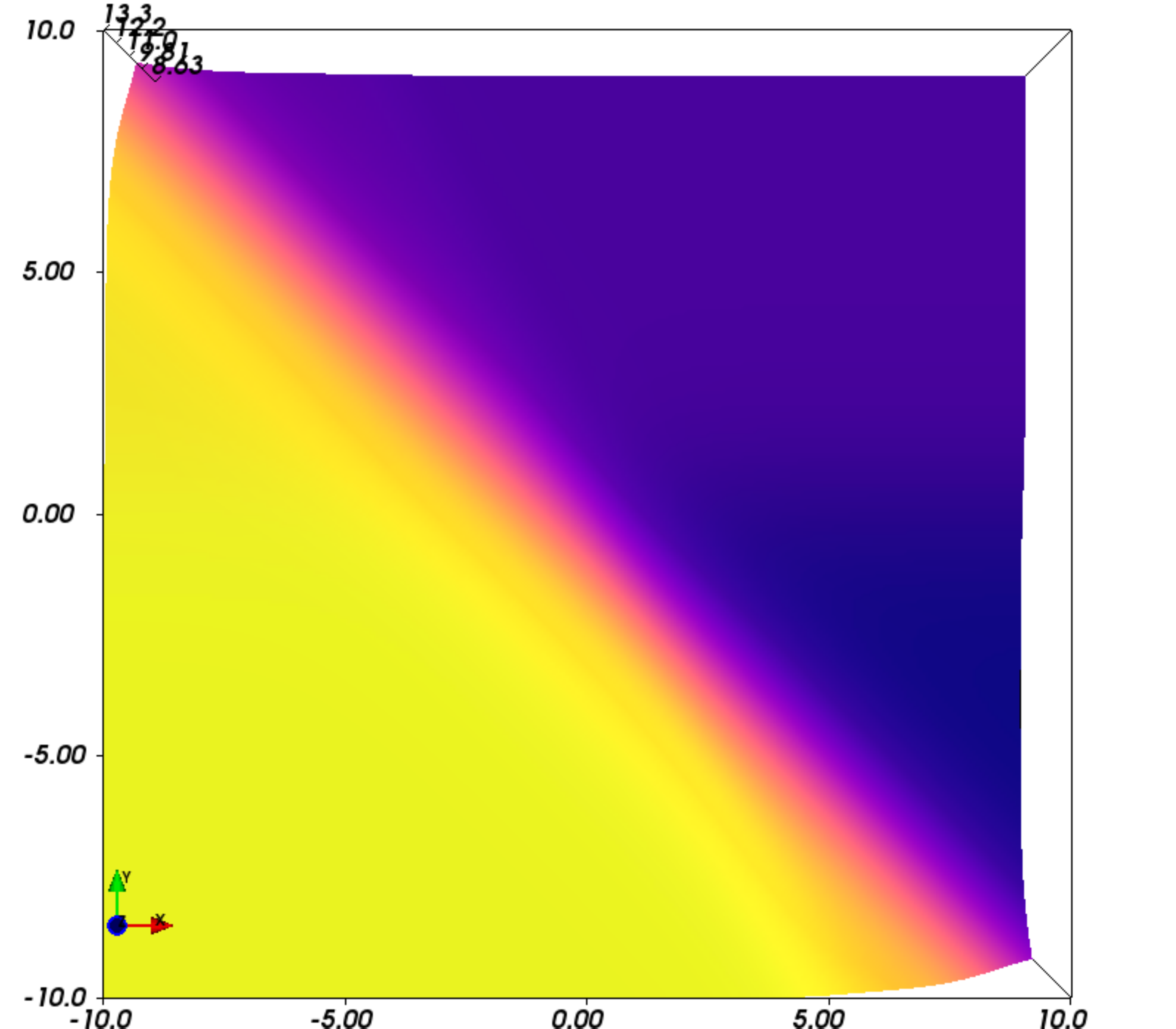}
\subcaption*{Gain}
\end{subfigure}

\begin{subfigure}{0.210\textwidth}
\includegraphics[width=\textwidth]{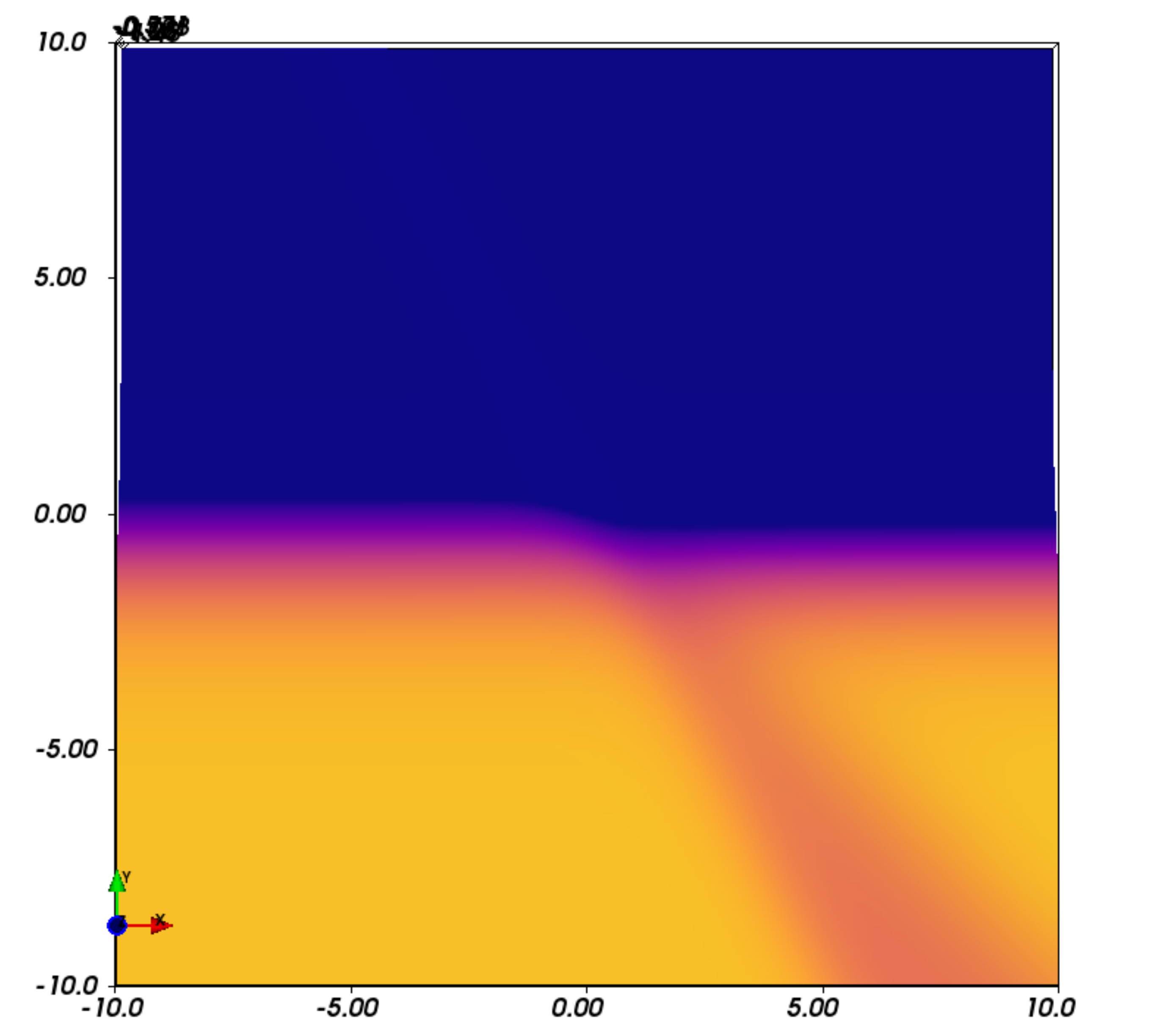}
\subcaption*{$\gamma=0.35$}
\end{subfigure}
\begin{subfigure}{0.210\textwidth}
\includegraphics[width=\textwidth]{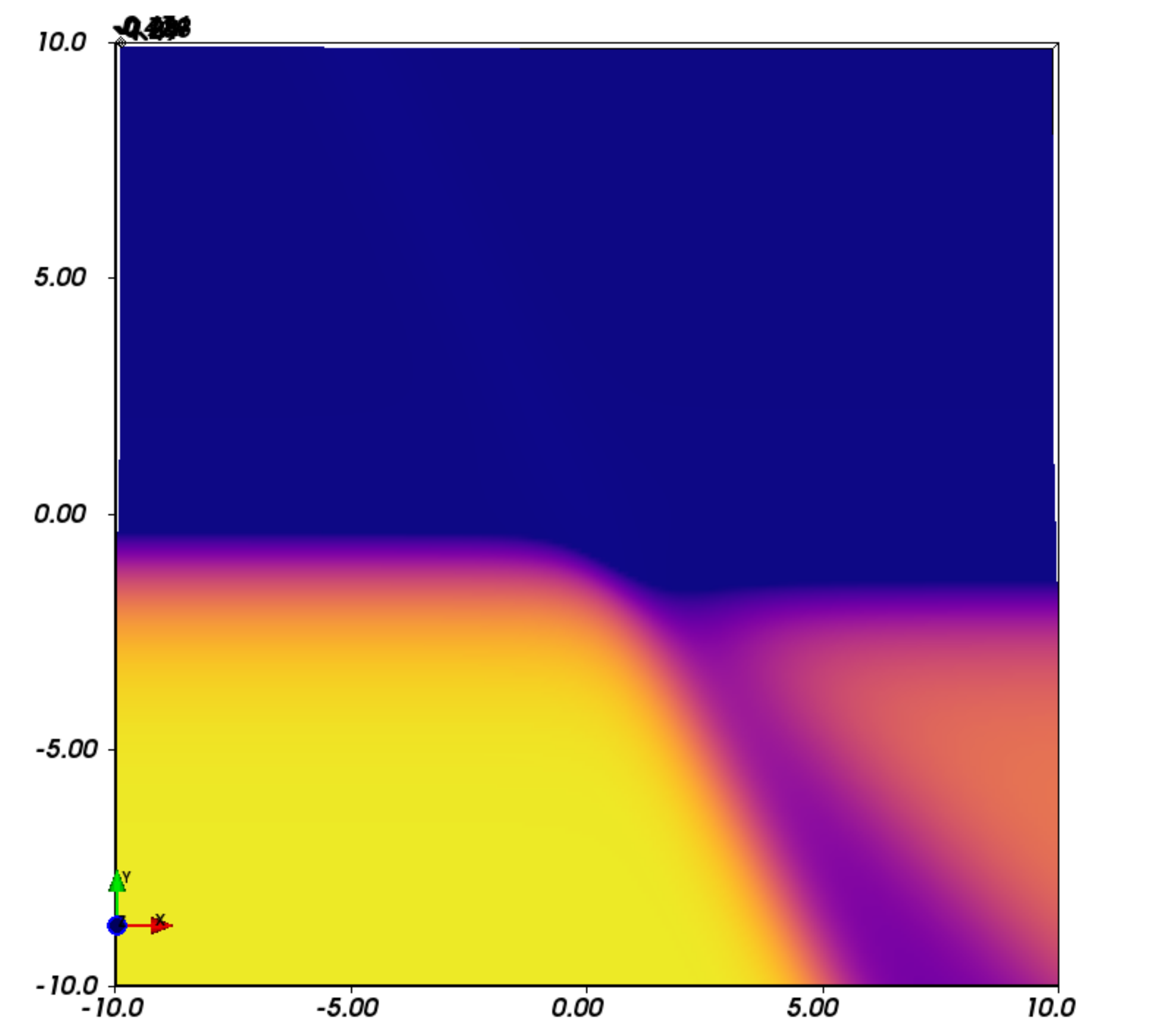}
\subcaption*{$\gamma=0.50$}
\end{subfigure}
\begin{subfigure}{0.210\textwidth}
\includegraphics[width=\textwidth]{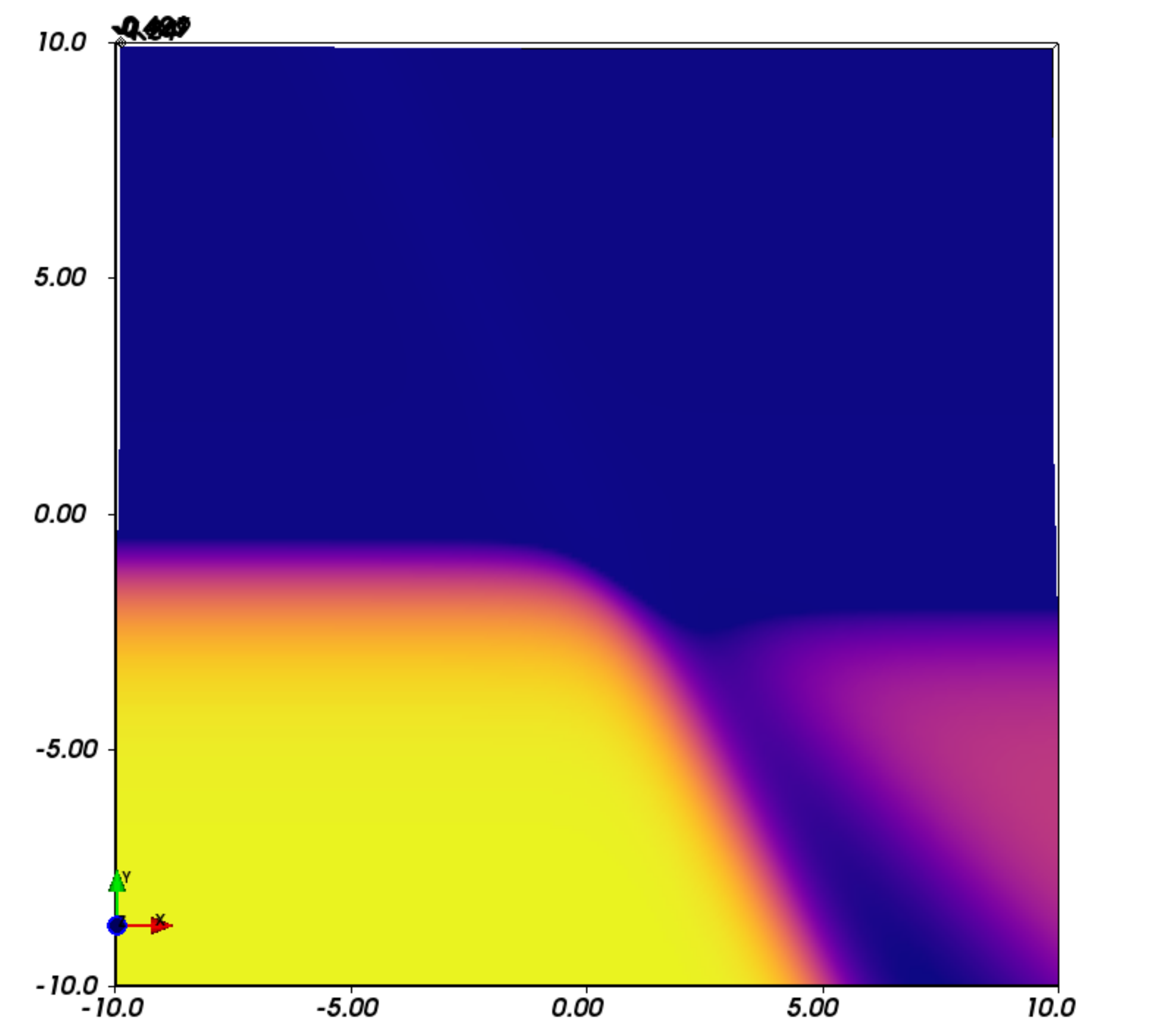}
\subcaption*{{\color{red}$\gamma=0.55 \approx \gammabw$}}
\end{subfigure}
\begin{subfigure}{0.210\textwidth}
\includegraphics[width=\textwidth]{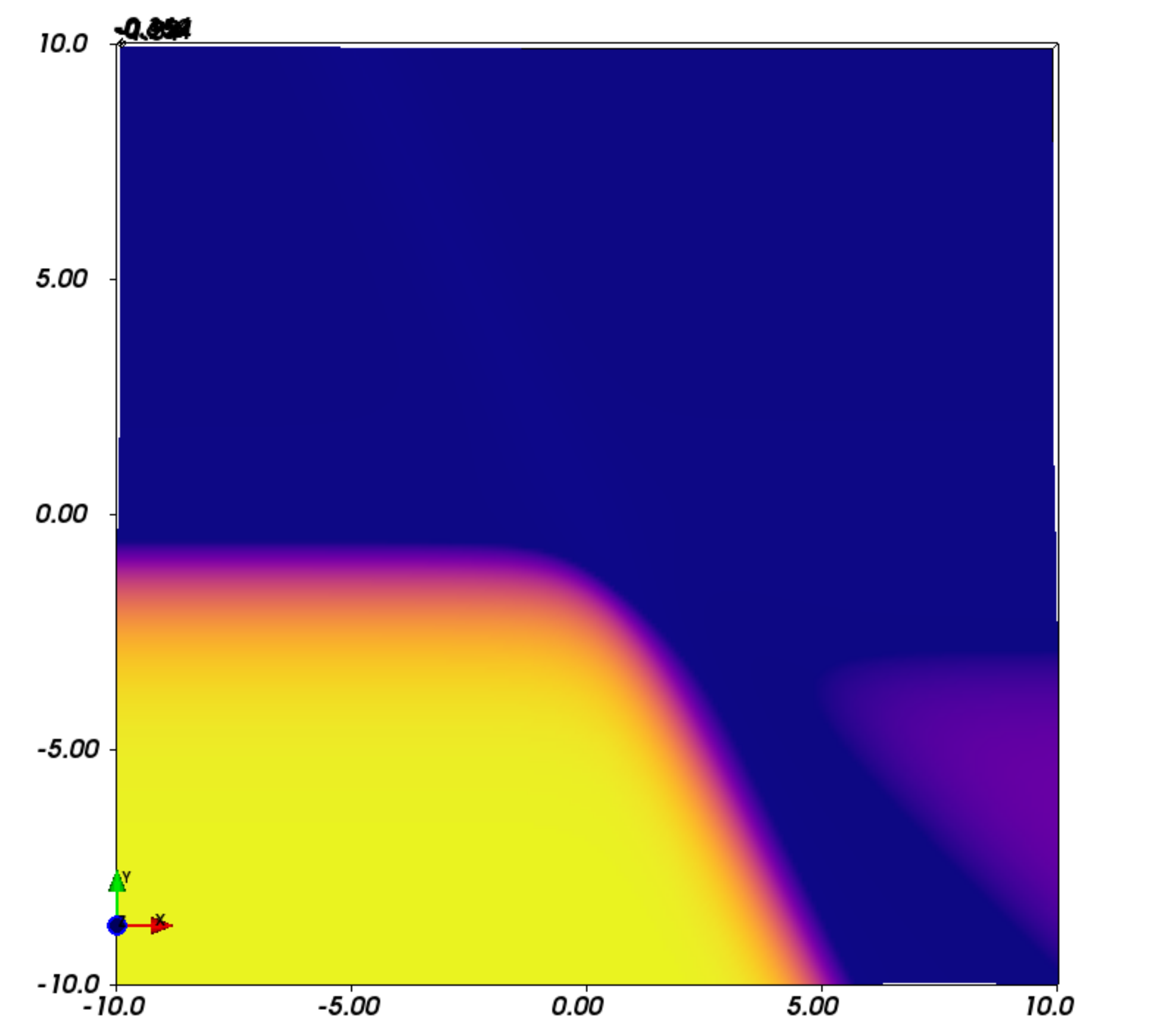}
\subcaption*{$\gamma=0.60$}
\end{subfigure}
\begin{subfigure}{0.210\textwidth}
\includegraphics[width=\textwidth]{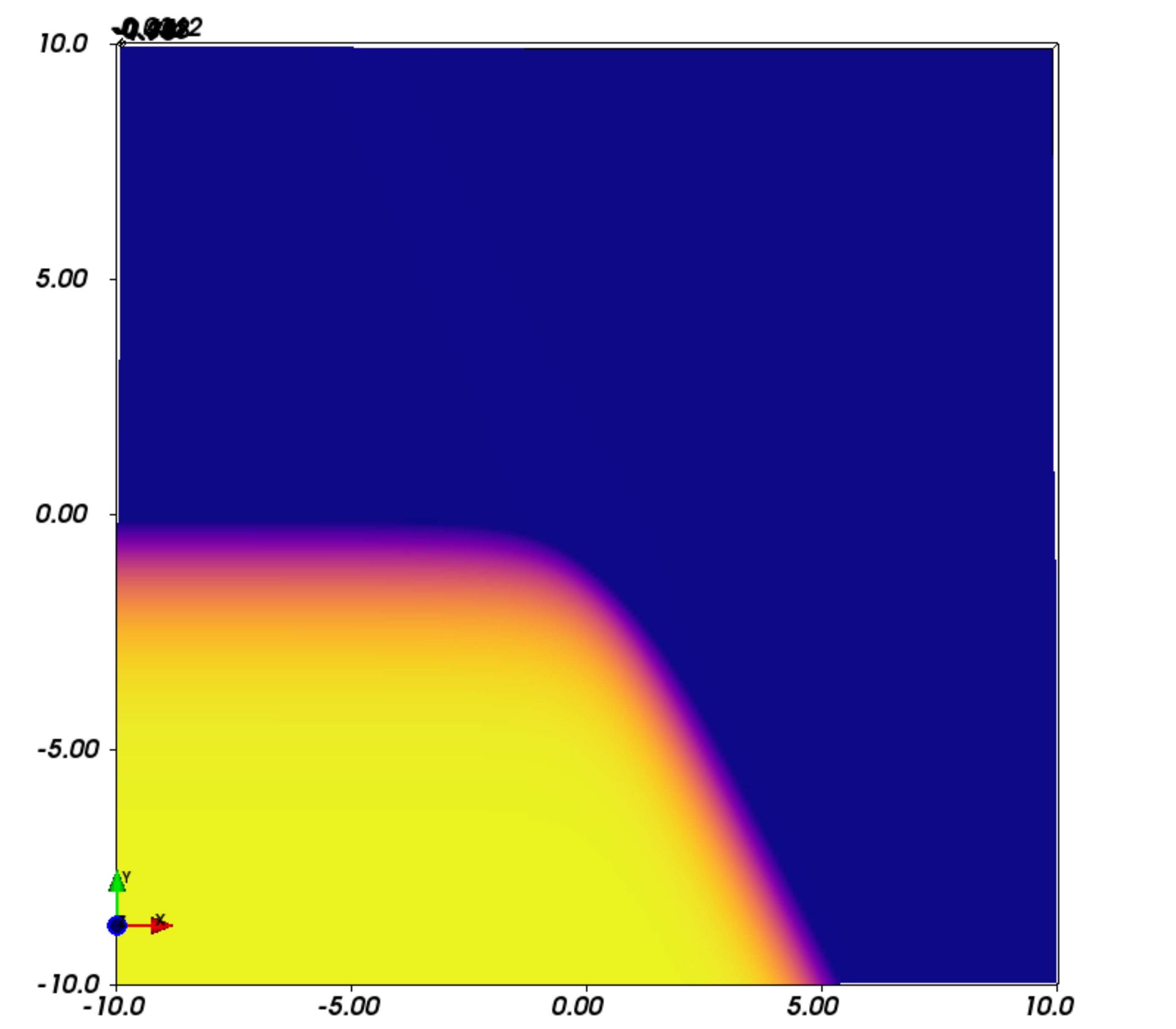}
\subcaption*{$\gamma=0.75$}
\end{subfigure}
\begin{subfigure}{0.210\textwidth}
\includegraphics[width=\textwidth]{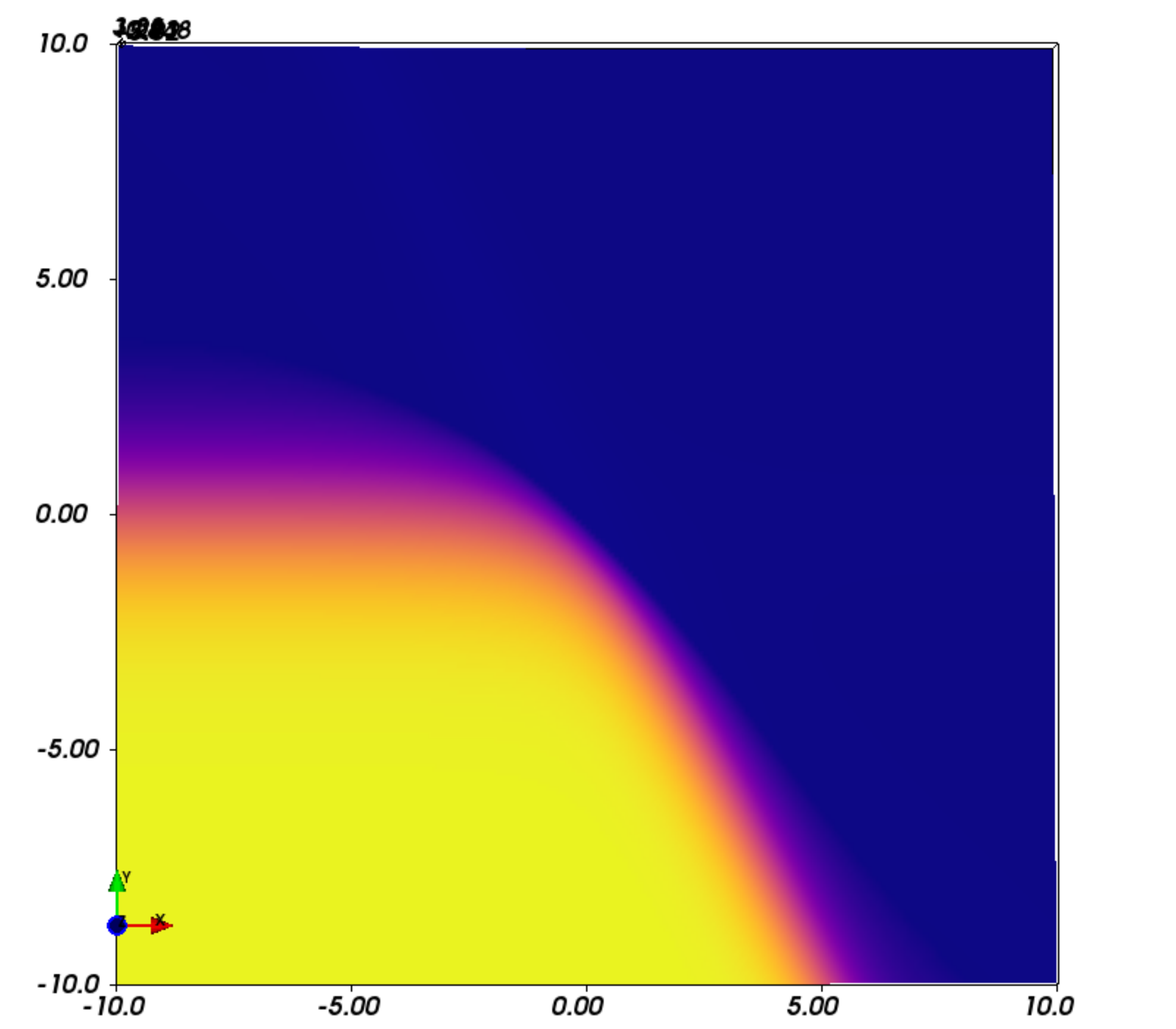}
\subcaption*{$\gamma=0.95$}
\end{subfigure}
\begin{subfigure}{0.210\textwidth}
\includegraphics[width=\textwidth]{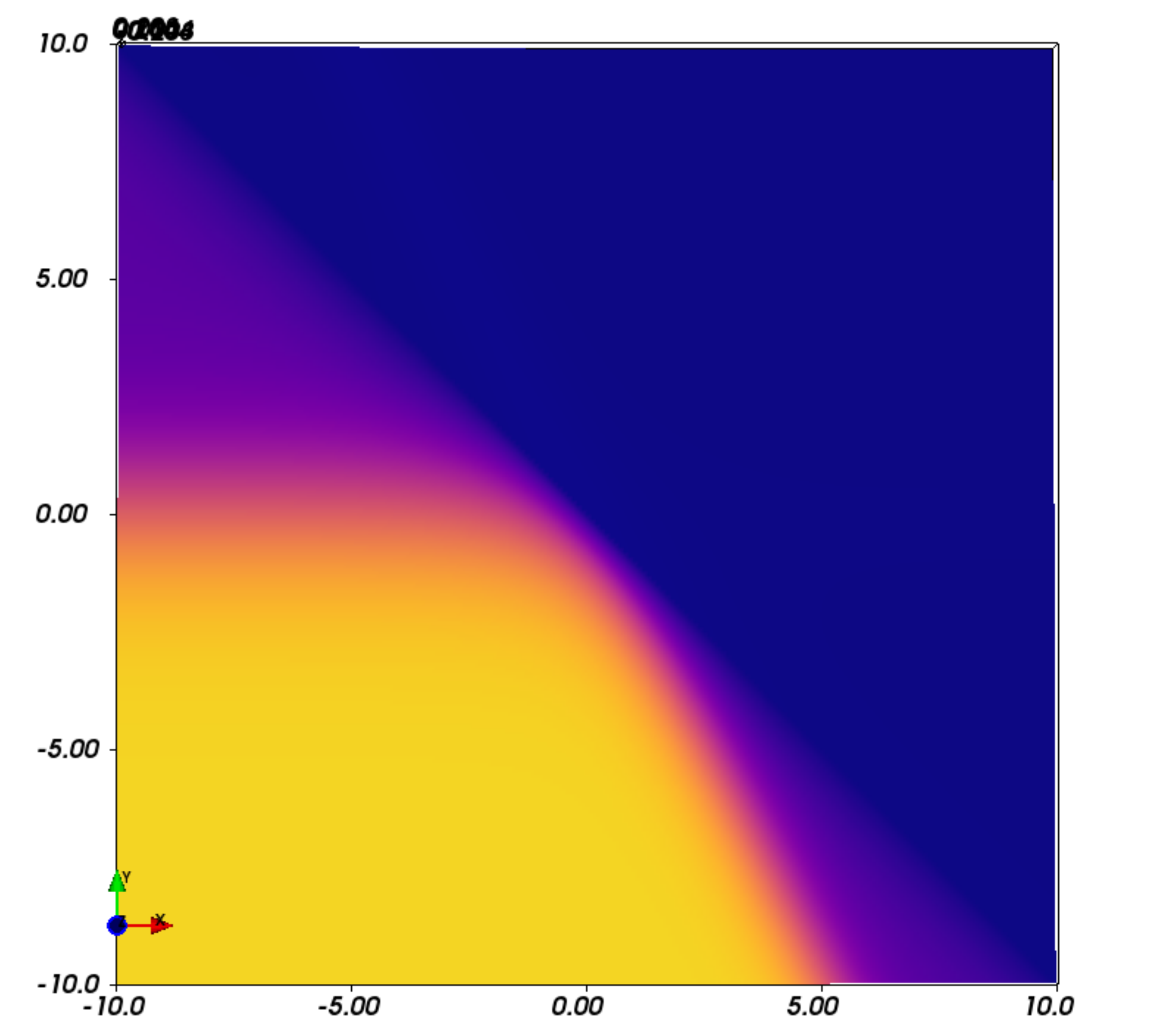}
\subcaption*{Gain}
\end{subfigure}

\caption{Policy-value landscapes as a function of two policy parameters
$\vecb{\theta} \in \real{2}$ (\ie horizontal and vertical axes)
on three environments (from top to bottom rows: Chain, Taxicab, and Torus,
which all induce recurrent MDPs where the gain optimality is the most selective).
All columns, except the rightmost (the gain~$v_g$), correspond to
the scaled discounted reward $(1 - \gamma) v_\gamma$ of
randomized policies $\pi(\vecb{\theta})$, measured from a single initial state.
The color maps the lowest value to dark-blue, and the highest value to yellow.
These lowest and highest values are respectively of the ``worst'' and
of the optimal \emph{deterministic} stationary policies
with respect to the corresponding criterion
(such that different columns (hence, subplots) have different lowest and highest values).
As anticipated, the scaled $v_\gamma$-landscape becomes more and more similar to
its gain counterpart as $\gamma$ approaches~1.
Particularly, $\gamma \ge \gammabw$ induces a scaled $v_\gamma$-landscape, whose
global maximum coordinate coincides with that of the gain.
Note that the gain landspace in the bottom row only has a yellow-ish region
because there is a significant difference between
the maximum gain $v_g^*$ of the optimal deterministic policy
\emph{and} the maximum gain $\hat{v}_g^* \approx v_g^* $ across the shown landscape
(which is only of a portion of the policy parameter space).
Their absolute difference, \ie $\abs{(\hat{v}_g^* - v_g^*)/v_g^*}$, is of $7.070\%$,
where $v_g^*=0.224$ and $\hat{v}_g^*=0.208$.
For experimental setup, see \secref{sec:xprmt_setup_avgdisrew}.
}
\label{fig:optim_landscape}
\end{figure}
\end{landscape}

%% file: avgrew.tex
\section{Benefits of maximizing the average reward in recurrent MDPs}
\label{sec:avgrew}

In this Section, we enumerate the benefits of directly maximizing
the average reward (gain) optimality criterion in recurrent MDPs.
Loosely speaking, such a recurrent structure is found in continuing environments\footnote{
    Continuing environments are those with no terminal state;
    \cf episodic environments (\fnoteref{fnote:episodic_env}).
}
with cyclical events across all states.
For episodic environments, we can obtain their recurrent MDPs by
explicitly modelling the episode repetition, as explained in \secref{sec:ep_repeat}.
For a non-exhaustive list of continuing and episodic environments, refer to
\ifpaper
\citep{dewanto_2020_avgrew}.
\fi

The combination of recurrent MDPs and gain optimality is advantageous.
There is no discount factor~$\gamma$ involved (as a by-product),
removing the difficulties due to artificial discounting (\secref{sec:gamma}).
Other benefits are described in the following sub-sections.

\input{avgrew_selective.tex}
\input{avgrew_uniformoptim.tex}
\input{avgrew_convergence.tex}
\input{avgrew_alignedmetric.tex}
\input{avgrew_eprep.tex}

%% file: avgrew_selective.tex
\subsection{Unconditionally the most selective criterion}

Gain optimality is the most selective criterion for recurrent MDPs.
This is because all states are recurrent so that the gain
is all that is needed to quantitatively measure the quality of
any stationary policy from those states
(such a gain quantity turns out to be constant for all states in
recurrent or more generally unichain MDPs).
Recall that the gain is concerned with the long-run rewards \eqref{equ:v_gain},
and recurrent states are states that are re-visited infinitely many times
in the long-run (\secref{sec:mdpclf}).

Gain optimality therefore is equivalent to Blackwell optimality \emph{unconditionally},
as well as \emph{instantaneously} in that there is no need
for hyperparameter tuning for the optimality criterion
(which fundamentally determines the optimization objective function,
hence the overall optimization).
This is in contrast to $\gamma$-discounted optimality,
where it is equivalent to Blackwell optimality if $\gamma \ge \gammabw$
as in \eqref{equ:bw_optim}.
Moreover since $\gammabw$ is unknown,
tuning $\gamma$ is necessary (\secref{sec:disrew_bwoptim}).

%% file: avgrew_uniformoptim.tex
\subsection{Uniformly optimal policies}

Since recurrent (up to unichain) MDPs have only a single recurrent class (chain),
the gain of any policy is constant across all states.
As a result, average-reward policy gradient methods maximize an objective
\eqref{equ:avgpg_obj} that is independent of initial states,
or generally of initial state distributions.
The resulting gain-optimal policies are said to be \emph{uniformly optimal} because
they are optimal for all initial states or all initial state distributions
\citep[\deff{2.1}]{altman_1999_cmdp}.

In contrast, the discounted-reward counterpart maximizes an objective
\eqref{equ:dispg_obj} that is defined with respect to
some initial state distribution $\isd$ (since the discounted-reward value $v_\gamma$
depends on the state from which the value is measured, as in \eqref{equ:v_disc}).
Consequently, the resulting $\gamma$-discounted optimal policies
may not be optimal for all initial states; they are said to be
\emph{non-uniformly optimal}, as noted by \citet[\page{41}]{bacon_2018_thesis}.
This non-uniform optimality can be interpreted as a relaxation of
the uniform optimality in DP, which requires that
the superiority of optimal policies $\pi_\gamma^*$ holds in every state,
\ie $v_\gamma(\pi_\gamma^*, s_0) \ge v_\gamma(\pi, s_0)$,
for all states $s_0 \in \setname{S}$ and all policies $\pi \in \piset{S}$,
see \secref{sec:optcrit}.

The objectives of average- and discounted-reward policy gradient methods
are as follows.
\begin{align}
\text{Average-reward policy gradient objective:}
& \quad \argmax_{\vecb{\theta} \in \Theta} v_g(\pi(\vecb{\vecb{\theta}})),
    \label{equ:avgpg_obj} \\
\text{Discounted-reward policy gradient objective:}
& \quad \argmax_{\vecb{\theta} \in \Theta} \E{\isd}{v_\gamma(\pi(\vecb{\vecb{\theta}}), S_0)},
    \label{equ:dispg_obj}
\end{align}
where $\vecb{\theta} \in \Theta = \real{\dim(\vecb{\theta})}$ is
the policy parameter and $S_0 \sim \isd$ is the initial state random variable.

%% file: avgrew_convergence.tex
\subsection{Potentially higher convergence rates}

Without delving into specific algorithms, there are at least two reasons for
faster convergence of average-reward methods (compared to their discounted-reward counterparts),
as hinted by \citet[\secc{6}]{schwartz_1993_rlearn} and
\citet[\page{114}]{vanroy_1998_thesis}.

\textbf{First} is the common gain across states.
Such commonality eases the gain approximation in that no generalization is required.
That is, a single gain estimation is all that is needed for one true gain
of all states in unichain MDPs.

\textbf{Second}, average-reward methods optimize solely the gain term in
the Laurent series expansion of~$v_\gamma$ in \eqref{equ:laurent_expansion}.
On the other hand, their discounted-reward counterparts optimize
the gain, bias, and higher-order terms altogether simultaneously,
whose implication becomes more substantial as $\gamma$ is set further below~1,
as can be observed in \eqref{equ:laurent_expansion_truncated}.

%% file: avgrew_alignedmetric.tex
\subsection{Less discrepancy between the learning objective and the performance metric}

Since there is a finite number of timesteps (as training budget) and
no inherent notion of discounting,
the \emph{final} learning performance metric $\psitotasy$ in RL is typically measured by
the average \emph{return} over several last experiment-episodes\footnote{
    The term ``experiment-episode'' refers to one trial (run, roll-out, or simulation),
    which produces one sample of trajectory (of states, actions, and rewards).
    The term ``experiment-episode'' is applicable to both episodic and continuing environments.
    In contrast, the term ``episode'' only makes sense for episodic environments.
}
\citetext{\citealp[\secc{5.2}]{machado_2018_arcade}; \citealp[\fig{2}]{henderson_2018_rlmatter}}.
That is,
\begin{align}
\psitotasy
& \eqdef
\frac{1}{\nxeplast} \sum_{i= j + 1}^{\nxep} \Bigg[
\underbrace{
     \sum_{t=0}^{\tepmaxi} r_{t+1}^{(i)} \Big| \hat{\pi}^*
}_\text{Return} \Bigg]
\tag{with $j = \nxep - \nxeplast$} \\
& \approx
\underbrace{
\mathbb{E} \Bigg[ \sum_{t=0}^{\tepmax} R_{t + 1} \Big| \hat{\pi}^* \Bigg]
}_\text{Finite-horizon total reward}
\propto
\underbrace{
\frac{1}{\tepmax + 1}
    \mathbb{E} \Bigg[ \sum_{t=0}^{\tepmax} R_{t + 1} \Big| \hat{\pi}^* \Bigg]
}_\text{Finite-horizon average reward}
\approx \underbrace{v_g(\hat{\pi}^*)}_\text{Gain},
\label{equ:perfmetric_asymp}
\end{align}
where $\nxep$ denotes the number of i.i.d experiment-episodes,
from which we pick the last $\nxeplast$ experiment-episodes
once the learning is deemed converged to a final learned policy~$\hat{\pi}^*$.
Each $i$-th experiment-episode runs from $t=0$ to a finite $\tepmaxi \approx \tmax$,
where the reward realization at every timestep $t$ is denoted by $r_{t + 1}^{(i)}$.
The expectation in \eqref{equ:perfmetric_asymp} of the reward random variable
$R_{t + 1} \eqdef r(S_t, A_t)$ is with respect to
$S_0 \sim \isd$, $A_t \sim \hat{\pi}^*(\cdot|s_t)$ and $S_{t+1} \sim p(\cdot|s_t, a_t)$.

We argue that the learning performance metric $\psitotasy$ has less discrepancy
with respect to the gain~$v_g$ than to the discounted reward~$v_\gamma$.
In other words, whenever the criterion is $v_g$, what is measured during training
(\ie the learning performance metric) is more aligned to what the agent learns/optimizes
(\ie the learning objective).\footnote{
    In some cases, this alignment may be traded-off for more tractable computation/training/analysis.
}
There are three arguments for this as follows.

\textbf{First}, since the experiment-episodes are i.i.d, the expectation of
$\psitotasy$ in \eqref{equ:perfmetric_asymp} converges as $\nxeplast \to \infty$
to the expected finite-horizon total reward.
This is proportional to the expected \emph{finite-horizon} average reward,
which is an approximation to the gain $v_g(\hat{\pi}^*)$
due to the finiteness of~$\tepmax$.

\textbf{Second} is the fact that $\tepmax$ is typically fixed for
all experiment-episodes during training.\footnote{
    One common practice is to set $\tepmax$ to some constant far less than
    the training budget to obtain multiple experiment-episodes.
    This is implemented as for instance, the `max\_episode\_steps' variable at
    \url{https://github.com/openai/gym/blob/master/gym/envs/__init__.py}
    of a seminal and popular RL environment codebase: OpenAI Gym \citep{brockman_2016_gym}.
}
It is not randomly sampled from a geometric distribution $Geo(p = 1 - \gamma)$
even when maximizing $v_\gamma$ (seemingly because there is no inherent notion of discounting).
If it was sampled from such a geometric distribution, then the expectation of $\psitotasy$
would converge to $v_\gamma$, following the identity in \eqref{equ:rndtermination}.
This however would inherently pose a risk of miss-specifying $\gamma$ (\secref{sec:gamma_lo}).
This metric gap due to artificial discounting is highlighted by
\citet[\secc{3}]{schwartz_1993_rlearn}, \citet[\page{25}]{vanhasselt_2011_thesis},
\citet[\page{47}]{dabney_2014_thesis}, and \citet[\secc{2}]{seijen_2019_lowdiscount}.

\textbf{Third} is because the induced Markov chain may reach stationarity
(or close to it) within $\tepmax < \infty$ in some environments.
That is, $\ppimat^t \ppimat = \ppimat^{t+1} = \ppimat^t = \ppimat^\star$
for several timesteps $t \le \tepmax$, see \eqref{equ:pstar_pgamma_def}.
If this happens, then some number of rewards are generated from states sampled
from the stationary state distribution $p_\pi^\star$
(although they are not independent, but Markovian state samples).
This makes the approximation to the gain $v_g$ more accurate
(although it is inherently biased due to non-i.i.d state samples)
than to the discounted $v_\gamma$ since
$v_g(\pi) = \E{S \sim p_\pi^\star}{r_\pi(S) \eqdef \E{A \sim \pi}{r(S, A)}}$,
which can be shown to be equivalent to~\eqref{equ:v_gain}.

%% file: avgrew_eprep.tex
\subsection{Modelling the episode repetition explicitly in episodic environments}
\label{sec:ep_repeat}

\input{avgrew_eprep_model.tex}

In order to obtain an infinite-horizon recurrent MDP of an episodic environment,
we can model its episode repetition explicitly.
This is in contrast to modelling an episodic environment as
an infinite-horizon MDP with a 0-reward absorbing terminal state
(denoted as $\szrat$, which is recurrent under every policy).
This modelling (called $\szrat$-modelling) induces a unichain MDP,
where all states but $\szrat$ are transient.
In $\szrat$-model, the gain is trivially~0 for all stationary policies,
rendering the gain optimality underselective (\secref{sec:disrew_bwoptim}).
\figref{gridnav2_szrat} shows the diagram of a $\szrat$-model.

To explicitly model episode repetition, we augment the original state set
with a resetting terminal state~$\sreset$ (instead of $\szrat$)
that has a single available reset action $\areset$
(instead of a self-loop action $\azrat$).
This action $\areset$ is responsible for a transition from
$\sreset$ to an initial state $S_0 \sim \isd$, which yields a reward of~0.
We call this approach $\sreset$-modelling, whose example diagram is shown
in \figref{gridnav2_sreset}.
The conversion from a unichain $\szrat$-model to
a recurrent $\sreset$-model is as follows.

Let $\setname{S}$ and $\setname{A}$ denote the original state and action
sets of an episodic environment, respectively,
before the augmentation of a 0-reward absorbing terminal state $\szrat$.
Given a unichain $\szrat$-model with
a state set $\ssetzrat = \setname{S} \cup \{ \szrat \}$,
an action set $\asetzrat = \setname{A} \cup \{ \azrat \}$,
an initial state distribution $\isdzrat$,
a state transition distribution $\pzrat$, and
a reward function $\rzrat$,
then the corresponding recurrent $\sreset$-model has the following components.
\begin{itemize}
\item A state set $\ssetreset = \setname{S} \cup \{ \sreset \}$,
    an action set $\asetreset = \setname{A} \cup \{ \areset \}$, and
    an initial state distribution $\isdreset = \isdzrat$,
    where $\isdreset(\sreset) = \isdzrat(\szrat) = 0$.
\item A state transition distribution $\preset$ and
    a reward function $\rreset$, where
    \begin{itemize}
    \item[$\bullet$] $\preset(s'|s, a) = \pzrat(s'|s, a)$
        and $\rreset(s, a, s') = \rzrat(s, a, s')$,
        for all ${s, s' \in \setname{S}}$ and for all ${a \in \setname{A}}$,
    \item[$\bullet$] $\preset(\sreset |s, a) = \pzrat(\szrat|s, a)$
        and $\rreset(s, a, \sreset) = \rzrat(s, a, \szrat)$,
        for all ${s, s' \in \setname{S}}$ and for all $a \in \setname{A}$,
        \emph{as well as}
    \item[$\bullet$] $\preset(s |\sreset, \areset) = \isdzrat(s)$
        and $\rreset(\sreset, \areset, s) = 0$,
        for all $s \in \setname{S}$.
    \end{itemize}
\end{itemize}
The above conversion requires that
i)~reaching $\szrat$ is inevitable with probability 1 under all stationary policies
(this is equivalent to inevitable termination in an episodic environment), and
ii)~there is no inherent notion of discounting that operates from $t = 0$
until the end of an episode.
For example diagrams of both $\szrat$- and $\sreset$-models, refer to \figref{fig:eprep_model}.

The $\sreset$-modelling suits the fact that in practice,
an RL-agent is expected to run in multiple episodes, \eg to play some game repeatedly.
By using $\sreset$-modelling, we can train an agent operating on episodic environments
in the same practical way as if it were operating on continuing environments.
Similar to the zero state-value of the terminal state (\ie $v_\gamma^\pi(\szrat) = 0$)
in discounted-reward $\szrat$-modelling,
we may have a zero state-value of the resetting state (\ie $v_b^\pi(\sreset) = 0$)
in the average-reward $\sreset$-modelling, where
$v_b$ denotes the relative value of the bias \eqref{equ:laurent_expansion_truncated}.

\input{avgrew_eprep_xprmt.tex}

We compare $\szrat$- and $\sreset$-modelling by running experiments with
two training schemes as follows.
\begin{itemize}
\item Scheme-A uses an $\szrat$-model and the total reward criterion
    (hence, $Q_{\mathrm{tot}}$-learning).
\item Scheme-B uses an $\sreset$-model and the average reward criterion
    (hence, $Q_b$-learning).
\end{itemize}
\figref{fig:eprep_xprmt} depicts the learning curves of $Q_x$-learning trained
under Scheme-A and Scheme-B on two episodic environments.
Both schemes are trained with the same experiment-episode length\footnote{
    During training on episodic environments, an agent is always transported back
    to the initial state after an episode ends, regardless of the training scheme.
    This transportation to initial states is captured by $\sreset$-modelling,
    but not by $\szrat$-modelling (see \figref{fig:eprep_model}).
}, and gauged by the same performance metric, \ie
the finite-time average reward \eqref{equ:perfmetric_asymp}.
As can be observed, both schemes converge to the same value, but
Scheme-B empirically leads to a higher rate of convergence.
This indicates that both $\szrat$- and $\sreset$-models induce
the same optimal policy (in the original states $s \in \setname{S}$),
and that the $\szrat$- to $\sreset$-model conversion is sound.
More importantly, conversion to an $\sreset$-model enables obtaining
the optimal policy using an average-reward method since such a model is recurrent,
for which gain optimality is the most selective.

\citet[\secc{3.1}]{mahadevan_1996_avgrew} mentioned the idea about
episode repetitions in an episodic grid-navigation environment.
He however, did not provide a formal conversion.
\cite{pardo_2018_timelim} also used a similar conception to the episode repetition
for proposing a technique that bootstraps from the value of the state at
the end of each experiment-episode for random-horizon episodic environments.
Another important related work is of \cite{white_2017_taskspec} who
introduced a unification of episodic and continuing environment specifications.
Such a unification is carried out via transition-based discounting,
where the discount factor from the terminal to initial states is set to~0.
She noted that the transition-based discounting breaks
the Laurent-expansion-based connection between
the discounted- and average-reward criteria in \eqref{equ:laurent_expansion}.

%% file: avgrew_eprep_model.tex
\begin{figure} [t]
\centering

\begin{subfigure}{0.35\textwidth}
\resizebox{\textwidth}{!}{
\begin{tikzpicture}[
node distance = 3cm and 3cm, on grid,
-{Latex[length=3mm]}, %
semithick, %
state/.style={circle, top color = white, bottom color = white,
    draw, text=black, minimum width = 1cm},
state_terminal/.style={regular polygon, regular polygon sides=4,
    top color = white, bottom color = white,
    draw, text=black, minimum width = 2cm}
]
\node[state](A)[] {$s^{0}$}; %
\node[state](B)[right=of A] {$s^{1}$}; %
\node[state](C)[below=of A] {$s^{2}$}; %
\node[state_terminal](D)[below=of B] {{\color{red} $\szrat$}};
\path (A) edge [bend left] node[above] {} (B);
\path (A) edge [bend right] node[left] {} (C);
\path (B) edge [bend left] node[right] {} (D);
\path (C) edge [bend right] node[below] {} (D);
\path (D) edge [red, out=90, in=180, loop] node[right]
    {{\color{red}$0$}} (D);
\end{tikzpicture}
} %
\subcaption{{\color{red} A unichain $\szrat$-model}}
\label{gridnav2_szrat}
\end{subfigure}
\begin{subfigure}{0.35\textwidth}
\resizebox{\textwidth}{!}{
\begin{tikzpicture}[
node distance = 3cm and 3cm, on grid,
-{Latex[length=3mm]}, %
semithick, %
state/.style={circle, top color = white, bottom color = white,
    draw, text=black, minimum width =1 cm},
state_reset/.style={regular polygon, regular polygon sides=4,
    top color = white, bottom color = white,
    draw, text=black, minimum width = 2cm}
]
\node[state](A)[] {$s^{0}$}; %
\node[state](B)[right=of A] {$s^{1}$}; %
\node[state](C)[below=of A] {$s^{2}$}; %
\node[state_reset](D)[below=of B] {{\color{blue}$\sreset$}};
\path (A) edge [bend left] node[above] {} (B);
\path (A) edge [bend right] node[left] {} (C);
\path (B) edge [bend left] node[right] {} (D);
\path (C) edge [bend right] node[below] {} (D);
\path (D) edge [blue, left] node[right] {{\color{blue}$0$}} (A);
\path (D) edge [out=135, in=45, blue] node[below] {{\color{blue}$0$}} (C);
\path (D) edge [out=135, in=225, blue] node[right] {{\color{blue}$0$}} (B);
\end{tikzpicture}
} %
\subcaption{{\color{blue} A recurrent $\sreset$-model}}
\label{gridnav2_sreset}
\end{subfigure}

\caption{Diagrams of
{\color{red} a unichain $\szrat$-model} and
{\color{blue} a recurrent $\sreset$-model}
of an episodic environment with an (original) state set
$\setname{S} = \{ s^0, s^1, s^2 \}$.
The red edge indicates a deterministic transition via
a single self-loop action $\azrat$ in a 0-reward absorbing terminal state $\szrat$.
On the other hand, the blue edges indicate possible transitions via
a single reset action $\areset$ in a resetting state $\sreset$.
This $\areset$ leads to a next state that is distributed according to
the initial state distribution~$\isd$
(which here has all states $s \in \setname{S}$ as its supports).
Transitions via $\azrat$ and $\areset$ yield a zero reward as indicated by
the red and blue edge labels.
The other black unlabeled edges represent examples of transitions with
positive probabilities and any reward values.
For a formal $\szrat$- to $\sreset$-model conversion, refer to \secref{sec:ep_repeat}.
}
\label{fig:eprep_model}
\end{figure}
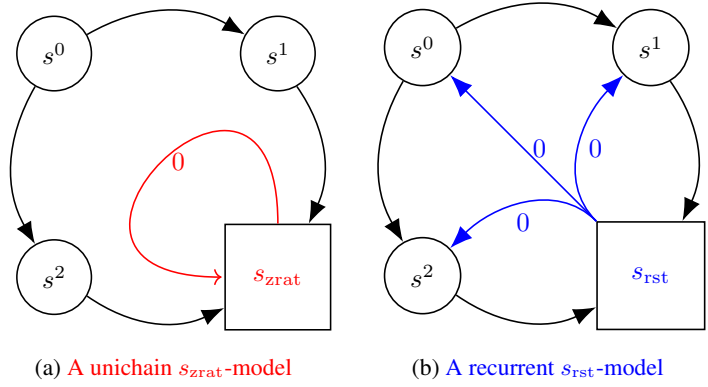

%% file: avgrew_eprep_xprmt.tex
\begin{figure*} [t]
\centering

\begin{subfigure}{0.475\textwidth}
\includegraphics[width=\textwidth]{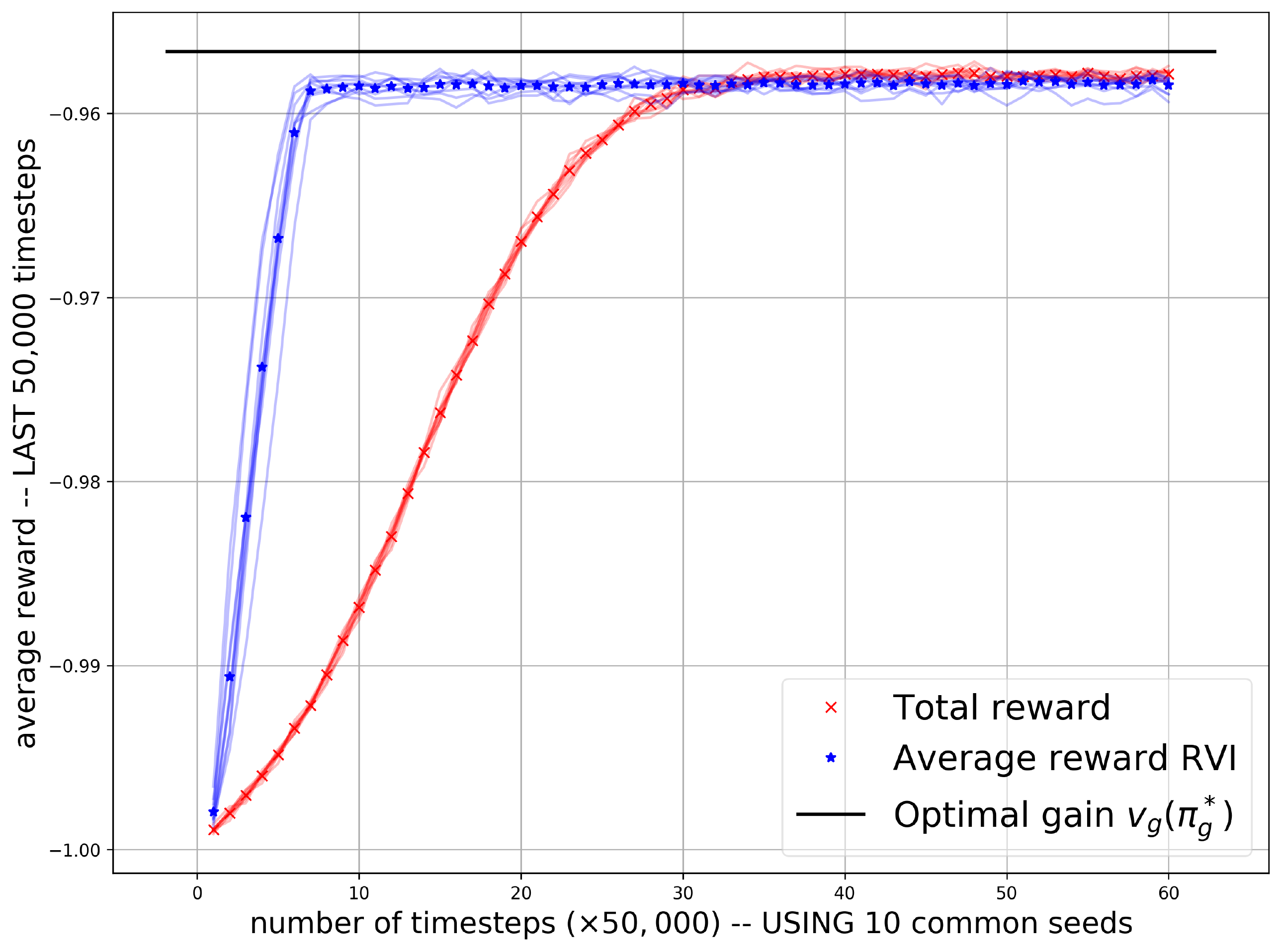}
\subcaption{Episodic: GridNav-25}
\label{lc_gn25}
\end{subfigure}
\begin{subfigure}{0.475\textwidth}
\includegraphics[width=\textwidth]{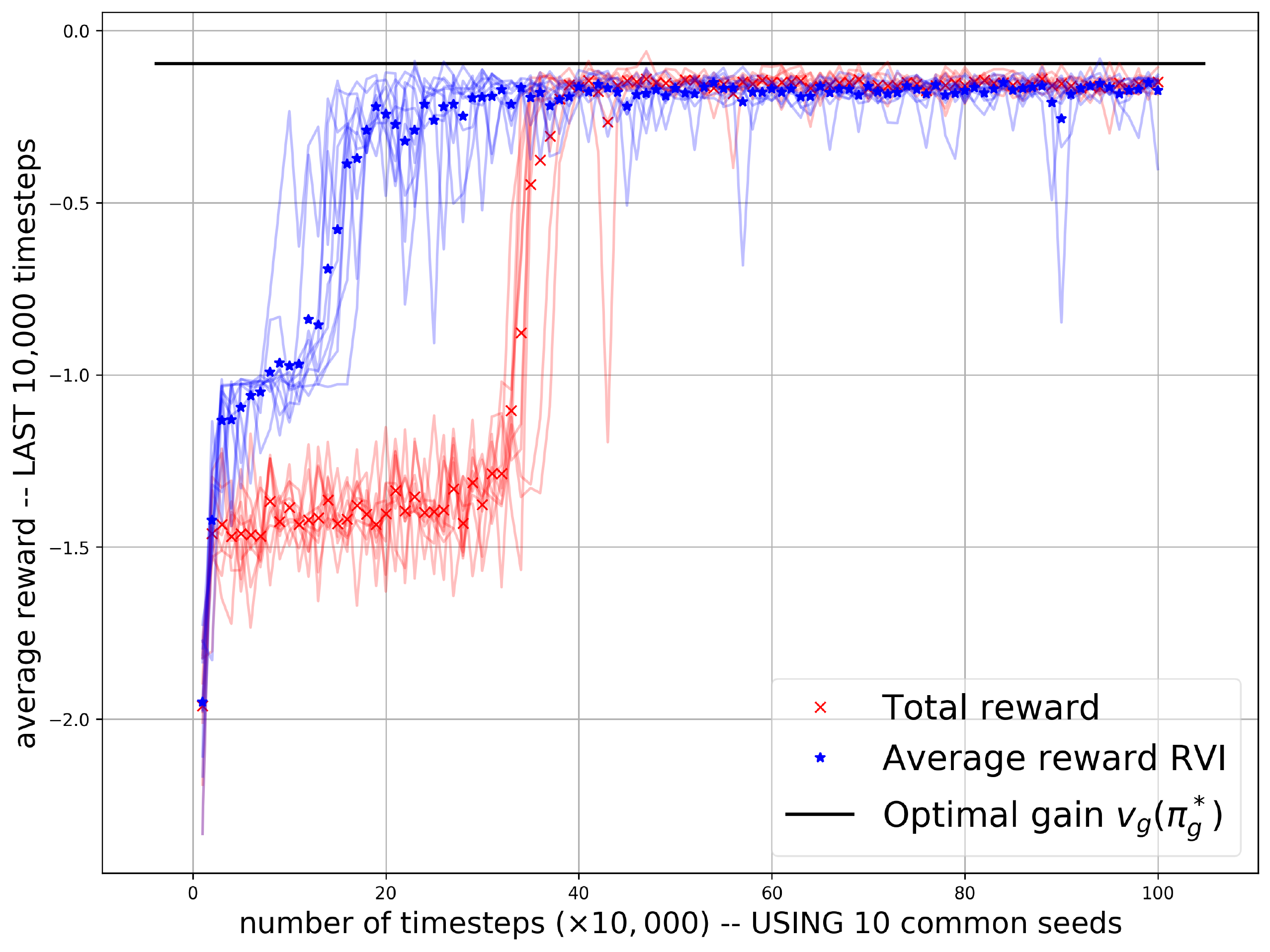}
\subcaption{Episodic: Taxi-15}
\label{lc_taxi15}
\end{subfigure}

\caption{Learning curves of
{\color{red} $Q_\mathrm{tot}$-learning on $\szrat$-model} and
{\color{blue} $Q_b$-learning on $\sreset$-model},
evaluated on two episodic environments.
$Q_\mathrm{tot}$-learning maximizes the total reward criterion,
whereas $Q_b$-learning maximizes the average-reward criterion.
For experimental setup, see \secref{sec:xprmt_setup_avgdisrew}.
}
\label{fig:eprep_xprmt}
\end{figure*}

%% file: discuss.tex
\section{Discussions} \label{sec:discuss_avgdisrew}

The route of using the discounted reward to approximately maximize
the average reward in RL seems to follow
Blackwell's $\gamma$-discounted approach (1962) to Howard's average reward (1960)
in DP.\footnote{
    Recall that both Blackwell's $\gamma$-discounted and Howard's average-reward
    approaches were aimed to deal with the infiniteness of the total reward
    in infinite-horizon MDPs, refer to \secref{sec:optcrit}.
}
As explained in \secref{sec:optcrit}, the major successive development of
such Blackwell's approach came from Veinott who introduced
a family of $n$-discount optimality criteria (1969), which is discounting-free.
Then, he developed an algorithm for obtaining $n$-discount optimal policies,
from which policies that are considered nearly-optimal and optimal by Blackwell
emerge as special cases, namely
$(n=0)$- and $(n=\infty)$-discount optimal policies, respectively.
Both imply the average reward optimality.

Like in DP, the route in RL should be completed by devising methods based on
Veinott optimality criteria.
To this end, average reward RL methods constitute the first step
in the following senses.
\textbf{First}, they yield a set of approximately $(n=-1)$-discount optimal policies,
from which higher $n$-discount optimal policies are sought,
as illustrated in \figref{fig:ndiscount_venn}.
This follows from the hierarchical property of $n$-discount optimality,
whose selectivity increases as $n$ increases.
Note that such a property is exploited in the $n$-discount policy iteration in DP
\citep[Ch 10.3]{puterman_1994_mdp}.
\textbf{Second}, average reward RL methods evaluate the gain $\vecb{v}_g \equiv \vecb{v}_{-1}$
and the bias $\vecb{v}_0$ that are likely to be useful for attaining higher $n$-discount criteria.
This is because the expansion coefficients $\vecb{v}_n$ in \eqref{equ:laurent_expansion}
are interconnected.
For instance, the first three coefficients satisfy three nested equations below,
\begin{equation}
\text{\emph{i)}}\ \vecb{v}_{-1} = \mat{P} \vecb{v}_{-1},\qquad
\text{\emph{ii)}}\ \vecb{v}_0 = \vecb{r} - \vecb{v}_{-1} + \mat{P} \vecb{v}_0, \qquad
\text{\emph{iii)}}\ \vecb{v}_1 = -\vecb{v}_0 + \mat{P} \vecb{v}_1,
\label{equ:vn_nested}
\end{equation}
where $\vecb{r}$ is the reward vector, $\mat{P}$ is the one-step transition matrix,
and $\vecb{v}_1$ is the policy evaluation entity for $(n=1)$-discount optimality.
Thus, advancement on gain and bias estimations is transferable to
higher $n$-discount methods towards obtaining Blackwell optimal policies.

\input{fig/venn_gamma_vs_n.tex}

Directly maximizing the average reward brings several benefits,
as described in \secref{sec:avgrew}.
It is also evidently free from any $\gamma$-related difficulties (\secref{sec:gamma}).
It is interesting now to discuss whether those benefits outweigh
the loss of all the virtues of $\gamma$ (\secref{sec:disrew}).

Approximating the average reward via discounting (\secref{sec:avgrewapprox})
is not without caveats.
Taking $\gamma$ really close to~1 slows the convergence, as well as
increases the variance of, for example, policy gradient estimates.
On the other hand, lowering $\gamma$ poses the risk of suboptimality
(with respect to the most selective criterion).
This trade-off can be potentially guided by
the critical discount factors $\gammabw$ (\secref{sec:disrew_bwoptim}).
Specifically in RL, we conjecture that
it is not the exact value of $\gammabw$ that is needed, but some value around it
because of the interplay among $\gamma$-dependent approximation layers,
such as those related to policy value and gradient estimations.
Thus, fine-tuning $\gamma$ is always a crucial yet non-trivial task,
dealing with possibly non-concave and non-differentiable learning performance functions
with respect to~$\gamma$
(even though the domain of such a univariate function is limited to $[0, 1)$).
Research on this front has been going on with some progress
\citep{zahavy_2020_stdrl, paul_2019_hoof, xu_2018_mgrad}.
In that regard, Veinott's discounting-free criteria (including average-reward optimality)
can be viewed as alternatives towards achieving the most selective Blackwell optimality in RL.

There are several ways to compensate for the other merits of discounting
(\secref{sec:discrew_mathplus}), which are missed due to directly maximizing the average reward.
We describe some of them as follows.

\paragraph{The chain-classification independence merit:}
The need for chain-type determination (or assumption) can be thought of as a way to
exploit the chain structural property to be able to apply more simple average-reward methods.
That is, for a constant gain across all states, we impose a unichain assumption,
which is relatively not restrictive in that it already generalizes the recurrent type.
Nevertheless, the ultimate goal remains: an average-reward RL method that
can handle non-constant gains across states as in multichain MDPs.
Because of its generality (hence, sophistication), it does not require
prior determination of the chain type.

The chain-type assumption can also be interpreted as an attempt to
break down the difficulty in attaining Blackwell-optimal policies in
one go (regardless of the chain structure) as in the $\gamma$-discounted optimality,
see \figref{fig:gammadiscounted_venn}.
Recall that in such a criterion,
only ($\gamma \ge \gammabw$)-discounted optimal policies are Blackwell optimal
(however the critical $\gammabw$ is generally unknown,
otherwise this one-go technique would be favourable).
On the other hand, the average reward (($n=-1$)-discount) optimality is
already equivalent to the Blackwell optimality whenever the MDPs are recurrent.
For more general unichain MDPs, ($n=0$)-discount optimality is
equivalent to the Blackwell optimality but only for
some transition and reward configuration
(for other configurations, higher $n$-discount criteria should be applied,
as shown in \figref{fig:ndiscount_venn}).

\paragraph{The contractive merit:}
The \emph{relative} value-iteration (VI) has been shown to alleviate
the non-contractive nature of \emph{basic} VI in average rewards
\citep{abounadi_2001_rviqlearn}.
There are also policy gradient methods whose convergence generally follows
that of stochastic gradient ascents.
Their policy evaluation part can be carried out via stochastic gradient descents.
That is, by minimizing some estimation loss derived for instance,
from the average-reward Bellman evaluation equation,
which involves the substraction of gain (similar to that of the relative VI).

\paragraph{The variance reduction merit:}
The variance of trajectory-related estimation can be controlled by
directly varying (truncating) the experiment-episode length
(as an alternative to varying the artificial discount factor).
One may also apply baseline substraction techniques for variance reduction.
In general however, lower variance induces higher bias-errors, leading to a trade-off.

We conclude that directly maximizing the average reward has a number of benefits
that make it worthwhile to use and to investigate further.
It is the root for approaching Blackwell optimality through
Veinott's criteria, which are discounting-free
(eliminating any complication due to artificial discounting).
Future works include examination about exploration strategies:
to what extent strategies developed for the discounted rewards applies to
RL aiming at discounting-free criteria.

%% file: fig/venn_gamma_vs_n.tex
\begin{figure*}
\centering

\begin{subfigure}{0.475\textwidth}
\resizebox{\textwidth}{!}{
\begin{tikzpicture}[set/.style = {circle, minimum size = 4.3cm, fill=blue!50}]
\node [set,
    fill= red,
    label={}] (A) at (0,0) {};
\node[set,
    fill=blue!75,
    label={}] (B) at (2.75,0) {};

\begin{scope}
\clip (0,0) circle(2.15cm);
\clip (2.75,0) circle(2.15cm);
\fill[purple](0,0) circle(2.15cm);
\end{scope}
\draw (0,0) circle(2.15cm);
\draw (2.75,0) circle(2.15cm);
\node at (1.4,0) [align=center]{may be \\ empty};

\node[draw, circle, minimum size=3.25cm, fill=green, label={}]
    (gammabw_set) at (3.25, -3.85){};
\node[draw, circle, minimum size=3.0cm, fill=brown, label={}]
    (gamma21_set) at (-0.25, -3.85){};
\draw [draw=black, line width=0.5mm]
    (5.5,2.25)
    rectangle
    (-2.25,-5.65);
\node at (-1,0){\Large $\Pi^*_{0 \le \gamma < \gamma_2}$};
\node at (3.5,0) {\Large $\Pi^*_{\gamma_1 \le \gamma < \gammabw}$};
\node at (-0.25, -3.85) {\Large $\Pi^*_{\gamma_2 \leq \gamma < \gamma_1}$};
\node at (3.25, -3.85){$\Pi^*_{\gammabw \le \gamma < 1} \eqdef \pistarbwset$};
\node at (5,1.5) {\large $\piset{S}$};
\end{tikzpicture}
} %
\subcaption{$\gamma$-discounted optimality for $\gamma \in [0, 1)$, whose
optimal policy sets are denoted as $\Pi_\gamma^*$.
Theoretically, there are a finite number of intervals.
For simplicity here, we assume that there are four intervals, namely: %
{\scriptsize $\textcolor{red}{[0=\gamma_3, \gamma_2)},
\textcolor{brown}{[\gamma_2, \gamma_1)},
\textcolor{blue}{[\gamma_1, \gamma_0=\gammabw)},
\textcolor{darkgreen}{[\gammabw, \gamma_{-1}=1)}$}.
}
\label{fig:gammadiscounted_venn}
\end{subfigure}
\hspace{3mm}
\begin{subfigure}{0.485\textwidth}
\resizebox{\textwidth}{!}{
\begin{tikzpicture}
\node[draw, circle, minimum size=8.5cm, fill=cyan, label={}]
    (gain_set) at (0,2.65){};
\node[draw, circle, minimum size=7.0cm, fill=orange, label={}]
    (bias_set) at (0,2.0){};
\node[draw, circle, minimum size=5.5cm, fill=pink, label={}]
    (n1_set) at (0,1.35){};
\node[draw, circle, minimum size=4.0cm, fill=yellow, label={}]
    (n2pp_set) at (0,0.7){};
\node[draw, circle, minimum size=2.5cm, fill=green, label={}]
    (nmax_set) at (0,0){};
\draw [draw=black, line width=0.5mm]
    (4.5, 7) %
    rectangle
    (-4.5,-1.9); %
\node at (0, 0.0) [align=center]
    {$\Pi_{\setsize{S} - 2}^* = \pistarbwset$};
\node at (0, 1.75) [align=center]
    {$\Pi_{\ldots}^* = \pistarbwset$ \\ (may be)};
\node at (0, 3.25) [align=center]
    {$\Pi_{1}^* = \pistarbwset$ \\ (may be)};
\node at (0, 4.75) [align=center]
    {$\Pi_{0}^* = \pistarbwset$ \\ (may be)};
\node at (0, 6) [align=center]
    {$\Pi_{-1}^* = \pistarbwset$ \\ (if recurrent)};
\node at (3.75, 6.25){\Large $\piset{S}$};
\end{tikzpicture}
} %
\subcaption{$n$-discount optimality for $n=-1, 0, \ldots, {\setsize{S} -2}$,
whose optimal policy sets are denoted by $\Pi_n^*$.
In recurrent MDPs, $\Pi_{-1}^* = \pistarbwset$.
When transient states are present, such an equivalency to $\pistarbwset$ may
not be achieved till $n=\setsize{S} - 2$ at most in unichain MDPs.
}
\label{fig:ndiscount_venn}
\end{subfigure}

\caption{Two Venn diagrams of optimal policy sets $\Pi^*$ of
$\gamma$-discounted and $n$-discount optimality criteria, applied to unichain MDPs.
There is only one green circle that is present in both Venn diagrams.
It indicates a subset $\pistarbwset$ containing stationary policies
that are guaranteed to be optimal with respect to the most selective Blackwell criterion.
Note that the sizes of circles in both diagrams are not to scale.
The left Venn diagram of $\gamma$-discounted optimality is
invariant to MDP classification and is based on
\citep{blackwell_1962_ddp, smallwood_1966_reg, jiang_2016_shp}, particularly
we conjecture that $\Pi^*_{\gammabw \le \gamma < 1}$ is always disjoint with
the other optimal policy subsets.
The right Venn diagram of $n$-discount optimality is based on
\citetext{\citealp[\thm{10.1.5, 10.3.6 }]{puterman_1994_mdp};
\citealp[\fig{2}]{mahadevan_1996_sensitivedo}}.
}
\label{fig:gamma_n_venn}
\end{figure*}
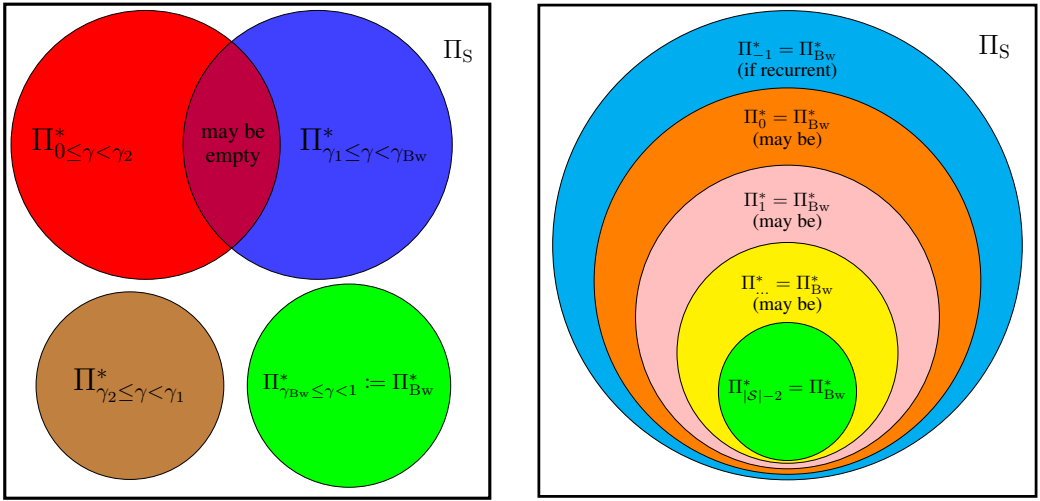

%% file: xprmt_setup.tex
\section{Experimental setup} \label{sec:xprmt_setup_avgdisrew}

In this Section, we describe the experimental setups used to produce
\figrefc{fig:gamma_xprmt}{fig:optim_landscape}{fig:eprep_xprmt}.
They are about environments (\secref{sec:avgdisrew_env})
and learning methods (\secref{sec:avgdisrew_learningmethod}).

\subsection{Environments} \label{sec:avgdisrew_env}

\paragraph{\textbf{GridNav-$n$ (episodic)}}
refers to an $n$-by-$n$ grid-world with no obstacle.
An agent has to navigate from an initial state to the goal state.
There are $n^2 - 1$ states, excluding the goal state.
Every state has four available actions,
\ie moving in North, East, South and West compass directions.
The state-transition stochasticity is governed by an action-slip parameter.
That is, the agent actually moves according to
the chosen (intended) directional action with a probability $q$.
Otherwise, it stays or moves in one of three other directions;
hence there are four alternatives due to slipping, each is with probability $(1 - q)/4$.
If an action brings the agent beyond the grid-world,
then the agent stays at the current grid.
There is a reward of $+1$ for transitioning to the goal;
any other movement costs $-1$.

\paragraph{\textbf{Taxi-$n$ (episodic)}}
refers to the $n$-by-$n$ grid Taxi environment,
which is adopted from OpenAI Gym \citep{brockman_2016_gym}.
The number of obstacles is increased accordingly with respect to~$n$,
whereas the number of passengers and pickup/drop-off locations is kept the same.

\paragraph{\textbf{The environment families 1, 2, and 3}}
are adopted from the access-control queueing task
\citep[\page{252}]{sutton_2018_irl},
the Chain problem \citep[\fig{1}]{strens_2000_bfrl}, and
the torus MDP \citep[\fig{2}]{morimura_2010_sdpg},
respectively.
In \figref{gammabw_envspecific}, the first two were used as Families 1 and 2,
whose numbers of states are varied, whereas the third was used as Family 3,
whose reward constant is varied.

\subsection{Learning methods} \label{sec:avgdisrew_learningmethod}

The learning method used for experiments in \figrefand{fig:gamma_xprmt}{fig:eprep_xprmt}
is $Q$-learning.
It is an iterative method that relies on the Bellman optimality equation (BOE)
to produce iterates approximating the optimal action value (denoted by $q^*$).
Different optimality criteria have different BOEs, inducing
different types of $Q$-learning as follows.
\begin{align*}
\text{Discounted rewards ($Q_\gamma$-learning):}
& \ \hat{q}_\gamma^*(s_t, a_t)
    \gets (1 - \alpha_t) \hat{q}_\gamma^*(s_t, a_t)
    + \alpha_t \{ r_{t+1}
    + \gamma \max_{a \in \setname{A}} \hat{q}_\gamma^*(s_{t+1}, a) \}, \\
\text{Average rewards ($Q_b$-learning):}
& \ \hat{q}_b^*(s_t, a_t)
    \gets (1 - \alpha_t) \hat{q}_b^*(s_t, a_t)
    + \alpha_t \{ r_{t+1} - \hat{v}_g^*
    + \max_{a \in \setname{A}} \hat{q}_b^*(s_{t+1}, a) \}, \\
\text{Total rewards ($Q_{\mathrm{tot}}$-learning):}
& \ \hat{q}_{\mathrm{tot}}^*(s_t, a_t)
    \gets (1 - \alpha_t) \hat{q}_{\mathrm{tot}}^*(s_t, a_t)
    + \alpha_t \{ r_{t+1}
    + \max_{a \in \setname{A}} \hat{q}_{\mathrm{tot}}^*(s_{t+1}, a) \},
\end{align*}
with a positive learning rate~$\alpha_t$ (here, we used a fine-tuned constant $\alpha$).
The estimate $\hat{q}^*$ is initialized optimistically to large values
to encourage exploration in the outset of learning.
For $Q_b$-learning, we set the optimal gain estimate
$\hat{v}_g^* \gets \max_{a \in \setname{A}} \hat{q}_b^*(\sref, a)$
with a prescribed (arbitrary but fixed) reference state~$\sref$,
following the relative-VI (RVI) technique
\ifpaper
by \citet[\secc{2.2}]{abounadi_2001_rviqlearn}.
\fi
Note that $Q_{\mathrm{tot}}$-learning converges as long as the total reward is
finite \citep[\page{2}]{schwartz_1993_rlearn}.